\documentclass[conference,10pt]{IEEEtran}
\usepackage{cite}
\usepackage{amsmath,amssymb,amsfonts}
\usepackage{algorithmic}
\usepackage{graphicx}
\usepackage{textcomp}
\usepackage{xcolor}
\def\BibTeX{{\rm B\kern-.05em{\sc i\kern-.025em b}\kern-.08em
    T\kern-.1667em\lower.7ex\hbox{E}\kern-.125emX}}

\usepackage{etex}

\usepackage{wrapfig}
\usepackage{graphicx}

\usepackage[mathscr]{eucal}
\usepackage{verbatim}

\usepackage{comment}
\usepackage{paralist}
\usepackage{mathtools}
\usepackage{amsfonts}
\usepackage{color}
\usepackage{graphicx}
\usepackage{tikz}
\usepackage{xspace} 
\usepackage{pgf}
\usetikzlibrary{arrows,automata}

\usepackage[usenames,dvipsnames]{pstricks}
\usepackage{epsfig}
\usepackage{pst-grad} 
\usepackage{pst-plot} 
\usepackage{pstricks-add}

\usepackage{float}
\usepackage{psfrag}
\usepackage{hyperref}
\usepackage{url}

\newcommand{\Sadegh}[1]{\textcolor{magenta}{#1}}

\tikzset{
	block/.style = {draw, rectangle,
		minimum height=1cm,
		minimum width=2cm},
	input/.style = {coordinate,node distance=1cm},
	output/.style = {coordinate,node distance=4cm},
	arrow/.style={draw, -latex,node distance=2cm},
	pinstyle/.style = {pin edge={latex-, black,node distance=2cm}},
	sum/.style = {draw, circle, node distance=1cm},
}

\usepackage{hyperref}

\begin{document}

\title{Perception-in-the-Loop Adversarial Examples}

\author{}

 \author{\IEEEauthorblockN{Mahmoud Salamati}
 \IEEEauthorblockA{\textit{MPI-SWS} \\
 msalamati@mpi-sws.org}
\and
 \IEEEauthorblockN{Sadegh Soudjani}
 \IEEEauthorblockA{\textit{Newcastle University} \\
 Sadegh.Soudjani@newcastle.ac.uk}
 \and
 \IEEEauthorblockN{Rupak Majumdar}
 \IEEEauthorblockA{\textit{MPI-SWS} \\
 rupak@mpi-sws.org}
 }

\maketitle

\begin{abstract}
We present a scalable, black box, perception-in-the-loop technique to find adversarial examples for
deep neural network classifiers.
Black box means that our procedure only has input-output access to the classifier, and not to
the internal structure, parameters, or intermediate confidence values.
Perception-in-the-loop means that the notion of proximity between inputs can be directly queried from
human participants rather than an arbitrarily chosen metric.
Our technique is based on covariance matrix adaptation evolution strategy (CMA-ES), a black box optimization
approach. 
CMA-ES explores the search space iteratively in a black box manner, by generating populations of candidates according
to a distribution, choosing the best candidates according to a cost function, and updating the posterior
distribution to favor the best candidates.
We run CMA-ES using human participants to provide the fitness function, using the insight that
the choice of best candidates in CMA-ES can be naturally modeled as a perception task: pick the top $k$ inputs
perceptually closest to a fixed input.

We empirically demonstrate that finding adversarial examples is feasible using small populations and few iterations.
We compare the performance of CMA-ES on the MNIST benchmark with other black-box approaches using $L_p$ norms
as a cost function, and show that it performs favorably both in terms of success in finding adversarial examples and in
minimizing the distance between the original and the adversarial input.
In experiments on the MNIST, CIFAR10, and GTSRB benchmarks, we demonstrate that CMA-ES can find perceptually similar
adversarial inputs with a small number of iterations and small population sizes when using perception-in-the-loop.
Finally, we show that networks trained specifically to be robust against $L_\infty$ norm can still be susceptible to 
perceptually similar adversarial examples.
Overall, we demonstrate that adversarial examples can be constructed in a black box way through direct human interaction.
\end{abstract}



\section{Introduction}

It is well known by now that deep neural networks are susceptible to \emph{adversarial examples}:
given an input, one can add a small perturbation to produce a new, perceptually similar, input that is classified differently
but with high confidence (e.g. \cite{Goodfellow:2014,Szegedy:2013}).
Adversarial examples exist across many different domains and can be reproduced in physical artifacts. 
Not only do adversarial examples exist, they can be found comparatively easily through many different techniques, even when
access to the classifier is limited to a black box setting.
Clearly, they give rise to immediate safety and security concerns.
Understanding the scope and prevalence of adversarial inputs, leading to their mitigation, 
is therefore a key challenge facing the deployment of DNNs in security or safety critical domains.

A key implicit assumption in an adversarial example is that the resulting input is \emph{perceptually identical} to the original input.
Indeed, in many situations, one desires the classifier to replace human decision making and an example should be considered 
adversarial if the classification task is easy for human observers but where the machine is wrong in unpredictable ways.
In research on generation or mitigation of adversarial inputs, though, it has been difficult to perform perception in the loop:
the underlying search technique exercises the network thousands of times, and it is impractical to ask the user at each step.
Thus, most research has focused on using a metric on the input space, typically the $L_0$, $L_1$, $L_2$, or $L_\infty$ norms, as a
surrogate for human perception.
Informally, for an image classification task, the $L_0$ norm measures how many pixels differ, the $L_1$ 
norm measures the sum of all pixel differences, the $L_2$ norm the Euclidean distance, and the $L_\infty$ norm the maximum difference
in corresponding pixel values. 
Thus, algorithms to generate adversarial examples minimize the $L_p$ norm between an original image and a new image that has a different
classification and algorithms to mitigate against adversarial examples assume that a network is robust if it classifies close
inputs in the $L_p$ norm in the same way.

Unfortunately, this fundamental assumption may not hold.
Several recent results have pointed out that these norms do \emph{not} form reasonable measures of human perception of image similarity: 
inputs imperceptible to human observers may have high $L_p$ norms, and vice versa \cite{Sharif:2018}. 
This presents two technical obstacles to developing ``human-level'' classifiers.
First, we know how to optimize for the wrong thing (an $L_p$ norm)
and we do not know how to scale the search with humans in the loop.
Second, humans are  better at providing \emph{preferences} but not at providing quantitative norms.
Thus, there is a problem of translating human perception of similarity into a norm usable by existing techniques. 

In this paper, we present a method to find adversarial examples in DNNs in a \emph{black box} setting that
can directly use \emph{human perception in the loop}.
Black box means that we only have access to the input and the output of the network, but not to any other information such as the structure,
the confidence levels at the output, or parameters of internal nodes.
Perception in the loop means that our optimization procedure can explicitly ask human subjects 
their perceptual preferences in order to select inputs during the search.

We overcome the two technical difficulties in the following way.
First, we base our search for adversarial examples on an \emph{evolutionary search} technique.
The specific search strategy we use is \emph{covariance matrix adaptation}-evolutionary strategy (CMA-ES) \cite{Hansen:2005}.
CMA-ES is a black box, gradient-free, search.
It keeps a population of candidates, generated according to a distribution, and updates the distribution
based on the ``fittest'' subset of candidates according to a fitness function.
Specifically, in our context, CMA-ES works without any information about the structure of the network or 
confidence levels or gradients.
In a first step, we show that CMA-ES with $L_p$ norms can already outperform existing grey-box and black-box approaches
(including those in \cite{Papernot:2017,Mustafa:2018,Marta:2018}). 
Moreover, we empirically show that CMA-ES can find adversarial examples using small populations and few iterations:
this sets the stage to include humans in the loop without high performance penalties.

Nominally, the top candidates in CMA-ES are chosen by ranking the candidates
on the basis of a cost or ``fitness'' function, such as an $L_p$ metric.
Our insight is that CMA-ES does not, in fact, require a metric, but only an oracle to identify the ``top $K$'' fittest
individuals in the population. 
Thus, one could pick the top candidates based purely on perceptual preferences.
In this way, our search is metric-free: in each phase, our CMA-ES search asks a human participant to pick the ``top $K$''
inputs from the current population which are perceptually closest to a fixed, initial, input.
 
We show empirically that for standard benchmark sets (MNIST, CIFAR10, GTSRB), CMA-ES converges to an adversarial example
with only a small population (at most 20) and in a small number of iterations (at most 200).
Thus, it is feasible to ask a human participant a small number of questions (in our experiments, around 10) that already
finds a misclassified example.
We also demonstrate through a psychophysics experiment that perception of closeness among different human participants is
robust to individual differences, and can be different from an $L_p$ norm.

Altogether, we get a black-box adversarial input generator based directly on perceptual preferences.
We compare the performance of CMA-ES against three main techniques: substitute training algorithm from \cite{Papernot:2017}, 
the GA algorithm from \cite{Mustafa:2018}, and the MCTS algorithm from \cite{Marta:2018}. 
Our first set of experiments uses $L_1$ and $L_\infty$ norms and shows that black-box CMA-ES can be 
equal or more efficient in comparison with existing black- or grey-box techniques, both in the number of successes and in 
the average perturbation added.
Next, we use CMA-ES to find adversarial examples on different datasets using human perception in the loop.
Finally, we have also run our technique on networks that deliberately incorporate robustness in the training \cite{Vechev:2018}.
We show that while these networks are designed to be robust against $L_\infty$ norm, we still find adversarial examples
which are perceptually indistinguishable.
Thus, we believe mitigation techniques should explicitly incorporate perceptual preferences as well.

We summarize our contributions.
\begin{itemize}
\item We use CMA-ES, an efficient black box optimization technique, to craft adversarial examples in a fully black box manner, without
requiring knowledge about the network nor requiring training a new model.
 \item
We propose the first perception-in-the-loop search for crafting adversarial examples and empirically demonstrate
its efficiency on MNIST, CIFAR10, and GTSRB benchmarks. 
\item
We demonstrate experimentally that perceptual preferences are robust across individuals.
\item Using CMA-ES on robustly trained networks, we argue that networks should be made robust using human perception instead of other norms.
%
\end{itemize}

\section{Related Work}
\label{sec:related_work}

\paragraph{\textbf{White-box adversarial examples}}
Vulnerability of deep neural networks (DNNs) to adversarial inputs was pointed out in \cite{Szegedy:2013,Biggio:2013}. 
Such adversarial examples were initially constructed by assuming full knowledge of the network's architecture and its parameters.
For example, \cite{Szegedy:2013} optimized the network's prediction error and \cite{Goodfellow:2014} used a fast gradient sign method to
search for perturbed inputs with incorrect classification.
The latter technique 
was extended in \emph{DeepFool} \cite{Moosavi:2016} using variable gradient steps.

The papers \cite{Goodfellow:2014} and \cite{Moosavi:2016} formulate the problem in a non-targeted manner, i.e., 
crafting adversarial examples for having an arbitrary misclassification (i.e., a classification different from the original input). 
Later, \cite{papernot:2016} proposed a \emph{targeted} misclassification algorithm for finding perturbed inputs that 
result in a specific output classification.
All these algorithms are \emph{white box}: they use knowledge of the DNN architecture in the search for misclassifying perturbations.

\paragraph{\textbf{Black and grey-box adversarial examples}}
In contrast to the \emph{white-box} approaches, \cite{Papernot:2017} proposed a \emph{black-box} approach that does not require any 
knowledge of the network. 
Their approach requires access to the input-output data only 
but assumes accessibility of a fraction of the test data with which the DNN was evaluated.
The approach of \cite{Papernot:2017} has two stages for crafting adversarial examples.
First, it feeds the DNN with significant amount of data, stores collected input-output pairs, and uses that 
data as labeled examples to train a substitute neural network with possibly different architecture.
Once the substitute network is trained, they employ white-box methods such as the one in \cite{Goodfellow:2014} 
to craft misclassifying examples, relying on a transference heuristic.
The main disadvantage of this approach is that the adversarial examples are crafted on the substitute network, without 
any guarantee on misleading the original DNN. 
The implementation results of \cite{Papernot:2017} also reveal that the transferability of the crafted adversarial samples is in general not very high.

Grey-box approaches craft adversarial inputs based on the input-output behavior as well as some extra information such as the 
output of the DNN's softmax layer.
For example, \cite{Marta:2018} craft adversarial examples in image classification tasks by 
first identifying the key points of the input image using object detection methods such as SIFT (Scale Invariant Feature Transform) and
then using Monte Carlo tree search to select pixels around the key points which affect the softmax layer (the confidence levels) the most. 
In \cite{Shiva:2017} another query-based local-search technique is proposed to choose pixels to perturb which minimize confidence 
level associated with true class of the input image.  	
\cite{Mustafa:2018} provides a gradient-free approach for crafting adversarial 
examples, based on Genetic Algorithms (GA). 
Later, \cite{Rohan:2018} combines GA with a gradient-estimation technique to craft adversarial examples for applications in 
automatic speech recognition. 
\cite{Song:2017,Zoo}, use finite differences to estimate gradiant of loss function which is then leveraged 
in white-box based attacks. 
In \cite{IIyas:2017} a query-efficient method for estimating the gradient is proposed which is then leveraged to performs a projected gradient 
descent (PGD) update. However, their method requires several thousands of queries on average to mislead the DNN.

\paragraph{\textbf{Mitigation strategies}}
The widespread existence of adversarial attacks
led to research on mitigation strategies. 
The research can be divided into two categories: 
either improve the DNN robustness or detect adversarial examples.
Strategies in the first category target at improving the training phase in order to increase the robustness and classification performance of the DNN. 
\cite{Goodfellow:2014} suggests that adding adversarial samples to the training set can act as a regularizer. 
It is also observed in \cite{Papernot:2017} that training the network with adversarial samples makes finding new adversarial samples harder.
\cite{GAN14} shows that new samples can be obtained from a training set by solving a game between two DNNs.
Recent papers \cite{Vechev:2018,AI2Robust18} have used \emph{abstract interpretation} for training DNNs that are guaranteed to be robust.
In particular, \cite{Vechev:2018} constructs approximations of \emph{adversarial polytopes},
which describe the set of possible neural network outputs given the region of possible inputs,
 and trains the network on the entire input space. This results in networks that are more robust.
Most of this line of research pick a norm ($L_1$ or $L_\infty$) as a measure for robustness: the loss function
during training ensures there is a neighborhood (in the chosen norm) around test inputs with the same classification.
However, we show that our technique can find adversarial examples under the perceptual norm even on robustly trained
networks.

The second category of strategies is reactive to the inputs and tries to identify adversarial samples by evaluating the regularity 
of samples \cite{Papernot:2017}. 
These strategies are not effective against samples crafted by our approach: 
the optimization has human perception in the loop and crafts samples without irregularities seen in the outcome of other approaches.

\paragraph{\textbf{Choosing dissimilarity measures}}
An adversarial example is found by adding perturbation to the original input image until the classifier is misled. 
Obviously, the classifier is not expected to classify the adversarial example the same as for the original input image if the 
amount of perturbation is high. 
An implicit assumption in most work on adversarial examples is that the inputs are \emph{perceptually indistinguishable} to a human
observer (see e.g., \cite{Papernott:2016,Carlini:2017,Engstrom:2017}). 
Most approaches for crafting adversarial examples have been founded on optimising a cost (or loss) function, typically 
$L_p$ norms, $p\in\{\infty,0,1,2,\cdots\}$, as a proxy for human perception
\cite{Papernot:2017,Goodfellow:2014,Grosse:2016,Szegedy:2013}. 
For any two vectors $v = [v_1,v_2,\ldots,v_n]^T$ and $v' = [v'_1,v'_2,\ldots,v'_n]^T$ representing two images, their 
dissimilarity can be measured by $L_0$ norm as $||v-v'||_{0} =\#\{i,\,v_i\neq v_i'\}$, 
by $L_\infty$ norm as $||v-v'||_{\infty} =\max_i\{|v_i-v_i'|\}$, 
or by $L_p$ norm, $p\in\{1,2,\ldots\}$, as
\begin{equation*}
||v-v'||_{p}=(\Sigma_{i=1}^n |v_i-v_i'|^p)^{\frac{1}{p}}.
\end{equation*}
%
In the context of adversarial examples, $L_0$ measures the number of perturbed pixels, 
$L_1$ gives the sum of absolute perturbations, and  $L_{\infty}$ measures the maximum perturbation applied to the pixels.
The computation of $L_p$ norm can be performed rapidly in computational tools; this is important because current
techniques rely on a large number of evaluations of the dissimilarity measure. 
Thus, strategies for crafting adversarial examples and for mitigating them, rely on the
assumption that the $L_p$ norm evaluates dissimilarities close enough to the way humans interpret these differences.

Unfortunately, this assumption is not well-founded:
inputs close in the $L_p$ norms may be neither necessary nor sufficient for images to be perceptually similar \cite{Moosavi:2017,Engstrom:2017,Sharif:2018}.
%
Thus, a major open question is to incorporate perceptual measures directly in the training of machine learning classifiers.\footnote{
	Of course, there are situations in which perceptual similarity is neither possible (e.g., in classification tasks
	unrelated to perception) nor the goal of an attack \cite{Nguyen2015}; we focus on important scenarios where perception does
	play an important role.
}
The main challenges are that perceptual preferences do not directly provide a (differentiable) cost function,
and involving a perceptual task in an exploration loop makes the search very slow.

We solve the problem of crafting adversarial inputs with perceptual similarity in the following way.
First, our approach is black-box, and does not require finding gradients. 
Second, our approach using CMA-ES does not require a metric space on inputs, but only requires selecting the ``top $K$'' closest
inputs. 
Thus, we naturally incorporate an easy perception task (``pick the preferred $K$ images'') in the search.
We show empirically that the process converges in only a few iterations, making it feasible to find examples even with the 
cost of human perception in the loop.
Finally, we show that the perceptual task is robust across humans and that top-$K$ perceptual selection performed by humans can be
different from selecting the top $K$ according to an $L_1$ norm.

Use of perceptual preferences in learning and classification have been approached in other domains \cite{MartinGS12,Christiano:2017,Akrour:2017}.
Our work demonstrates the applicability of perceptual preferences in the task of finding adversarial examples.
We note that there is recent work on the ``dual problem'' of finding adversarial examples which fool both humans and machines \cite{Elsayed:2018}.
Whether our approach generalizes to these problems or whether mitigation measures can use perception-in-the-loop is left for future work.


\section{The Adversarial Example Problem}
\label{sec:problem_def}

\subsection{Deep Neural Networks}


A \emph{classifier} $G: \mathbb R^d \rightarrow \{1,2,\ldots, L\}$ is a function
that maps an input $x\in\mathbb R^d$ to one of $L$ possible labels.
A deep neural network (DNN) $\mathfrak N$ with \emph{depth} $n$ is a parameterised representation of a classifier
$\mathcal{G}(\boldsymbol{\theta}, \cdot):\mathbb R^d\rightarrow \mathbb R^{L}$ that is constructed by a forward combination of $n$ functions:
\begin{equation}
\label{eq:DNN}
\mathcal{G}(\boldsymbol{\theta}, x):=\mathcal G_n(\theta_n,\mathcal G_{n-1}(\theta_{n-1},\cdots \mathcal G_2(\theta_{2},\mathcal G_1(\theta_1,x)))),
\end{equation}
where $x$ denotes the input to the DNN and $\mathcal G_i(\theta_i,\cdot)$, $i\in\{1,2,\ldots,n\}$, denote non-linear functions (called \emph{layers}),
each parameterised by $\theta_i$. 
Each layer of the network $\mathcal G_i$ takes its input as a representation of the input data and transforms it to another 
representation depending on the value of parameters $\theta_i$. 
The label associated to input $x$ by the DNN $\mathfrak N$ is obtained by finding the component of $\mathcal{G}(\boldsymbol{\theta}, x)$ with the highest value: 
\begin{equation}
\label{eq:max_layer}
\mathfrak N(\boldsymbol{\theta},x)=\arg\max\limits_{j\in\{1,2,\cdots,L\}}\mathcal{G}^{(j)}(\boldsymbol{\theta}, x),
\end{equation}
where $\mathcal{G}^{(j)}$ is the $j$th element of the output vector $\mathcal{G}$.

During the \emph{training phase} of the network, the set of parameters $\boldsymbol{\theta}$ is learnt using 
a large number of labeled input-output pairs $\{(x_i,y_i)\mid i=1,2,\ldots,m\}$ 
in order to achieve the highest performance by $\mathfrak N(\boldsymbol{\theta},\cdot)$ in mimicking the 
original unknown classifier $G(\cdot)$ on the available data.
During the \emph{test phase} of the netowrk, $\mathfrak N$ is deployed with the fixed learned parameters $\boldsymbol{\theta}_\ell$ 
and used to make predictions about the lables of inputs not seen during the training. 
The network is expected to generalize the training data set and to make accurate classifications on the test data set.

\subsection{Adversary's Goal}

Given a network $\mathfrak N$ with learned parameters $\boldsymbol{\theta}_{\ell}$ and an input $x$, 
the goal of the adversary is to find an input $x'$ such that 
$\mathfrak N(\boldsymbol{\theta}_{\ell},x')\ne \mathfrak N(\boldsymbol{\theta}_{\ell},x)$, 
i.e., the two images are classified in two different classes, while $x,x'$ are ``similar'' as much as possible.
One way to assess similarity is to define a metric space $\|\cdot\|^*$ on the space $\mathbb R^d$ of inputs, for example,
using an $L_p$ norm.
Then, the \emph{non-targeted} misclassification problem is to find 
 \begin{equation}
	\label{eq:obj_fun}
	x_{\textsf{adv}}(x,\mathfrak N)=\arg \min_{x'} \left\{\|x-x'\|^*,\mid \mathfrak N(\boldsymbol{\theta}_{\ell},x')\ne \mathfrak N(\boldsymbol{\theta}_{\ell},x)\right\}.
\end{equation}
The problem is \emph{non-targeted} because we only require a misclassification and not any specific label for $x_{\textsf{adv}}$. 

Alternatively, we can require a specific label for $x_{\textsf{adv}}$, and define the \emph{targeted} misclassification problem: 
 \begin{equation*}
	\label{eq:obj_fun_targeted}
	x_{\textsf{adv}}(x,j,\mathfrak N)=\arg \min_{x'} \left\{\|x-x'\|^*,\,\, \mathfrak N(\boldsymbol{\theta}_{\ell},x') = j\right\},
\end{equation*}
for any $j\neq \mathfrak N(\boldsymbol{\theta}_{\ell},x)$.

We present our results for non-targeted misclassification but the results are applicable to the targeted case as well.
In the experimental results of Section~\ref{sec:simulations}, we use $L_1$ and $L_{\infty}$ norms for measuring distances between inputs 
in order to compare our results with the approaches in the literature. 

For perception in the loop, we use the following formulation:
  \begin{equation}
	\label{eq:obj_fun_percep}
	x_{\textsf{adv}}(x,\mathfrak N)=\arg \min_{x'} \left\{d_{\textsf{per}}(x,x'),\, \mathfrak N(\boldsymbol{\theta}_{\ell},x')\ne \mathfrak N(\boldsymbol{\theta}_{\ell},x)\right\},
\end{equation}
where we require that the function $d_{\textsf{per}}(x,x')$ is such that for any fixed $x$, $d_{\textsf{per}}(x,\cdot)$ induces a total
order on the set of inputs.
Formally, $d_{\textsf{per}}(x,\cdot)$ is antisymmetric and transitive, and for any $x', x''$, either 
$d_{\textsf{per}}(x,x')\le d_{\textsf{per}}(x,x'')$ or $d_{\textsf{per}}(x,x'')\le d_{\textsf{per}}(x,x')$. 
With these properties, we can solicit preferences from human users to select the top $k$ inputs from a set which are ``most similar'' to a given input.

\subsection{Adversary's Capabilities} 

We work in the black-box setting.
The adversary can query the DNN $\mathfrak N$ on a given input $x$ and receive the output label $\mathfrak N(\boldsymbol\theta_\ell, x)$.
However, the adversary does not know the depth, the parameters $\theta_\ell$, or the confidence levels.
This assumption makes adversaries considered in our work weaker but more realistic in comparison  with the ones in 
\cite{Szegedy:2013,Goodfellow:2014,Marta:2018}, since output labels contain less information about the model's behavior.
Note that gradient based methods of \cite{Szegedy:2013,Goodfellow:2014,papernot:2016}
cannot be utilized for crafting adversarial examples since the
adversary has no knowledge of the DNN's parameters or confidence levels.

\begin{figure}[t]
	\begin{center}
		\resizebox{9cm}{5cm}{
		\begin{tikzpicture}[auto, node distance=1cm,>=latex',scale=.5]

		\node  at (0,0)
		{\includegraphics[scale=.15]{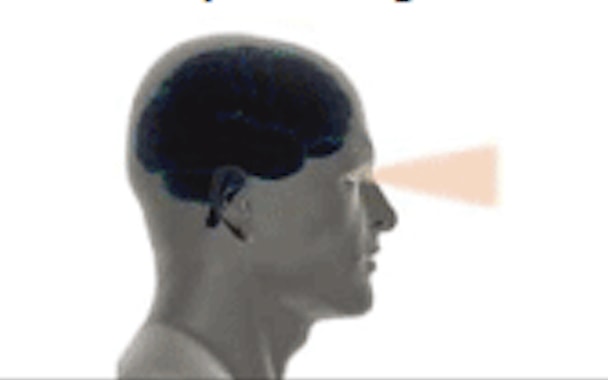}};
		
		\node (TV) at (10,1)
		{\includegraphics[scale=.29]{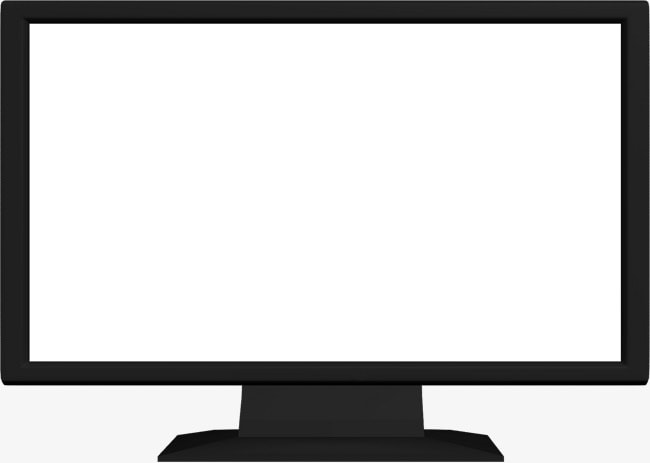}};
		
		\node at (11.75,1.5)
		{\includegraphics[scale=.045]{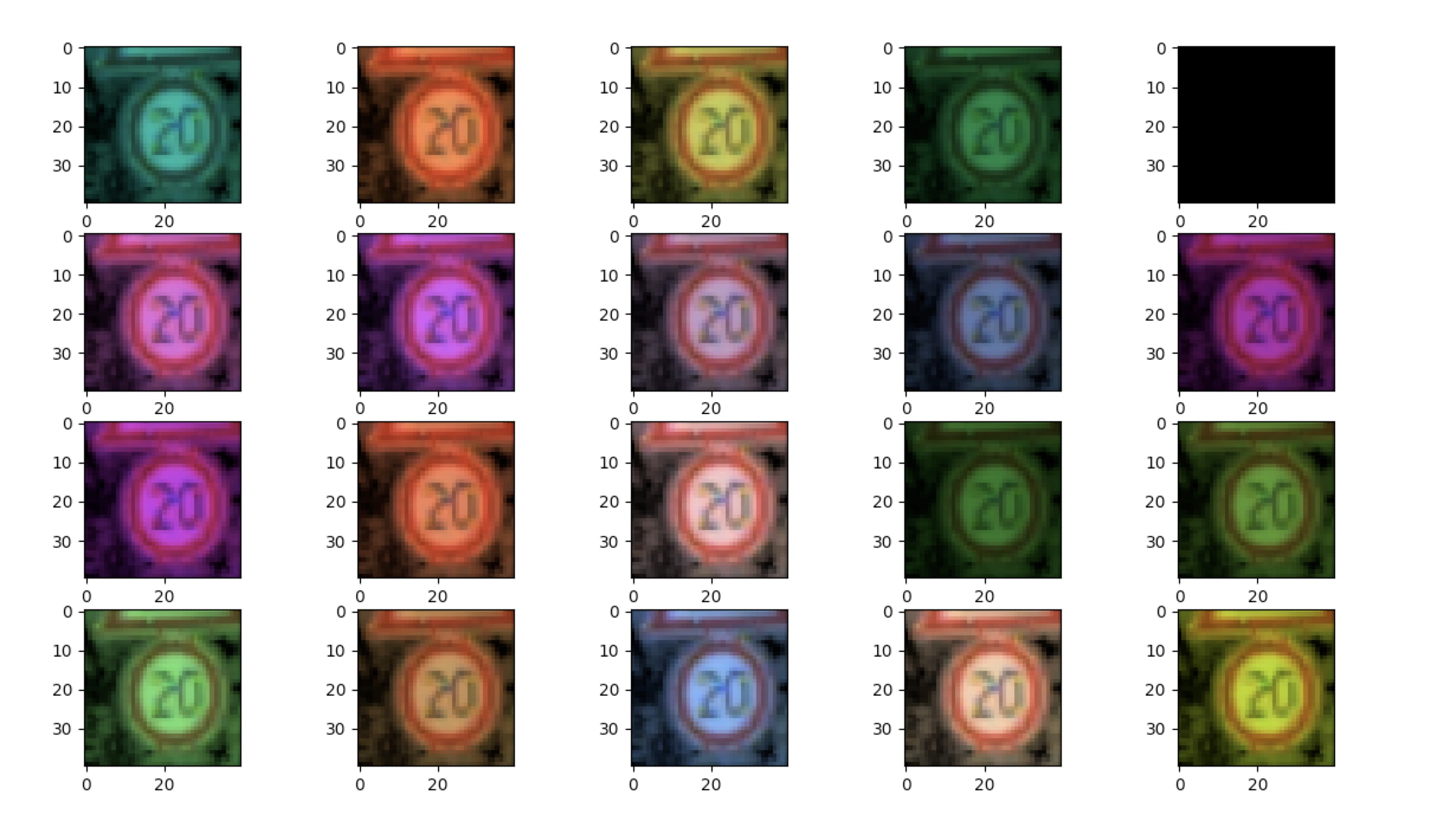}};
		
		\node at (5.75,1.75)
		{\includegraphics[scale=.04]{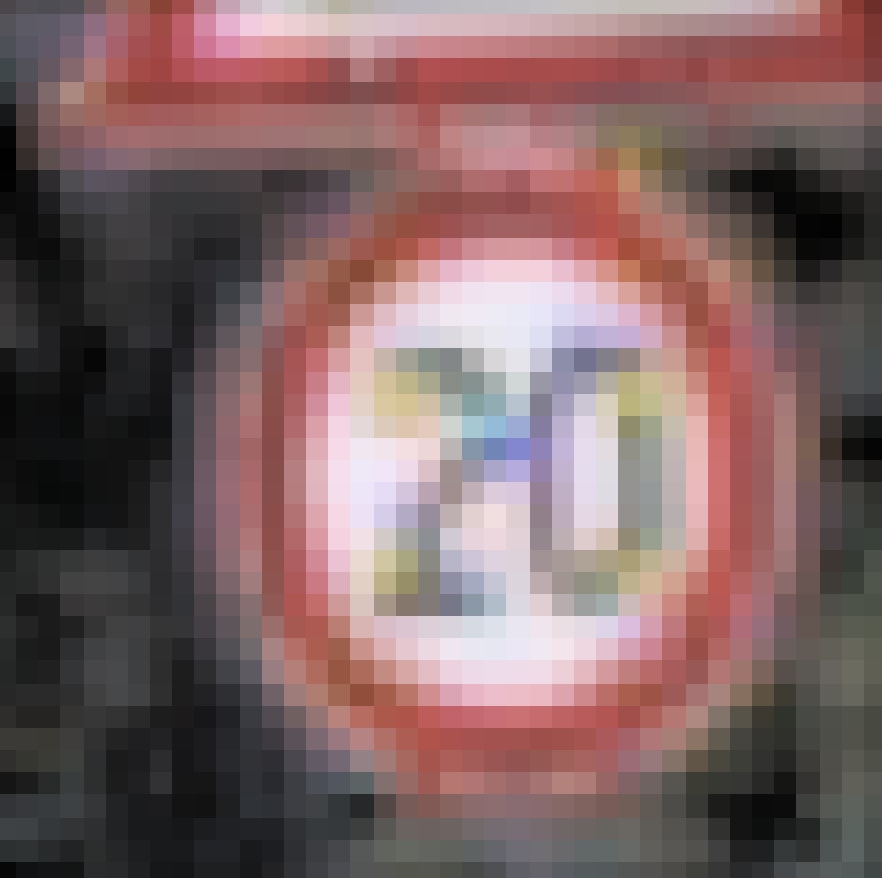}};
		
		\node at (23,8)
		{\includegraphics[scale=.25,width=1cm,height=1cm]{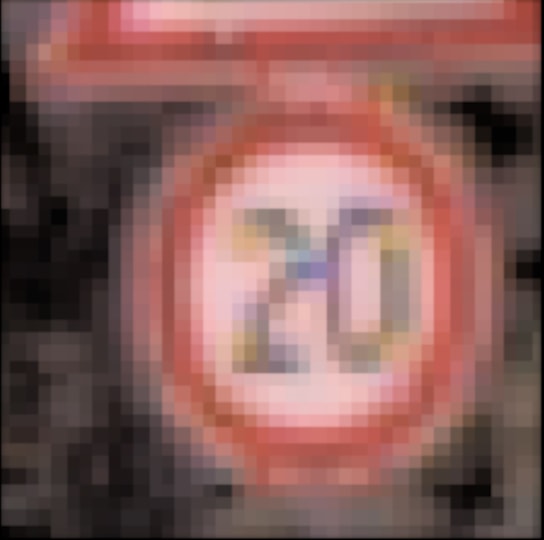}};
		\node at (23,5.5)
		{\includegraphics[scale=.25,width=1cm,height=1cm]{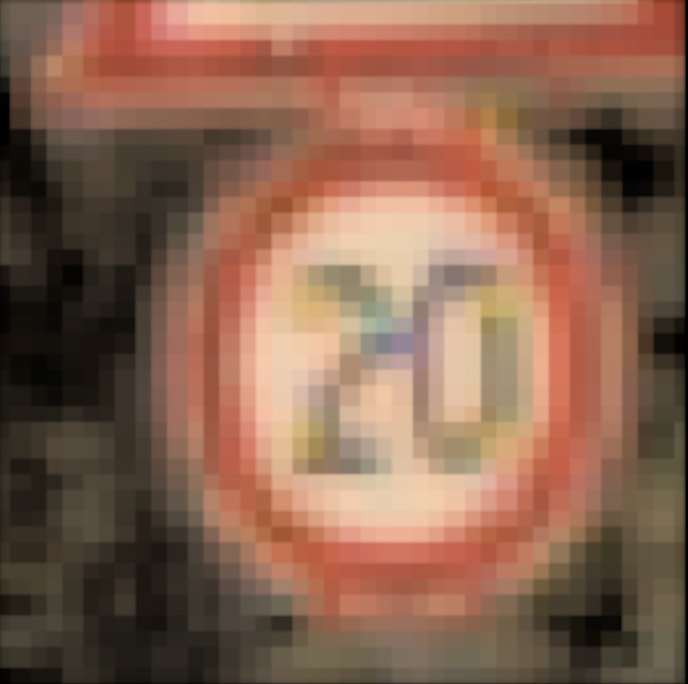}};
		\node  at (23,3)
		{\includegraphics[scale=.25,width=1cm,height=1cm]{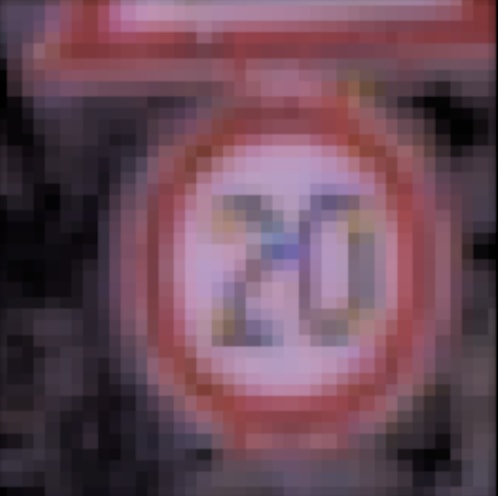}};
		\node (parents) at (23,.5)
		{\includegraphics[scale=.25,width=1cm,height=1cm]{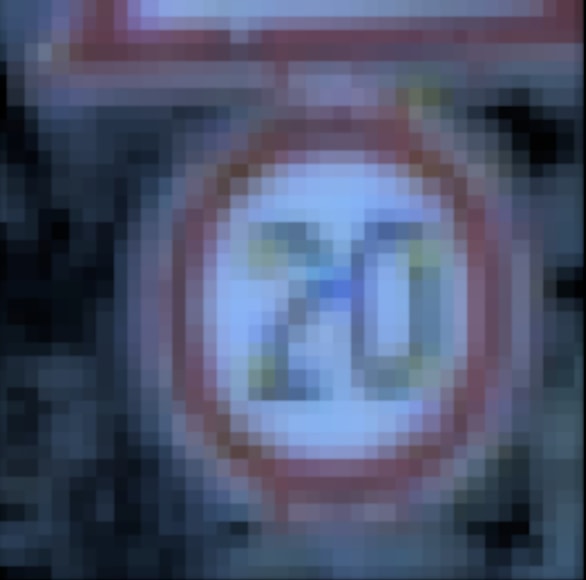}};
		\node at (23,-2)
		{\includegraphics[scale=.25,width=1cm,height=1cm]{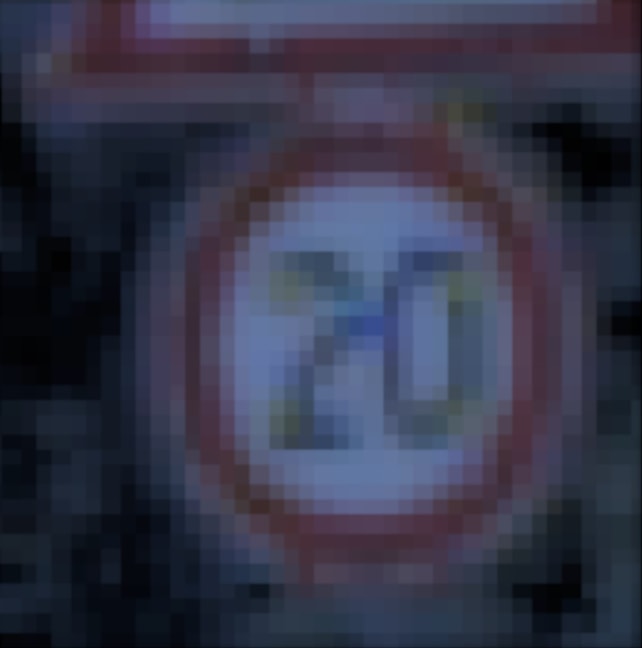}};
		
		\node [block,scale=1,line width=1mm] (system) at (15,10) {\textbf{CMA-ES Algorithm}};
		
		\node (input) at (0,11.5)
		{\includegraphics[scale=.25,width=2cm,height=2cm]{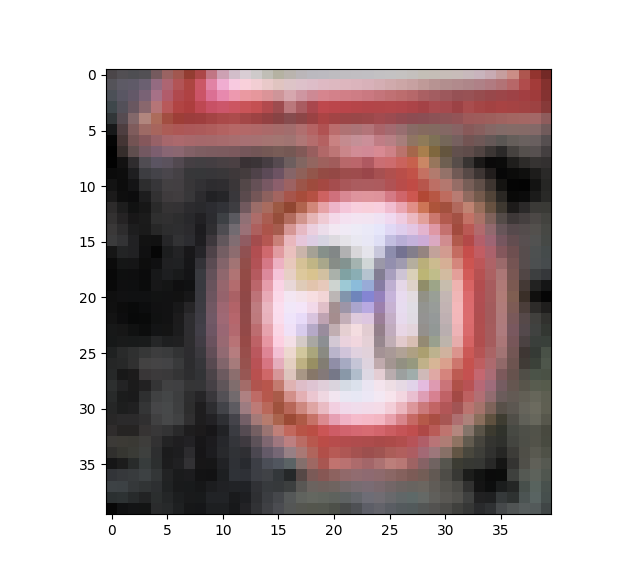}};

		\draw[->,line width=1mm] (2,12.5)-- (14.5,12.5)-|(14.5,11)  ;
		
		\draw[->,line width=1mm] (system)-|(TV)  ;
		\draw[->,line width=1mm] (16.5,0)--(21.5,0)  ;
		\draw[->,line width=1mm] (24.25,0)-- (27,0) |- (18,10)  ;
		\draw[draw=black,line width=1mm] (21.75,-3.25) rectangle (24.25,9.25);

		\node at (0,8.5) {\textbf{Input image}};
		\node at (8.5,8) {($\mathbf{1}$)};
		\node at (18.5,-1) {($\mathbf{2}$)};
		\node at (28,6) {($\mathbf{3}$)};
		\end{tikzpicture}
	}
	\end{center}
	\caption{Running CMA-ES with human in the loop that includes (1) generating certain number of adversarial candidates by the CMA-ES; (2) selecting top candidates most similar to the reference image by the user; (3) feeding selected candidates back to the CMA-ES.}
	\label{fig:CMAES_Process}
\end{figure}
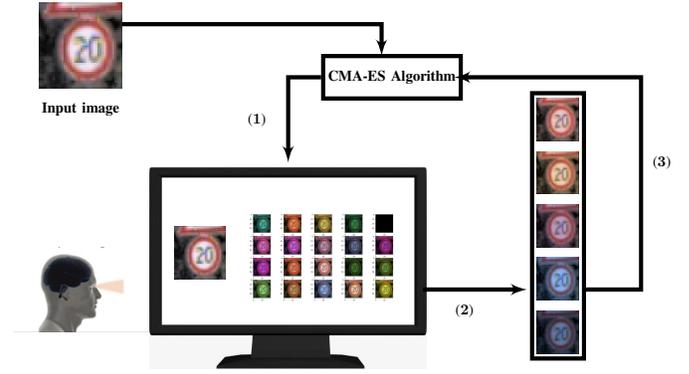

\section{Covariance Matrix Adaptation Evolution Strategy (CMA-ES)}
\label{sec:CMAES}

\subsection{Basic Algorithm}

CMA-ES \cite{Hansen:2005} is an efficient optimization method belonging to the class of evolutionary algorithms.
It is stochastic and derivative-free, and can handle non-linear, non-convex, or discontinuous optimization problems. 
It is based on the principle of biological evolution that is inspired by the repeated interplay of 
variation and selection.

The algorithm proceeds in generations. 
In each generation, it generates a new population of candidate solutions by \emph{variation} of
the ``best'' candidates of the previous generation in a stochastic way.
Then, it ranks a subset of the current population according to a fitness function; this subset
becomes the parents in the next generation.
This leads to get more and more individuals with better fitness over successive generations.
CMA-ES has been shown to be very efficient on a number of domains \cite{Hansen:2004,Hansen:2003},
without requiring large sample populations or carefully scaled fitness functions \cite{Hansen:2005}.

New individuals are sampled from a multivariate normal distribution whose parameters are updated using
the parents from the previous generation. 
Variation is captured by updating the mean value of the distribution and adding a zero-mean stochastic perturbation.   
CMA-ES adapts the covariance matrix of the distribution by learning a second order model of the underlying fitness function. 
%
CMA-ES requires only the ranking between candidate solutions in order to learn the sample distribution: 
neither derivatives nor the function values are needed.
%

We present the essential steps of the CMA-ES algorithm following \cite{Hansen:2005}. 
Let us fix the size of the population $L\ge 2$ for each generation and a parameter $K < L$.
The parameter $K$ determines the size of the ``most fit'' sub-population that will be used to update
the distribution in the next generation.
In each generation $g = 0, 1,\ldots$, the algorithm maintains an $n$-dimensional multivariate normal distribution 
$(\mu^g, C^g, \gamma^g)$ with mean $\mu^g$, covariance matrix $C^g$, and step size $\gamma^g$. 
Here, $n$ is the dimension of the solution.

To go from generation $g$ to generation $g+1$, we first
sample a population of $L$ new individuals using the multivariate distribution from the previous generation:
\begin{equation}
\label{eq:cma_basic}
	x^{(g+1)}_i = \mu^{(g)}+\gamma^{(g)}z^{(g+1)}_i,\quad z^{(g+1)}_i \sim\mathcal N(0,C^{(g)}),
\end{equation}
for $i=1,\ldots,L$.
%
Based on this population, we update the parameters $(\mu, C, \gamma)$ of the distribution for the $(g+1)$st generation.

The individuals $x^{(g+1)}_i$, $i\in\{1,\ldots,L\}$ are first ranked according to some fitness function. 
Let $\{i_1,\ldots,i_L\}$ be a permutation of $\{1,\ldots,L\}$ such that
$x^{(g+1)}_{i_1},x^{(g+1)}_{i_2},\ldots,x^{(g+1)}_{i_L}$ is a ranked version of the individuals 
in the $(g+1)^{st}$ generation from the best to the worst fitness. 

Then the mean value is updated according to
\begin{equation}
\label{eq:cma_mean}
\mu^{(g+1)}=\mu^{(g)}+c_{\mathsf m}\sum_{s=1}^{K}\omega_s \left[x_{i_s}^{(g+1)}-\mu^{(g)}\right],
\end{equation}
where
$c_{\mathsf m}\le 1$ is the learning rate, usually set to $1$, and
$\boldsymbol{\omega}=[\omega_1,\omega_2,\ldots, \omega_L]$ is a 
weight vector with the property that 
$\sum_{s=1}^{K}\omega_s=1$,  $\omega_1\geq \omega_2\geq \ldots \geq \omega_{K}>0$,
and $\omega_s\le 0$ for $s>K$.

The covariance matrix is updated as follows:
\begin{align}
	\label{cma_covariance}
	C^{(g+1)}& =\left[1-c_1-c_2\sum_{i=1}^{L}\omega_i\right]C^{(g)}\nonumber\\
	&+c_1P_c^{(g+1)}P_c^{{(g+1)}^T} + c_2\sum_{i=1}^L\omega_i z_i^{(g+1)}{z_i^{(g+1)}}^T,
\end{align}
where
$c_1$ and $c_2$ are learning rates corresponding to the rank one and the rank $K$ covariance updates.
The vector $P_c^{(g)}\in \mathbb{R}^n$ denotes the \emph{evolution path} in the $g^{\text{th}}$ generation and is 
updated according to the following adaptation law:
\begin{equation}	
	\label{eq:cma_path_cov}
	P_c^{(g+1)}=(1-c_3)P_c^{(g)}+
	\sqrt{c_3(2-c_3)\kappa}\;\frac{\mu^{(g+1)}-\mu^{(g)}}{\gamma^{(g)}},
\end{equation}
where, $c_3\leq 1$ is a learning rate for path evolution update and $\kappa:=1/\sum_{s=1}^{K}\omega_s^2$. 

Finally, the step size is updated as follows:
\begin{equation}
\label{eq:cma_stepsize}
	\gamma^{(g+1)}=\gamma^{(g)}\exp\left[\frac{c_{\gamma}}{d_{\gamma}}\left(\frac{||P_{\gamma}^{(g+1)}||}{E}-1\right)\right],
\end{equation}
where $c_{\gamma}<1$ and $d_{\gamma}$ denote respectively the learning rate and damping factor for the step size update.
We define the constant $E:=\sqrt{2}\Gamma(\frac{n+1}{2})/\Gamma(\frac{n}{2})$, where $\Gamma$ is the gamma function.
Finally,
$P_{\gamma}^{(g)}$ is the \emph{conjugate evolution path} with the adaptation law
\begin{align}	
\label{eq:cma_path_step}
P_{\gamma}^{(g+1)}& =(1-c_{\gamma})P_{\gamma}^{(g)}\nonumber\\
&+\sqrt{c_{\gamma}(2-c_{\gamma})\kappa}\;\left[C^{(g)}\right]^{-\frac{1}{2}} \frac{\mu^{(g+1)}-\mu^{(g)}}{\gamma^{(g)}},
\end{align}
where $C^{(g)^{-\frac{1}{2}}}$ is the square root of the covariance matrix defined as 
$C^{(g)^{-\frac{1}{2}}}:=BD^{-1}B^T$ with $C^{(g)}=BD^2 B^T$ being an eigen-decomposition of $C^{(g)}$.

We emphasize that the updates to the parameters \emph{only depend on the top $K$ individuals and not on their relative order}.
While a linear ranking function is one way to determine the top $K$ individuals, the CMA-ES algorithm
only requires determining the top $K$ individuals in each generation.

\subsection{Fitness Functions}
 
We now provide fitness functions in our setting.

Suppose a DNN $\mathfrak N(\boldsymbol{\theta}_{\ell},\cdot)$ and a reference input $x_{\textsf{ref}}$ are given.
When using $L_p$ norm, we rank the randomly generated inputs in the CMA-ES algorithm using the fitness function
\begin{equation}
\label{eq:CMA_obj}
f(x):=
\begin{cases}
\|x-x_{\textsf{ref}}\|_p & \text{if }\,\,\mathfrak N(\boldsymbol{\theta}_{\ell},x)\ne \mathfrak N(\boldsymbol{\theta}_{\ell},x_{\textsf{ref}})\\
M_1+M_2\|x-x_{\textsf{ref}}\|_p &  \text{if }\,\,\mathfrak N(\boldsymbol{\theta}_{\ell},x)= \mathfrak N(\boldsymbol{\theta}_{\ell},x_{\textsf{ref}}),
\end{cases}
\end{equation}
where $M_1,M_2>1$ are very large constants. Then $x_i^{(g+1)}$, $i=1,2,\ldots,L$ are ranked such that
 $$f(x_{i_1}^{(g+1)})\le f(x_{i_2}^{(g+1)})\le\ldots\le f(x_{i_L}^{(g+1)}).$$
Note that $f(\cdot)$ assigns very large values to inputs within the same class as $x_{\textsf{ref}}$ 
and reduces the chance of having them as parents for the next generation. 
Note that the fitness function \eqref{eq:CMA_obj} is not continuous.

When using human perception, the algorithm generates a population of new inputs
and filters any image in the same class as the original input
$x_{\textsf{ref}}$.
The filtered sub-population is displayed to the user, who is asked to pick the top $K$
rinputs from the sub-population perceptually similar to $x_{\textsf{ref}}$. 
Fig.~\ref{fig:CMAES_Process} shows a schematic of our implementation.


 
 
\subsection{Improving Performance of CMA-ES Using Bisection}

To find adversarial inputs, we augment the basic CMA-ES search using a bisection technique.
Briefly, we run CMA-ES for a number of iterations to find a misclassified input and then
search for another misclassified input closer to the original input by iteratively
reducing certain dimensions and checking if the resulting input is still misclassified.

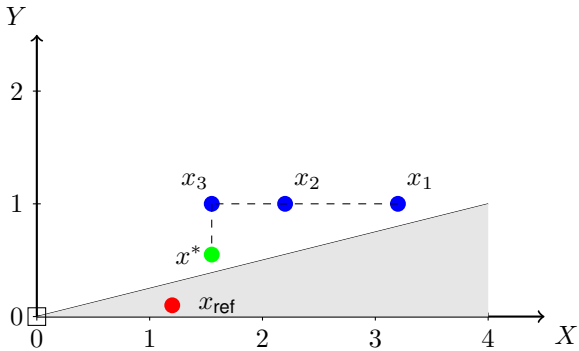
\begin{figure}[t] 
 \begin{tikzpicture}[scale=1.5]
 \tikzset{triangle/.append style={fill=green!100, regular polygon, regular polygon sides=3},
 	anytriangle/.append style={draw, append after command={  }}}
\draw[thick,->] (0,0) -- (4.5,0) node[anchor=north west] {$X$};
\draw[thick,->] (0,0) -- (0,2.5) node[anchor=south east] {$Y$};
\draw[] (0,0) -- (4,1) node[] {};
\foreach \x in {0,1,2,3,4}
\draw (\x cm,1pt) -- (\x cm,-1pt) node[anchor=north] {$\x$};
\foreach \y in {0,1,2}
\draw (1pt,\y cm) -- (-1pt,\y cm) node[anchor=east] {$\y$};

 \coordinate (c1) at (1.2,.1);

 \coordinate (c2) at (3.2,1);
 \draw node[] at (3.4,1.2) {$x_1$};
 \coordinate (c3) at (2.2,1);
  \draw node[] at (2.4,1.2) {$x_2$};
 \coordinate (c4) at (1.55,1);
  \draw node[] at (1.4,1.2) {$x_3$};
 \coordinate (c5) at (1.55,.55);
 \draw node[] at (1.35,.55) {$x^*$};

 \fill[red] (c1) circle (2pt);
 \fill[blue] (c2) circle (2pt);
 \fill[blue] (c3) circle (2pt);
 \fill[blue] (c4) circle (2pt);
 \fill[green] (c5) circle (2pt);
 \draw [dashed] (c2) -- (c4);
 \draw [dashed] (c4) -- (1.55,.6);
 
 \coordinate (A) at (0,0);
 \coordinate (B) at (4,0);
 \coordinate (C) at (0,1);
 
 \coordinate (r0) at (0,0);
 \coordinate (r1) at (4,0);
 \coordinate (r2) at (4,1);

 \fill[ fill=gray!20] (r0) -- (r1) -- (r2) -- cycle;
 \fill[red] (c1) circle (2pt);
 \draw node[] at (1.6,.1) {$x_{\textsf{ref}}$};
 \node [anytriangle] {};
 \end{tikzpicture}
 \caption{Graphical demonstration for bisection method; $x_{\textsf{ref}}$ is the two-dimensional input; $x_1$ is the example generated by CMA-ES after a few iterations; $x_2$, $x_3$ and $x^*$ are computed by running bisection.}
 \label{fig:bisection}
\end{figure}

Fig.~\ref{fig:bisection} summarizes the technique when there are two labels, grey and white,
separated by a linear boundary.
Suppose we start from an input $x_{\textsf{ref}}$ (shown as a red dot) and run CMA-ES to get a misclassified
example $x_1$.
Then, we consider the dimension $i$ of $x_1$ for which $|x_1^{i} - x_{\textsf{ref}}^{i}|$ is the largest.
We consider a new input which replaces the $i^\text{th}$ coordinate of $x_1$ by $(x_1^{i} +x_{\textsf{ref}}^{i})/2$.
In our example, this new point $x_2$ is still misclassified, and we repeat the process with $x_2$ and $x_{\textsf{ref}}$, 
getting subsequently a new point $x_3$, and so on.
In case the new point was not misclassified, we increase the $i^{\text{th}}$ coordinate and continue exploring the same dimension.
We continue in this way to find a new misclassified input $x^*$ closer to $x_{\textsf{ref}}$ for a pre-specified number of steps.
In our experiments, this bisection technique significantly reduced the distance between the original input and 
the misclassified input when $L_\infty$ is used for assessing similarities between images.

\section{Experimental Results: $L_p$ Norms}
\label{sec:simulations}

We have implemented the CMA-ES algorithm to find adversarial examples.
Our experiments had the following aims.
\begin{enumerate}
\item[\textbf{Baseline Performance}] Compare the performance of CMA-ES when using $L_1$ and $L_\infty$ norms with other grey- and
black-box techniques in the literature.
\item[\textbf{Perception-in-the-Loop}] Find adversarial examples using human perception in the loop.
As part of this process, we also evaluate the robustness of perceptual preferences.
\end{enumerate}
In this section, we report results of using CMA-ES in crafting adversarial examples using $L_1$ and $L_\infty$ norms,
and compare its performance with existing algorithms.
The aim of this comparison is to show that black-box CMA-ES is comparable in efficiency to existing black- or grey-box
techniques and to establish a basline performance that is used to determine population size and number of iterations
for perception-in-the-loop experiments.
In Section~\ref{sec:human}, we report on running CMA-ES using perception in the loop.

We have conducted experiments on three popular image recognition benchmarks: MNIST, GTSRB, and CIFAR10. 
We compare the performance of CMA-ES against three main techniques: substitute training algorithm from \cite{Papernot:2017}, 
the GA algorithm from \cite{Mustafa:2018}, 
and the MCTS algorithm from \cite{Marta:2018}. 
Recall that the first two are black-box approaches while the third is grey box. 
We use $L_1$ norm in the fitness function of all these algorithms with a fixed number of queries.
Given two images $x$ and $x'$ with $k$ channels and resolution $h\times w$, 
the \emph{average perturbation} in the $L_1$ norm sense is defined as:
\begin{equation}
\label{eq:avgperturb}
d(x,x')=\frac{||x-x'||_1}{k\times h\times w}.
\end{equation}


\subsection{Benchmarking CMA-ES on MNIST} 
\label{subsec:L1}

The MNIST dataset is considered to be the most common benchmark in many previous studies for crafting adversarial examples.
It consists of $60000$ grey scale images of handwritten digits, with resolution $28\times 28$ for training, as well as $10000$ images for testing.
We split the training dataset into training ($80\%$) and validation ($20\%$) sets.
We trained two DNNs proposed for MNIST, labeled DNN A (\cite{Papernot:2017}) and DNN B\footnote{Using available code at \url{https://mathworks.com}} 
in Table~\ref{tb:DNN_LIST}.
The DNNs are of type LeNet and consist of input, convolutional, max pooling, fully connected, and softmax layers. 
Both DNNs are trained for $10$ epochs with learning rate of $0.001$ and the
achieved accuracies for DNN A and DNN B on the $10000$ test images are $98.36\%$ and $97.73\%$ respectively, 
which is close to the accuracy of state-of-the-art classifiers on MNIST \cite{Marta:2018}.

We conducted our experiments in the following way.
For each class, we generated $10$ randomly selected images from test dataset, which are classified correctly by the DNN.
For each image, first, CMA-ES was run for $180$ iterations, with population size $4$ ($L=4$) and parent size $2$ ($K=2$).
This took about $15$ seconds in average. 
In order to compare our result with the black-box algorithm proposed in \cite{Papernot:2017}, 
we used their own implementation for MNIST dataset: 
$150$ samples from test data were used in input data augmentation; 
for training of substitute network, the DNN's output label was read 6400 times; 
the architecture of the substitute NN includes input and output layers as well as two fully connected layers with $200$ neurons each. 

For the genetic algorithm based crafting method \cite{Mustafa:2018}, we achieved the best performance by running GA for $120$ generations 
including $6$ individuals, and setting the mutation rate to be $0.05$, as given in \cite{Mustafa:2018}. 
For GA implementation, we used the same fitness function as CMA-ES. 
Notice that the query length in both GA and CMA-ES is set to $720$. 

Finally, we took one of the most powerful grey-box adversarial example crafting methods, proposed in \cite{Marta:2018} into account. 
Their algorithm is  based on Monte Carlo Tree Search (MCTS). 
Apart from GA, for the rest of methods, we ran authors' provided codes for both DNN A and B setting all the parameters to their default values.
Tables~\ref{tb:CMA-ES_Goodfellow_A} and \ref{tb:CMA-ES_Goodfellow_B} list results of running all four methods mentioned above. 
For GA and CMA-ES, the success rate is $100\%$, meaning that for $100\%$ of tested inputs, an adversarial example is found. 
However, success rate for substitute training and MCTS methods is $89\%$. 

We measure the performance in terms of average perturbation, using \eqref{eq:avgperturb}.
It can be noticed that apart from MCTS, CMA-ES's performance is significantly better, but the
performance of CMA-ES and MCTS is close.  
However, CMA-ES is superior in terms of success rate.

\begin{figure}[t]
	\begin{center}
	\begin{tabular}{cccc}
		\includegraphics[scale=.05]{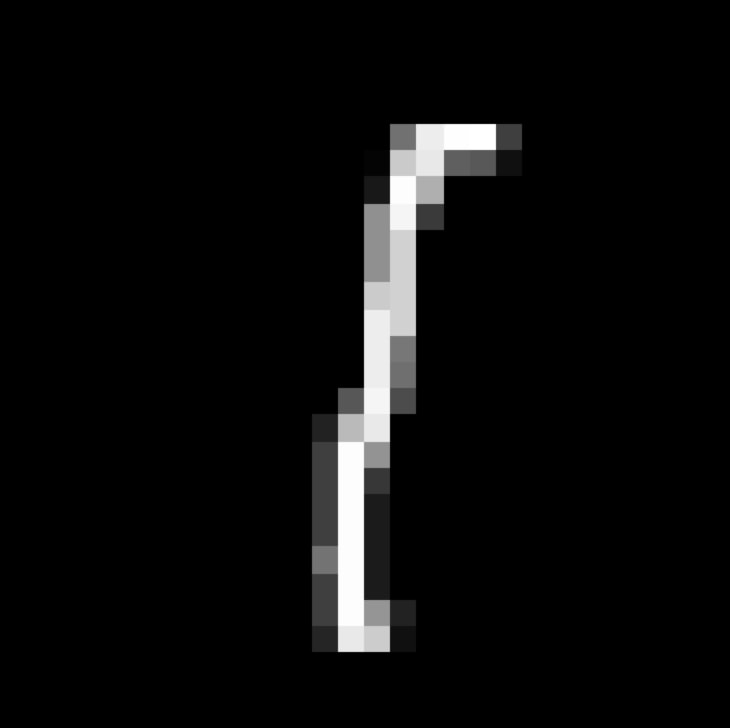}&\includegraphics[scale=.05]{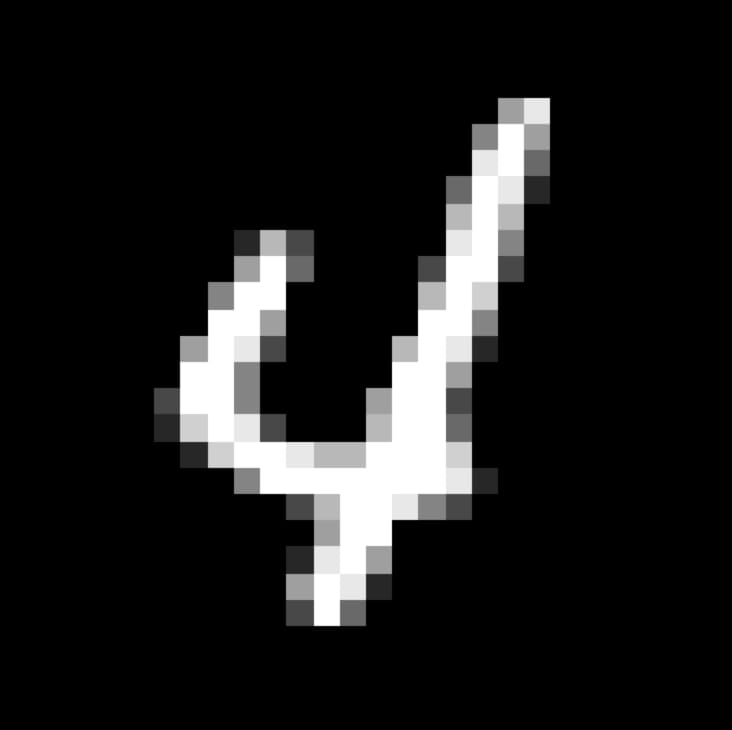}&\includegraphics[scale=.05]{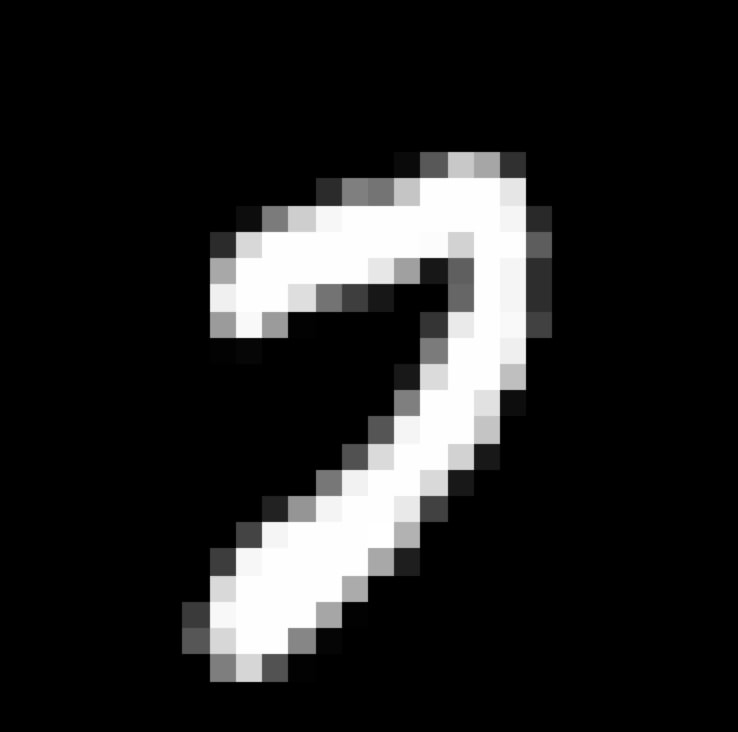}&\includegraphics[scale=.05]{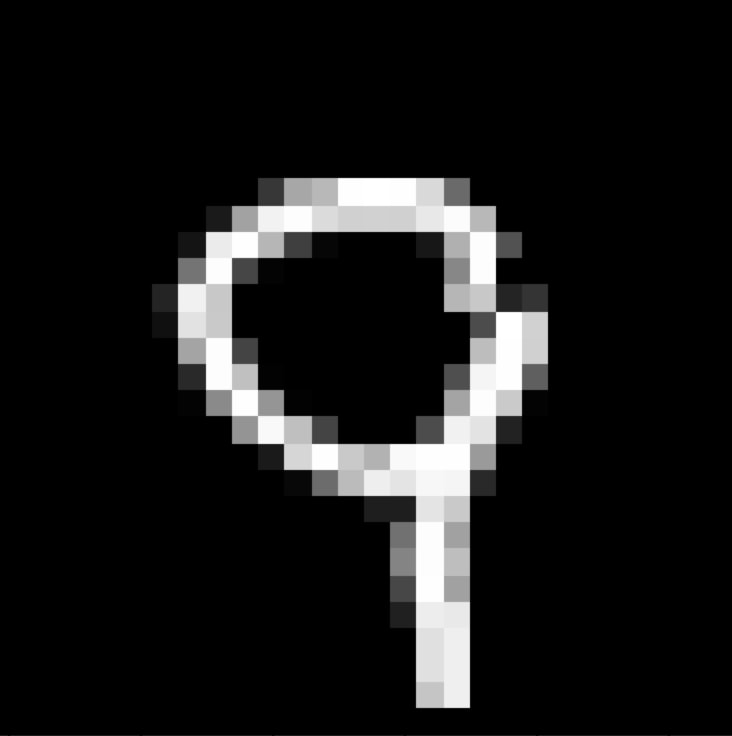}\\
		(1) & (4) & (7) & (9)\\
		\\
		\includegraphics[scale=.05]{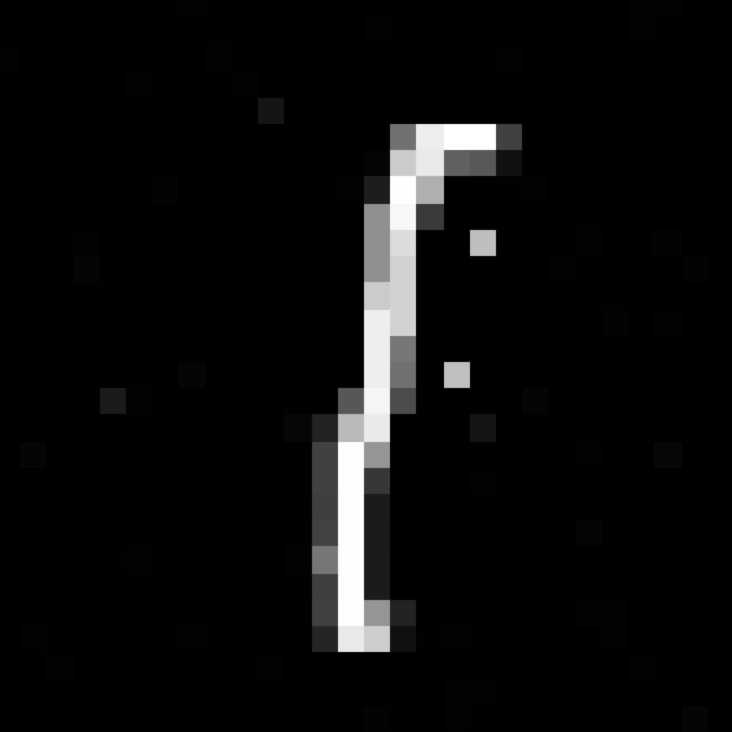}&\includegraphics[scale=.05]{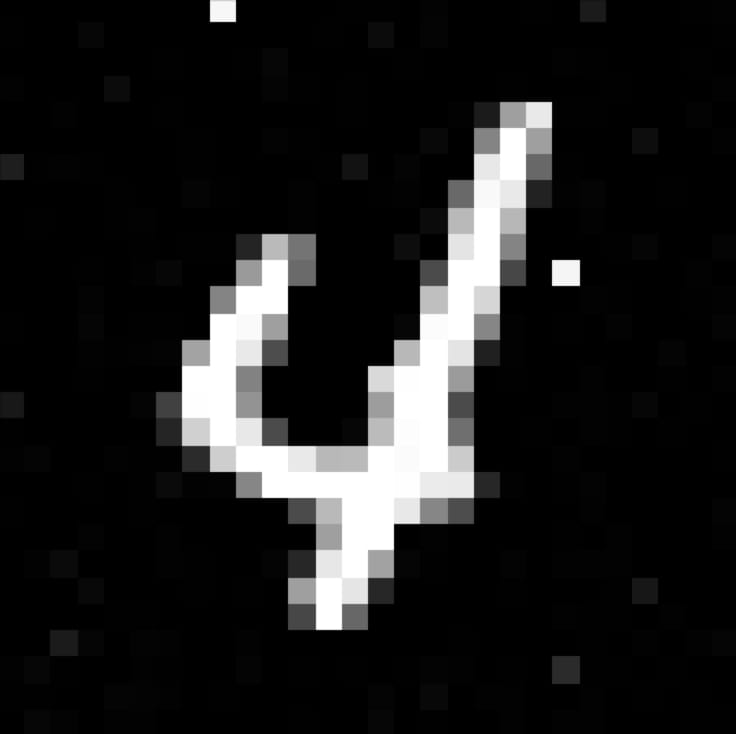}&\includegraphics[scale=.05]{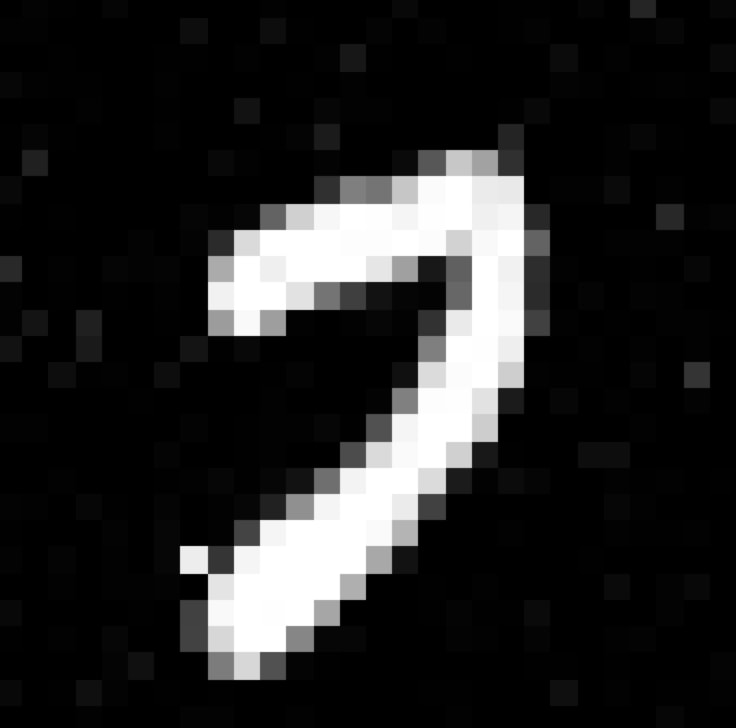}&\includegraphics[scale=.05]{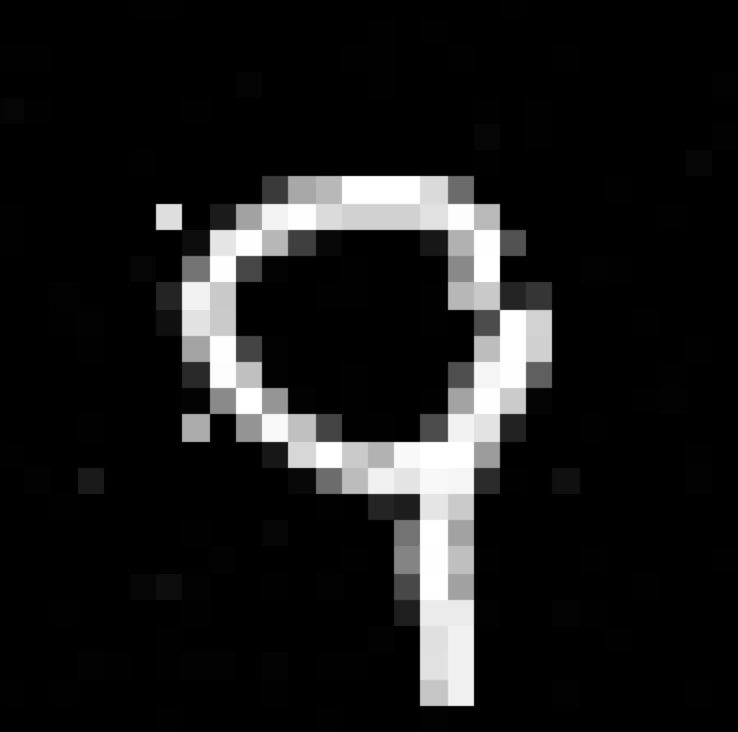}\\
		(8) & (6) & (3) & (7)\\
	\end{tabular}
	\end{center}
	\caption{Top: Original samples from MNIST dataset; Bottom: Misclassified examples generated by CMA-ES for inputs from top row}
	\label{fig:MNIST_ex}
\end{figure}

In all experiments, CMA-ES always found adversarial examples with less than $7\%$ perturbation in the $L_1$ norm sense. 
As a concrete example, Fig.~\ref{fig:MNIST_ex} illustrates the case for which the four input images shown on the top row are classified by the DNN correctly; 
however, CMA-ES is able to find within 15s the corresponding perturbed images, shown on the bottom row, 
which are misclassified by the DNN. 
Increasing the number of iterations for CMA-ES while fixing other parameters results in even smaller perturbations.

\begin{table*}[t]
\caption{Description of the DNN architectures used in this paper. ID: reference mentioned in the paper; Input: input dimension; Conv($a\times b$): convolutional layer with $a\times b$ kernels followed by max-pooling with $2\times 2$ kernel and ReLu as activation function; FC: fully connected layer; S: softmax layer.}
\label{tb:DNN_LIST}
	\begin{center}
		\begin{tabular}{|c|c|c|c|c|c|c|c|c|c|}
			ID & Input&Conv& Conv &  Conv & Conv & Conv & Conv & FC & S \\\hline
			A & 784& 32 ($2\times 2$)&  64 ($2\times 2$)& - & - & - & - & -& 10\\\hline
			B & 784 & 20 ($5\times 5$) & - & -&-& - & - & -& 10  \\\hline
			C & 784& 16 ($4\times 4$)&  32 ($4\times 4$)& - & - & - & - & 100& 10\\\hline
			D & 3072 & 32 ($3\times 3$)&  32 ($3\times 3$)& 64 ($3\times 3$) & 64 ($3\times 3$) & - & - & 512&10 \\ \hline
			E & 4800& 32 ($3\times 3$)& 32 ($3\times 3$) & 64 ($3\times 3$) &64 ($3\times 3$)& 128 ($3\times 3$) & 128 ($3\times 3$) & 512 &43 \\ \hline
		\end{tabular}
	\end{center}
\end{table*}

\begin{table*}[t]
	\caption{Comparing CMA-ES with other blackbox methods for crafting adversarial examples on MNIST data (DNN A)}
	\label{tb:CMA-ES_Goodfellow_A}
	\begin{center}
		\begin{tabular}{|c|c|c|c|c|c|c|c|c|c|c|}
			Method & 0& 1 &  2 & 3 & 4 & 5 & 6 & 7&8 &9 \\\hline
			CMA-ES & 1.80\%& 0.91\% &  2.99\% & 3.66\% & 3.96\% & 6.59\% & 4.51\% & 2.30\%&3.51\% &1.04\% \\\hline
			Substitute Training \cite{Papernot:2017} & $32\%$ & $38.33\%$ & $57\%$ & $47.5\%$ &$42.66\%$& $67.33\%$ & $41.33\%$ & $39.66\%$ &$40.37\%$& $37.33\%$ \\\hline
			GA   & $15.69\%$ & $12.81\%$ & $18.52\%$ &$19.22\%$& $18.05\%$ & $19.14\%$ & $18.8\%$ &$15.56\%$& $16.31\%$ & $15.12\%$ \\ \hline
			MCTS \cite{Marta:2018} & $2.44\%$ & $2.18\%$ & $1.97\%$ &$2.09\%$& $2.1\%$ & $1.96\%$ & $2.81\%$ &$2.11\%$& $2.45\%$ & $1.73\%$ \\ \hline
		\end{tabular}
	\end{center}
\end{table*}

\begin{table*}[t]
	\caption{Comparing CMA-ES with other blackbox methods for crafting adversarial examples on MNIST data (DNN B)}
	\label{tb:CMA-ES_Goodfellow_B}
	\begin{center}
		\begin{tabular}{|c|c|c|c|c|c|c|c|c|c|c|}
			Method & 0& 1 &  2 & 3 & 4 & 5 & 6 & 7&8 &9 \\\hline
			CMA-ES & 0.75\%& 0.93\% &  0.33\% & 1.36\% & 1.94\% & 2.27\% & 3.9\% & 0.9\%&2.9\% &0.72\% \\\hline
			Substitute Training \cite{Papernot:2017} & $22\%$ & $25.18\%$ & $51.33\%$ &$37.61\%$& $27.66\%$ & $37.03\%$ & $45.71\%$ &$23.33\%$& $37.66\%$ & $38\%$\\\hline
			GA   & $10.07\%$ & $10.28\%$ & $15.23\%$ &$14.27\%$& $11.92\%$ & $14.55\%$ & $14.29\%$ &$9.81\%$& $11.18\%$ & $10.79\%$ \\ \hline
			MCTS \cite{Marta:2018} & $1.54\%$ & $1.61\%$ & $1.95\%$ &$2.11\%$& $1.98\%$ & $1.88\%$ & $1.89\%$ &$1.72\%$& $1.81\%$ & $1.61\%$ \\ \hline
		\end{tabular}
	\end{center}
\end{table*}

\begin{table*}[t]
	\begin{center}
		\caption{Crafting adversarial examples on CIFAR10 data using CMA-ES}\label{tb:CIFAR10_all_pixels}
		\begin{tabular}{|c|c|c|c|c|c|c|c|c|c|c|}
			Source class & plane& car &  bird & cat & deer & dog & frog & horse&ship &truck \\\hline
			Mean pert & 0.67\%&  0.83\% & 0.38\% & 0.49\% & 0.51\% & 0.50\% & 3.98\%&0.71\% & 0.90\%&1.38\% \\\hline
			Success rate & $100\%$ & $100\%$ & $100\%$ &$100\%$& $100\%$ & $100\%$ & $100\%$ &$100\%$& $100\%$ & $100\%$\\\hline
		\end{tabular}
	\end{center}
\end{table*}

%


\subsection{Benchmarking CMA-ES on GTSRB}
\label{subsec:GTSRB}
The German Traffic Sign Recognition Benchmark (GTSRB) includes colored ($3$-channel) images with different sizes. 
After resizing and removing the frames of images, the GTSRB dataset contains 39209 $3$-channel $40\times 40$ pixel images for training, 
as well as 12600 test images. 
We split the training set into training ($80\%$) and validation ($20\%$) sets. 
We used a previously proposed convolutional neural network (DNN E in Table~\ref{tb:DNN_LIST})\footnote{Using available code at \url{https://chsasank.github.io/keras-tutorial.html}}
of type LeNet and consisting of input, convolution ($\times 6$), max pooling ($\times 3$), fully connected ($\times 1$), softmax, and output layers. 
The DNN is trained for 32 epochs with learning rate of $0.0001$. 
The achieved accuracy of the trained DNN on $12600$ test data is $96.53\%$.

\begin{figure}[H]
	\centering
	\begin{tabular}{cccc}
		\includegraphics[scale=.05]{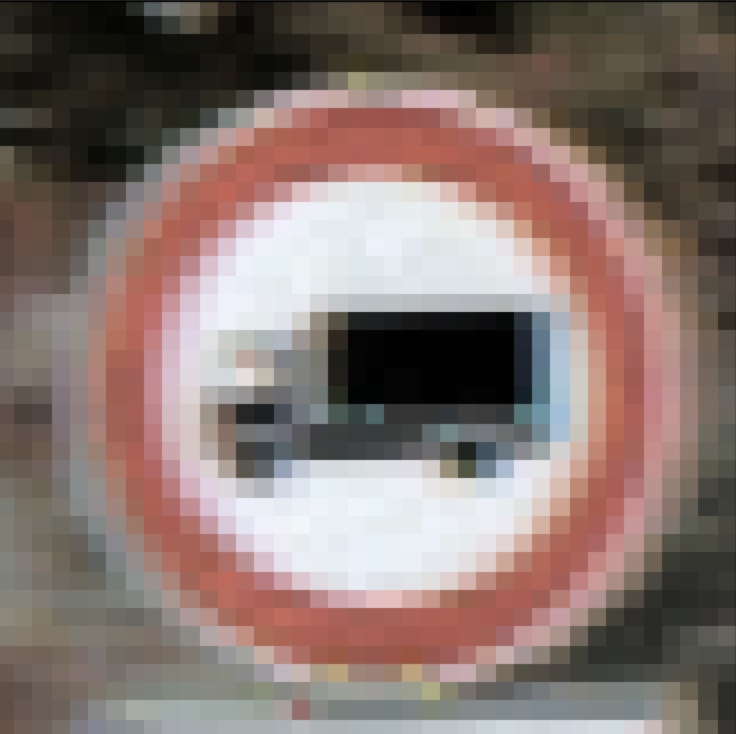}&\includegraphics[scale=.05]{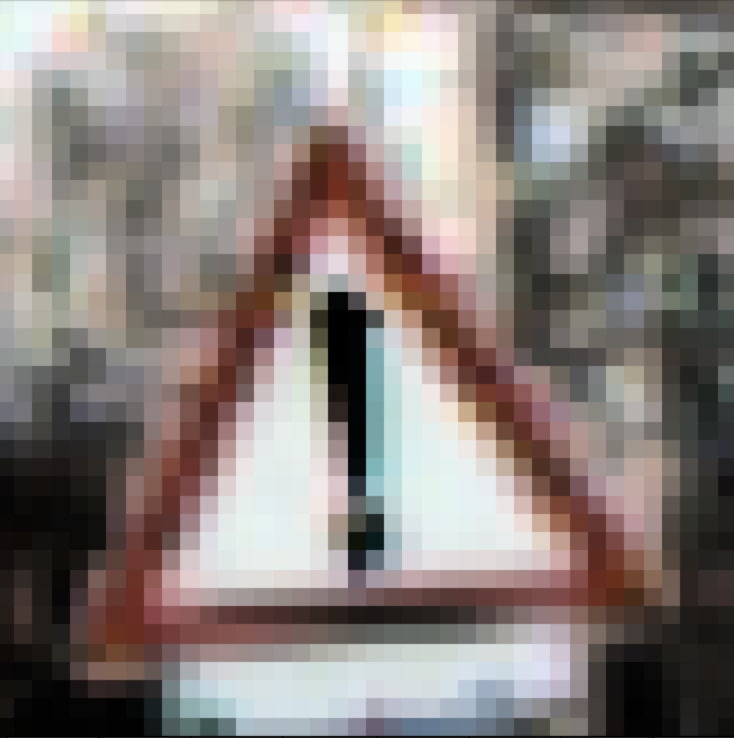}&\includegraphics[scale=.05]{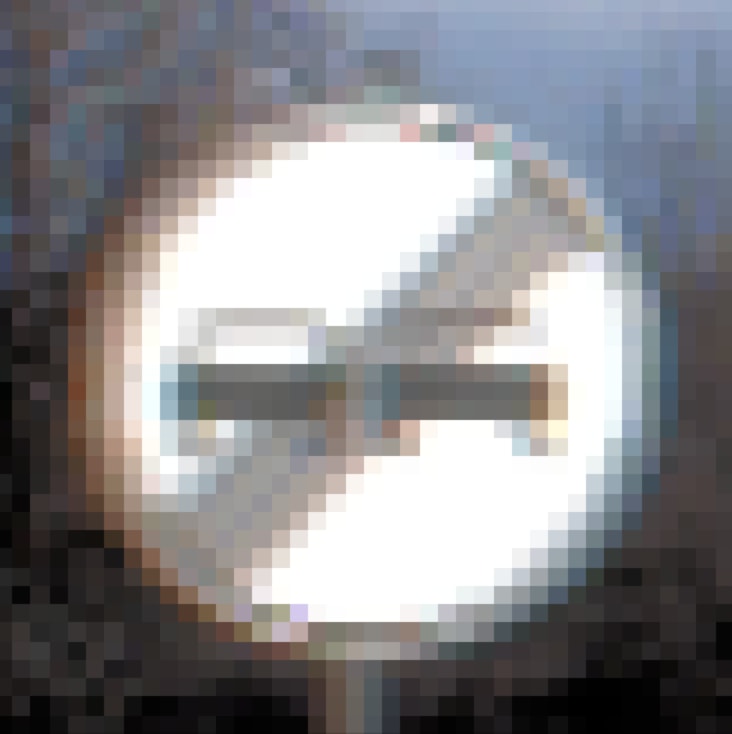}&\includegraphics[scale=.05]{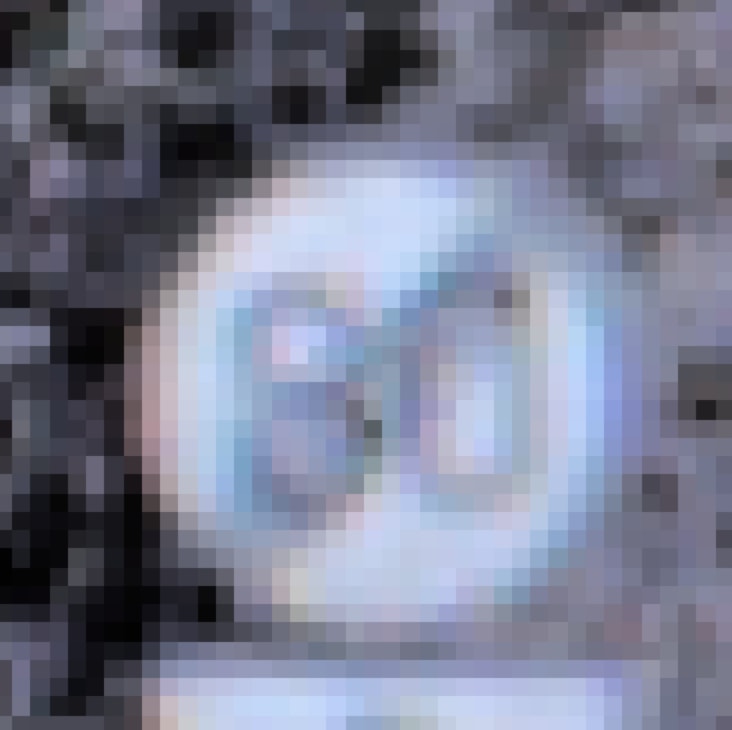}\\
		\includegraphics[scale=1,width=.2cm,height=.2cm]{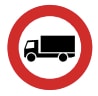}&\includegraphics[scale=1,width=.2cm,height=.2cm]{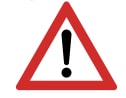}&\includegraphics[scale=1,width=.2cm,height=.2cm]{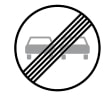}&\includegraphics[scale=1,width=.2cm,height=.2cm]{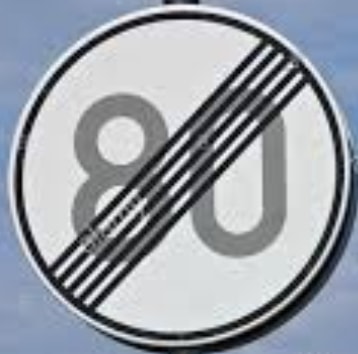}
		\\
		\includegraphics[scale=.05]{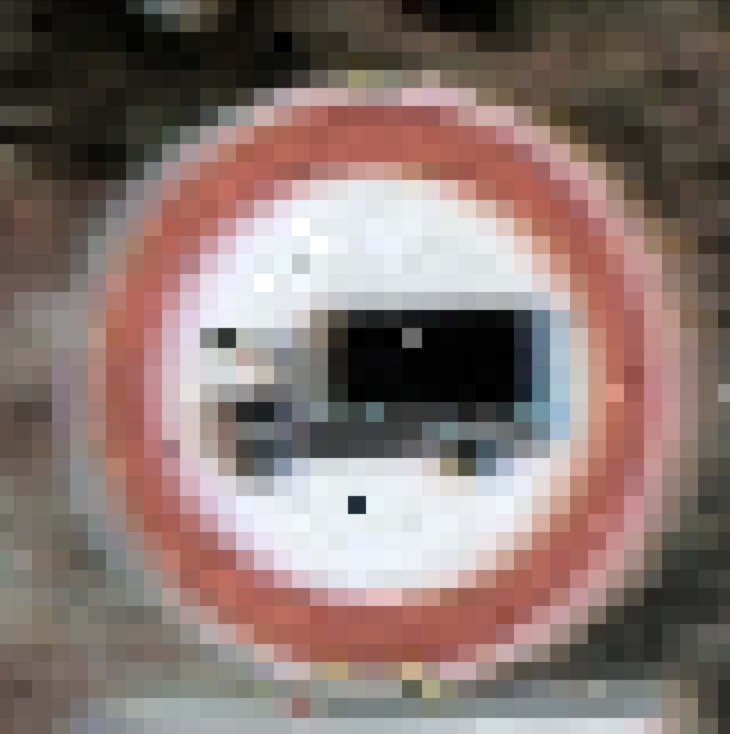}&\includegraphics[scale=.05]{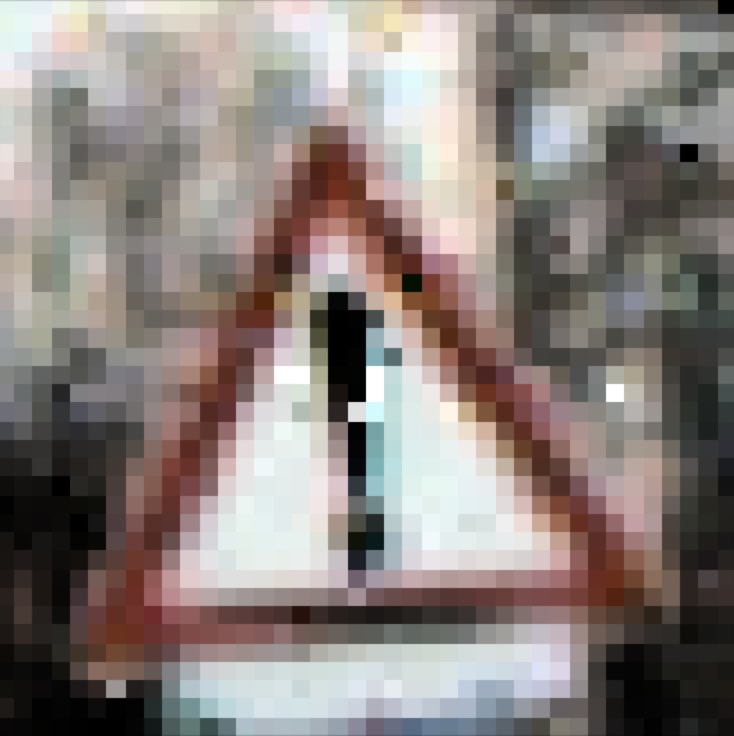}&\includegraphics[scale=.05]{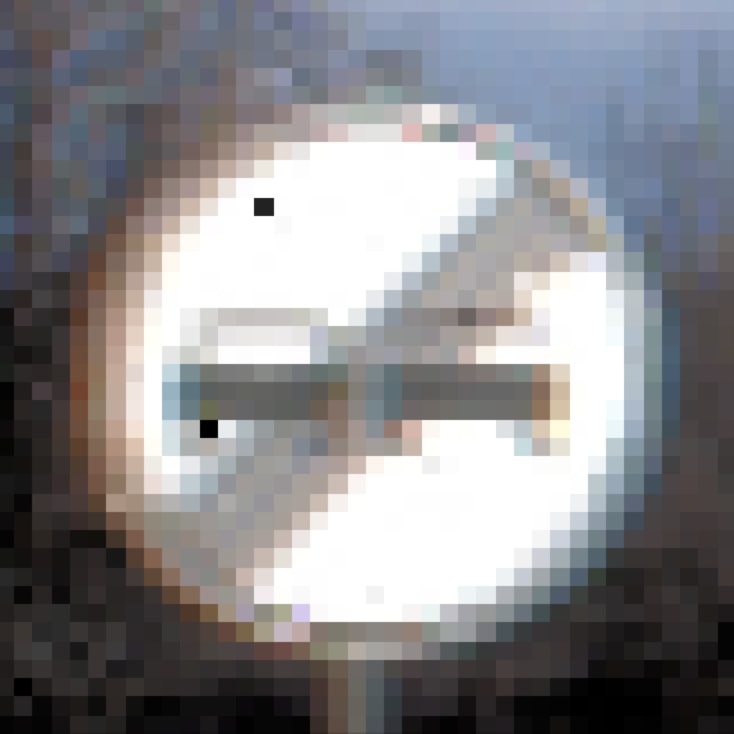}&\includegraphics[scale=.05]{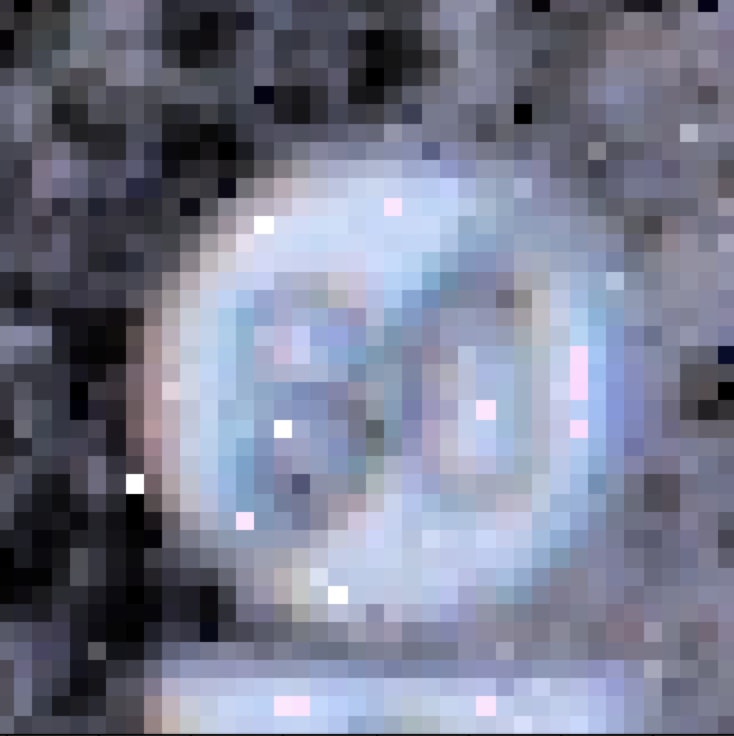}\\
		\includegraphics[scale=1,width=.2cm,height=.2cm]{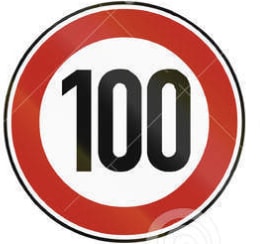}&\includegraphics[scale=1,width=.2cm,height=.2cm]{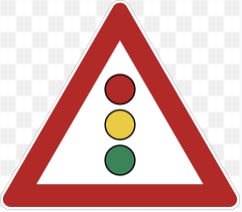}&\includegraphics[scale=1,width=.2cm,height=.2cm]{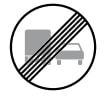}&\includegraphics[scale=1,width=.2cm,height=.2cm]{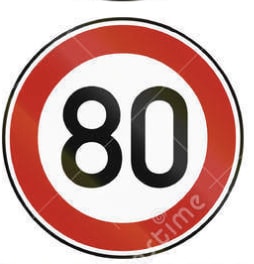}\\
	\end{tabular}
	\caption{Top: Original examples from GTSRB dataset; Bottom: Misclassified examples generated by CMA-ES for inputs from top row using pixel perturbation}
	\label{fig:GTSRB_ex_pixel}
\end{figure}

In order to generate adversarial examples, first, we took all the $1600$ pixels into consideration as optimization variables for CMA-ES. 
Again, for each of the classes, we chose $10$ random samples from test dataset and ran CMA-ES to find $1600$ values as elements of the perturbation vector. 
We set the number of iterations to $180$, population size to $4$ ($L=4$) and parent size to $2$ ($K=2$). 
The overall success rate was $97.3\%$ and average perturbation rate was $4.19\%$. 
Some of the misleading examples crafted by CMA-ES are demonstrated in Fig.~\ref{fig:GTSRB_ex_pixel}. 

\begin{figure}[t]
	\centering
	\begin{tabular}{cc}
		\includegraphics[scale=.1]{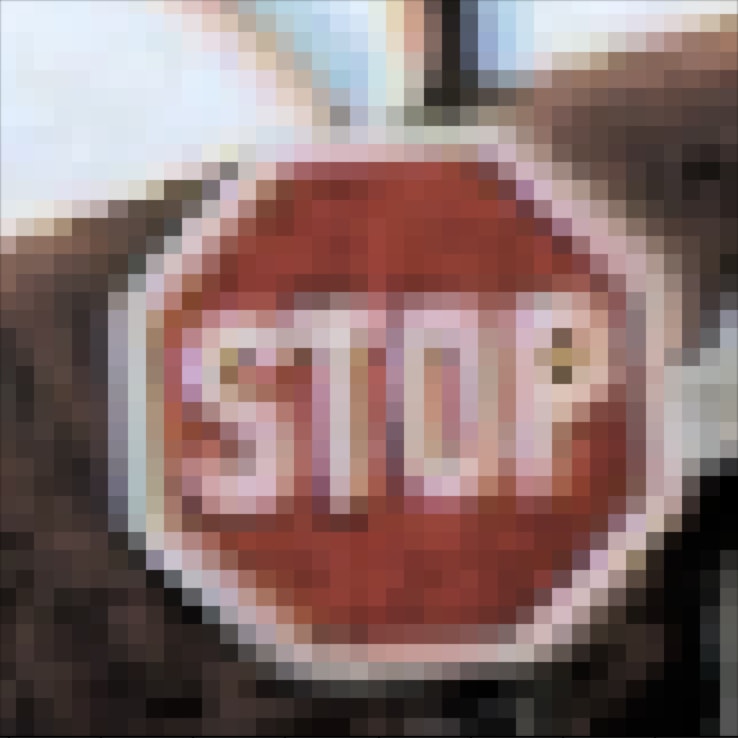}&\includegraphics[scale=.1]{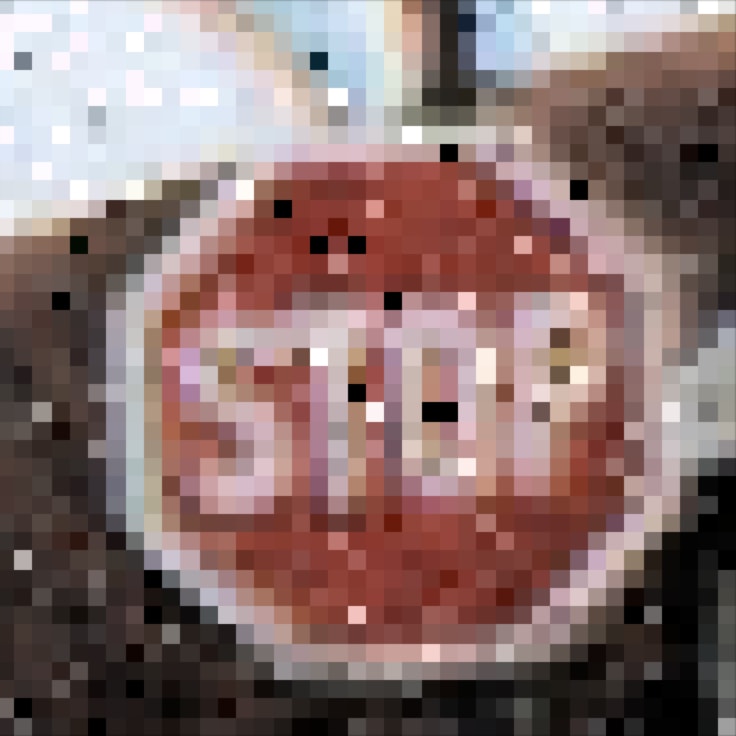}\\
		\includegraphics[scale=.1]{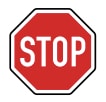}&\includegraphics[scale=.1]{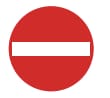}\\
	\end{tabular}
	\caption{Original image labeled by DNN as "STOP" (Left) and perturbed image classified as "No entry" after adding $4\%$ noise (Right)}
	\label{fig:GTSRB_full_ooo}
%
\end{figure}

\begin{figure}[t]
	\centering
	\begin{tabular}{cccc}
		\includegraphics[scale=.05]{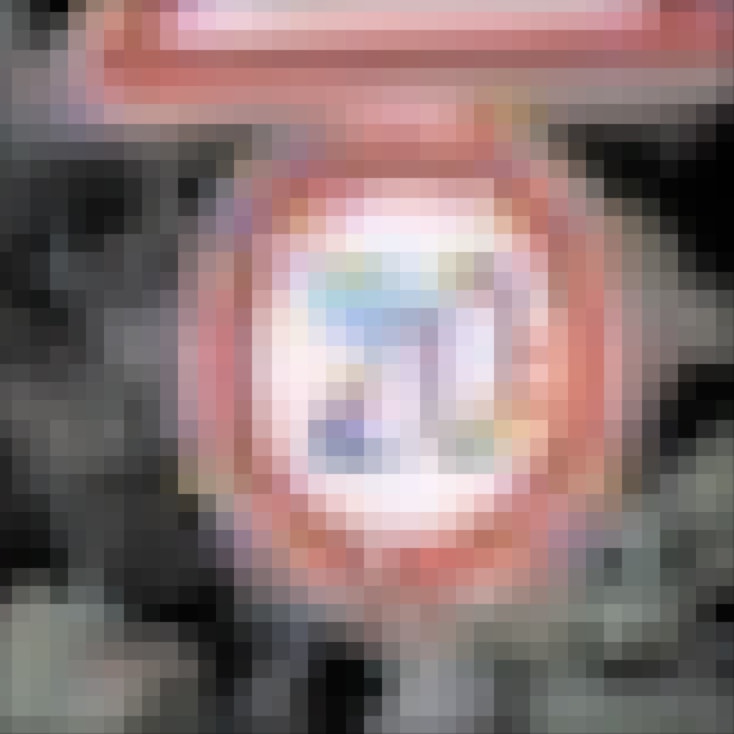}&\includegraphics[scale=.05]{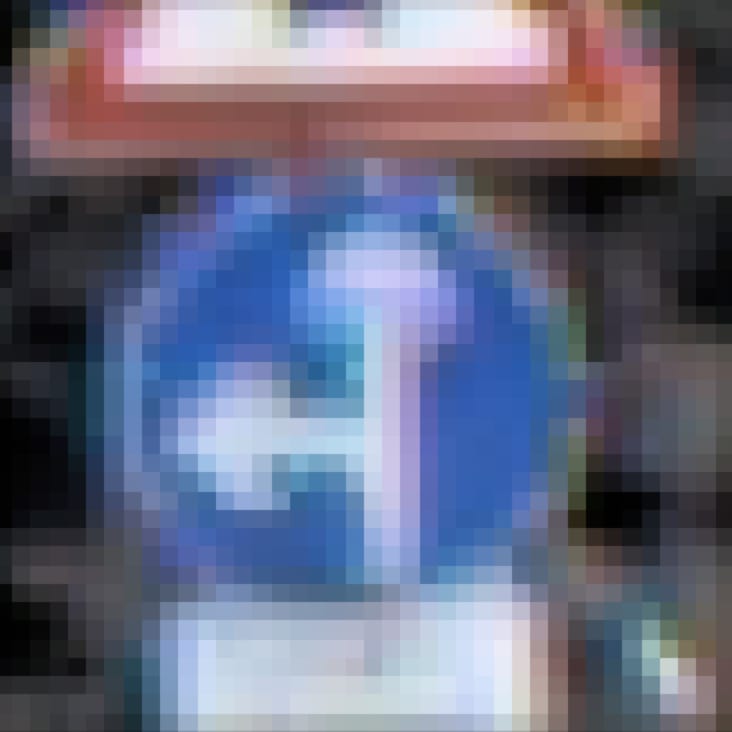}&\includegraphics[scale=.05]{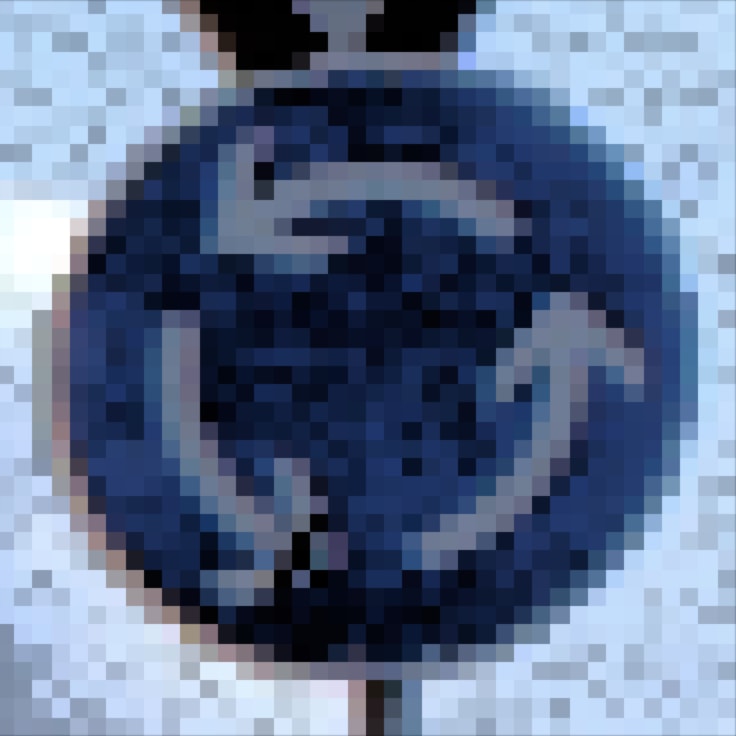}&\includegraphics[scale=.05]{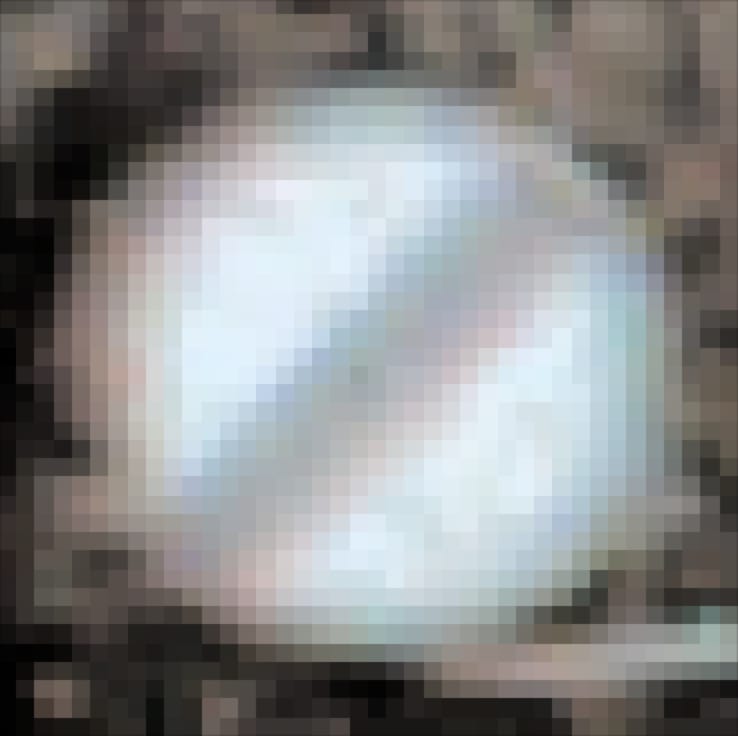}\\
		\includegraphics[scale=1,width=.2cm,height=.2cm]{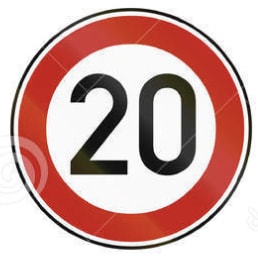}&\includegraphics[scale=1,width=.2cm,height=.2cm]{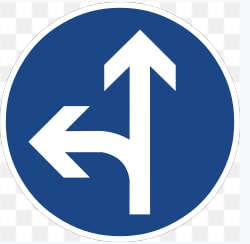}&\includegraphics[scale=1,width=.2cm,height=.2cm]{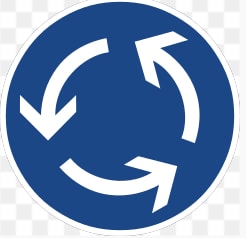}&\includegraphics[scale=1,width=.2cm,height=.2cm]{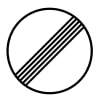}
		\\
		\includegraphics[scale=.05]{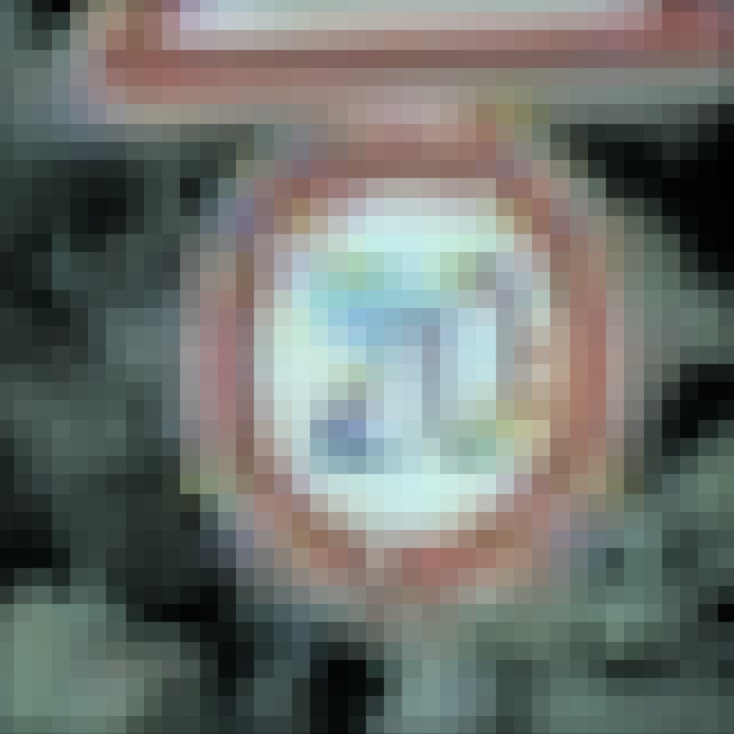}&\includegraphics[scale=.05]{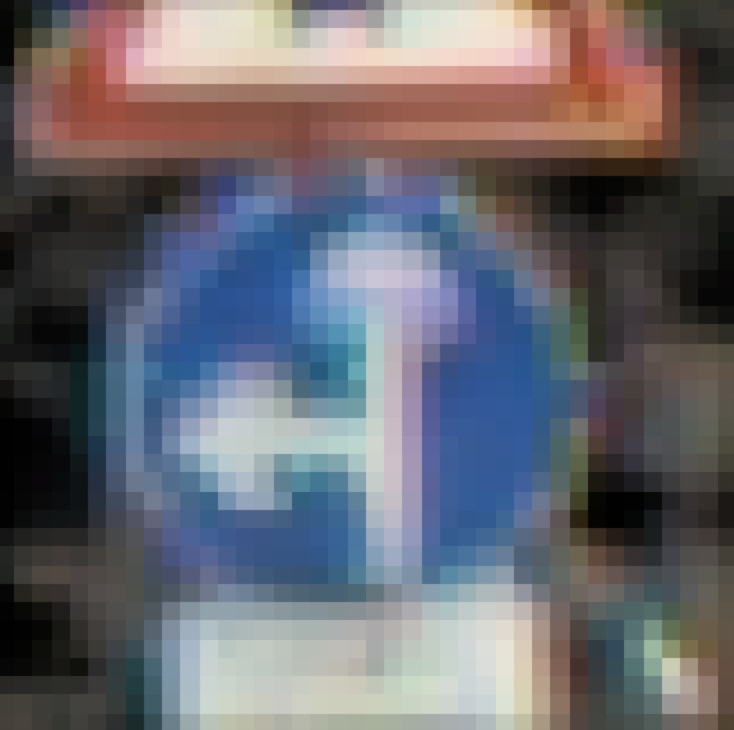}&\includegraphics[scale=.05]{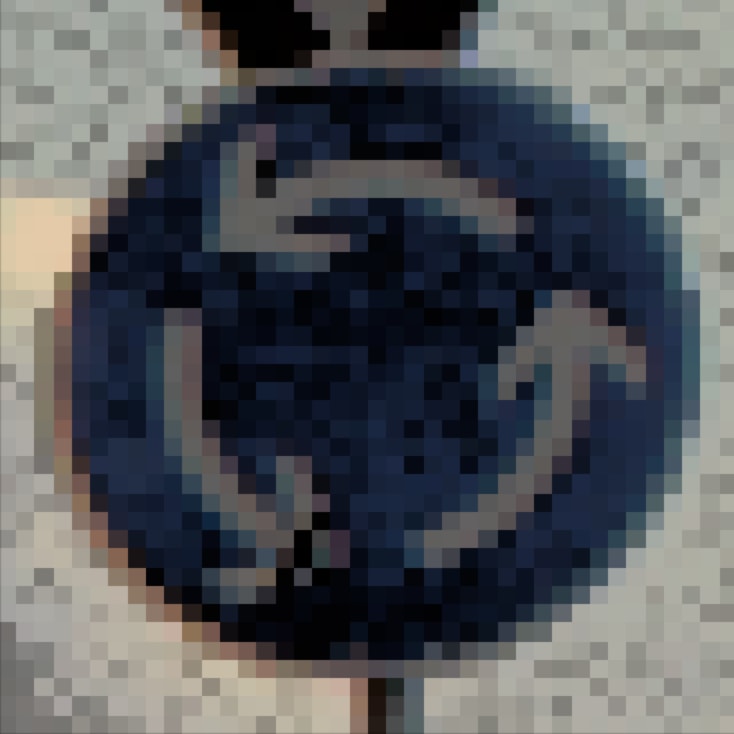}&\includegraphics[scale=.05]{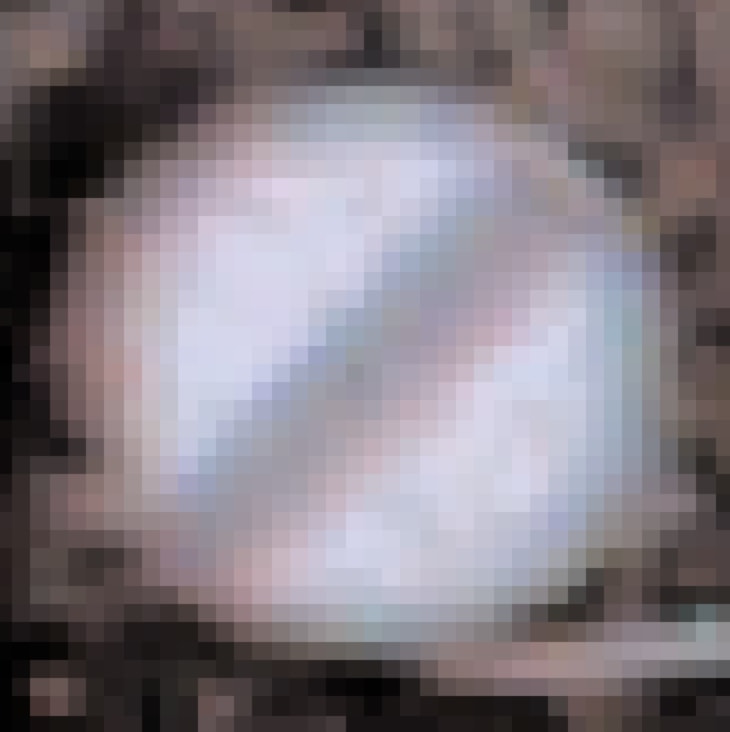}\\
		\includegraphics[scale=1,width=.2cm,height=.2cm]{Figs/40_main}&\includegraphics[scale=1,width=.2cm,height=.2cm]{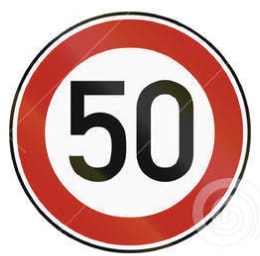}&\includegraphics[scale=1,width=.2cm,height=.2cm]{Figs/7_main}&\includegraphics[scale=1,width=.2cm,height=.2cm]{Figs/6_main}
		\\
	\end{tabular}
	\caption{Top: Original examples from GTSRB dataset; Bottom: Misclassified examples generated by CMA-ES for inputs from top row using color combination change}
	\label{fig:GTSRB_ex_col}
\end{figure}

We noticed, as in \cite{Sharif:2018}, that for RGB photos, there were cases in which small perturbations in the pixel level (in $L_1$ norm sense) 
was not compatible with human perception. 
One such example is shown in Fig.~\ref{fig:GTSRB_full_ooo} in which adding $4.01\%$ perturbation makes the image fuzzy.
%
Thus, in a second experiment, we changed the perturbation criterion to three values $\beta_r,\beta_g,\beta_b$ as darkening factors for red, green 
and blue channels respectively. These darkening factors are in the interval $[0,1]$, being equal to one means no perturbation and being equal to zero means fully darkened.
For this new setting, the objective function Eq.~\eqref{eq:CMA_obj} can be written as
\begin{equation}
\label{eq:CMA_col_obj}
f(x):=
\begin{cases}
3-(\beta_r + \beta_g + \beta_b) & \text{if }\,\,\mathfrak N(\boldsymbol{\theta}_{\ell},x)\ne \mathfrak N(\boldsymbol{\theta}_{\ell},x_{\textsf{ref}})\\
M &  \text{if }\,\,\mathfrak N(\boldsymbol{\theta}_{\ell},x)= \mathfrak N(\boldsymbol{\theta}_{\ell},x_{\textsf{ref}}),
\end{cases}
\end{equation}
where $M>3$ is a large number.
The number of iterations was set to $3$, population size was $240$, and parent size was $120$. 
Running CMA-ES in this setting results in $100\%$ success rate. 
Fig.~\ref{fig:GTSRB_ex_col} demonstrates some examples of adversarial images crafted by changing color combinations.


\subsection{Benchmarking CMA-ES on CIFAR10}
\label{subsec:CIFAR10}
The CIFAR10 dataset contains $60000$ $32\times 32$ color images (three channels) in 10 classes, with $6000$ images per class. 
There are $50000$ training images and $10000$ test images. 
We trained a previously proposed convolutional neural network (DNN D in Table~\ref{tb:DNN_LIST})\footnote{Using available code at \url{https://chsasank.github.io/keras-tutorial}} 
with convolution ($\times 2$), max pooling ($\times 2$), fully connected ($\times 1$), and softmax layers, which achieved $72.3\%$ test accuracy. 

\begin{figure}[t]
	\centering
	\begin{tabular}{cccc}
		\includegraphics[scale=.05]{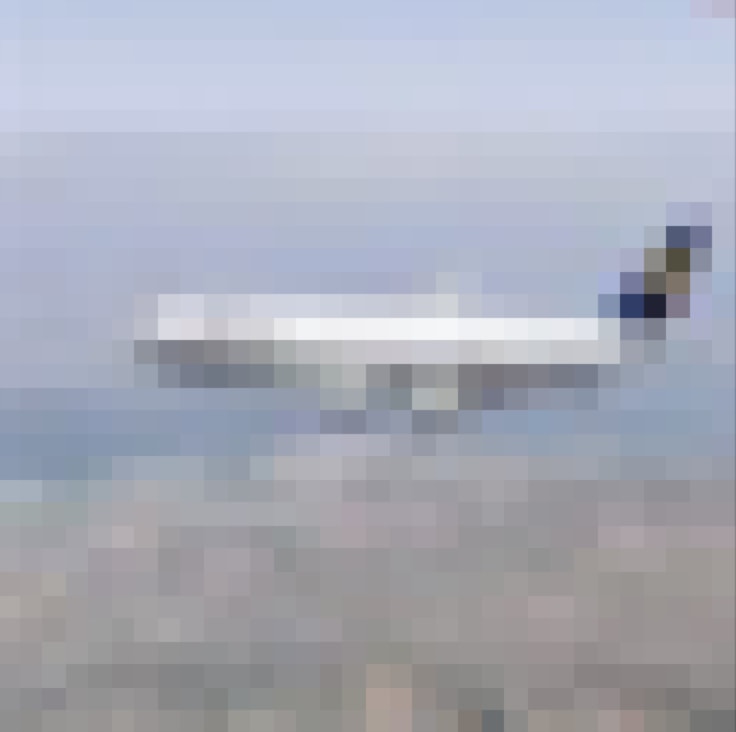}&\includegraphics[scale=.05]{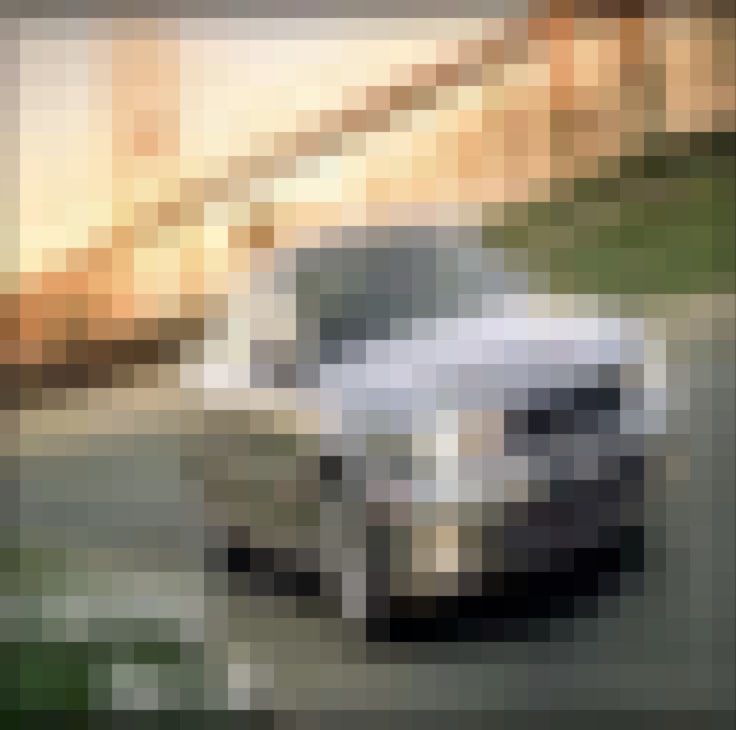}&\includegraphics[scale=.05]{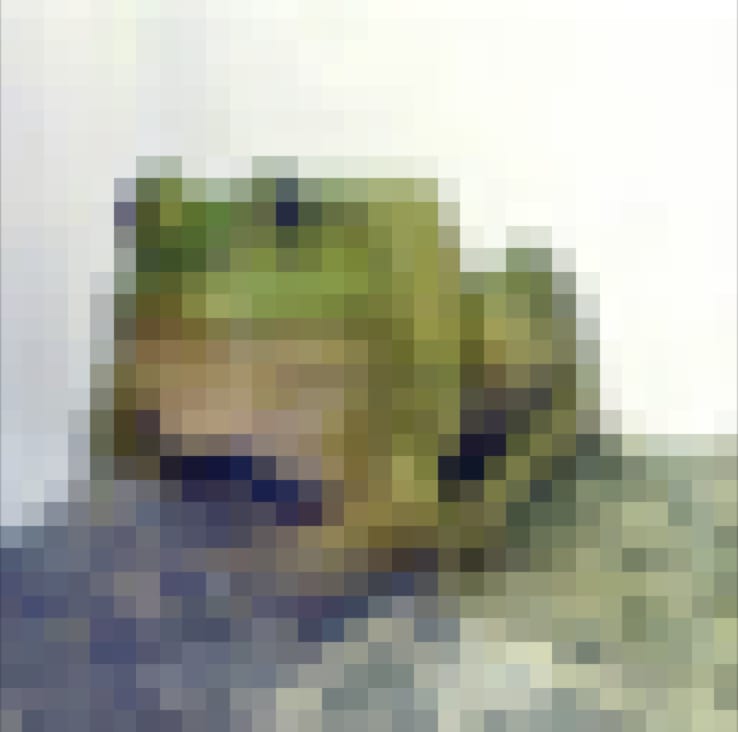}&\includegraphics[scale=.05]{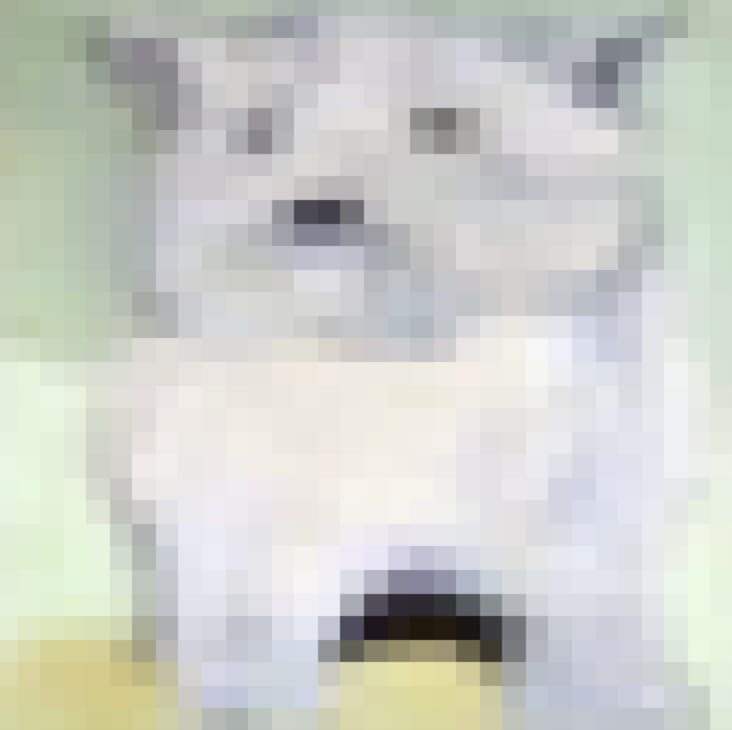}\\
		Plane & Car & Frog & Dog\\
		\\
		\includegraphics[scale=.05]{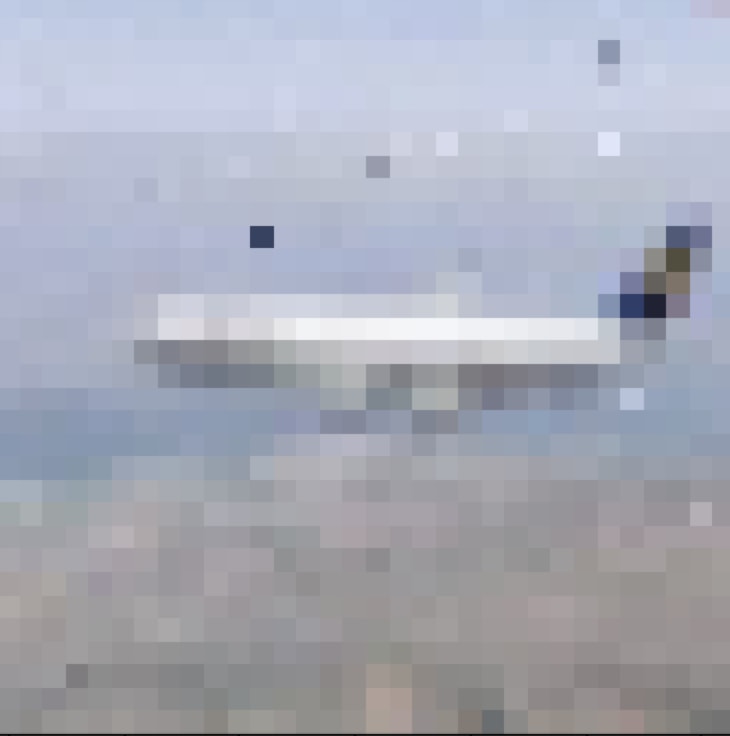}&\includegraphics[scale=.05]{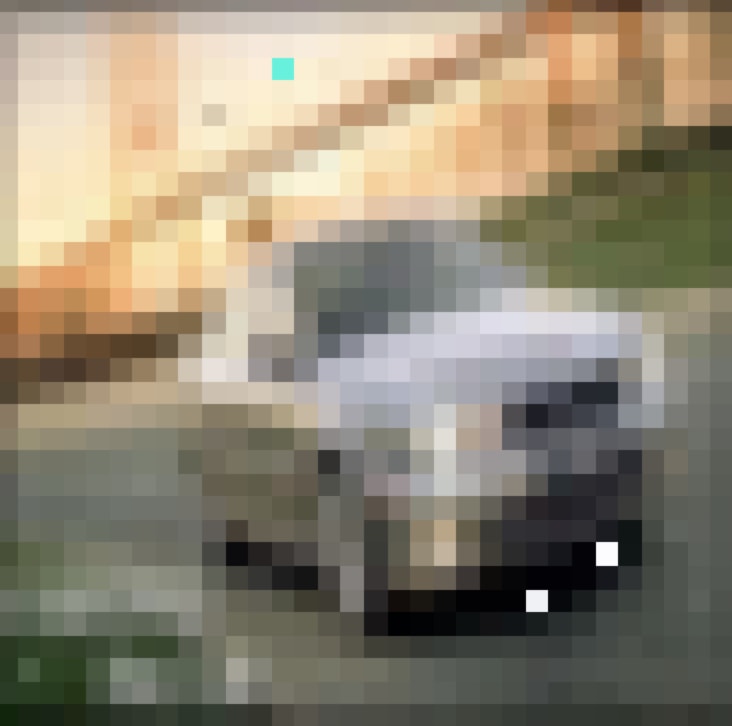}&\includegraphics[scale=.05]{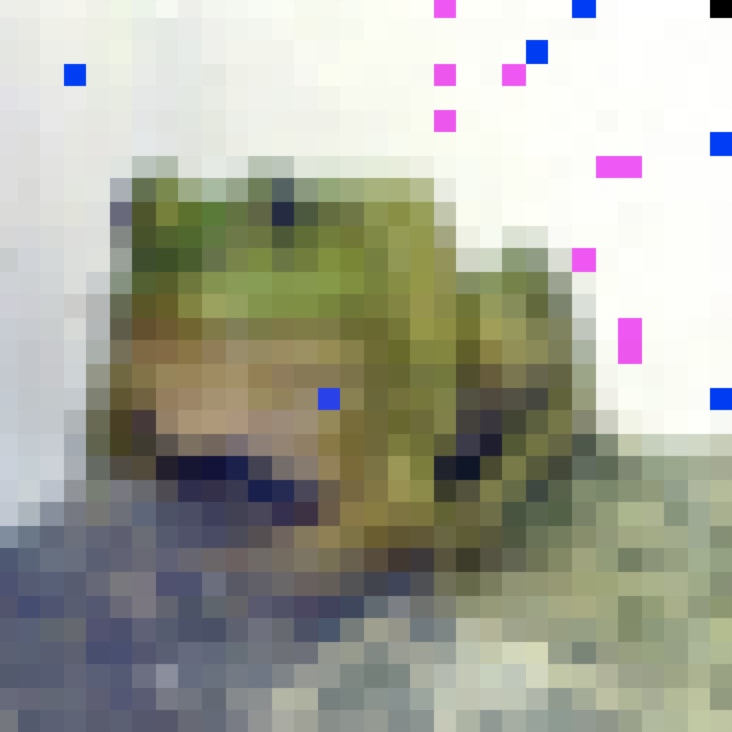}&\includegraphics[scale=.05]{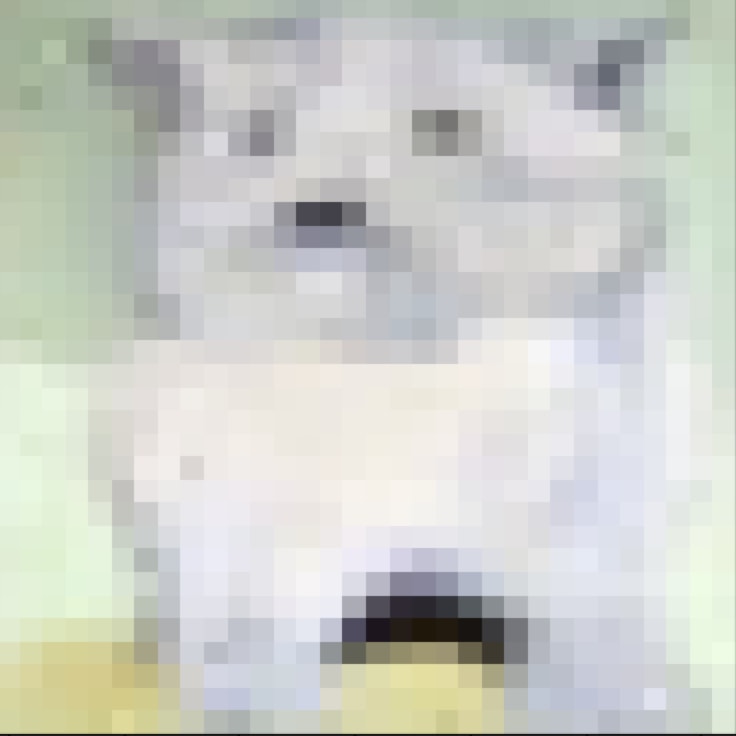}\\
		Ship & Frog & Truck & Deer\\
	\end{tabular}
	\caption{Top: Original examples from CIFAR10 dataset; Bottom: Misclassified examples generated by CMA-ES for inputs from top row using pixel perturbation}
	\label{fig:CIFAR_ex_pixel}
\end{figure}

In order to generate adversarial examples, we considered pixel perturbations with all $1024$ pixels as optimization variables of CMA-ES.  
The number of iterations was $180$, population size was $L = 4$, and parent size was $K = 2$. 
For each of the classes, we chose $10$ random samples from the test dataset and generated the adversarial examples.
Fig.~\ref{fig:CIFAR_ex_pixel} demonstrates four of these examples. 
Table~\ref{tb:CIFAR10_all_pixels} shows the average perturbation and the corresponding success rate for each of the classes. 
The overall success rate is $100\%$ with average perturbation $1.02\%$.

\begin{figure}[H]
	
	\centering
	\begin{tabular}{cccc}
		\includegraphics[scale=.05]{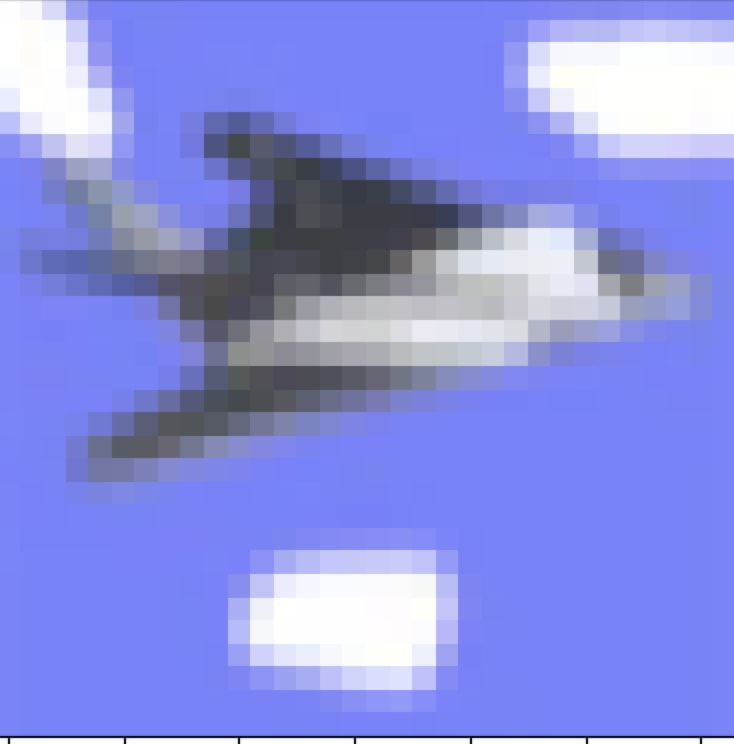}&\includegraphics[scale=.05]{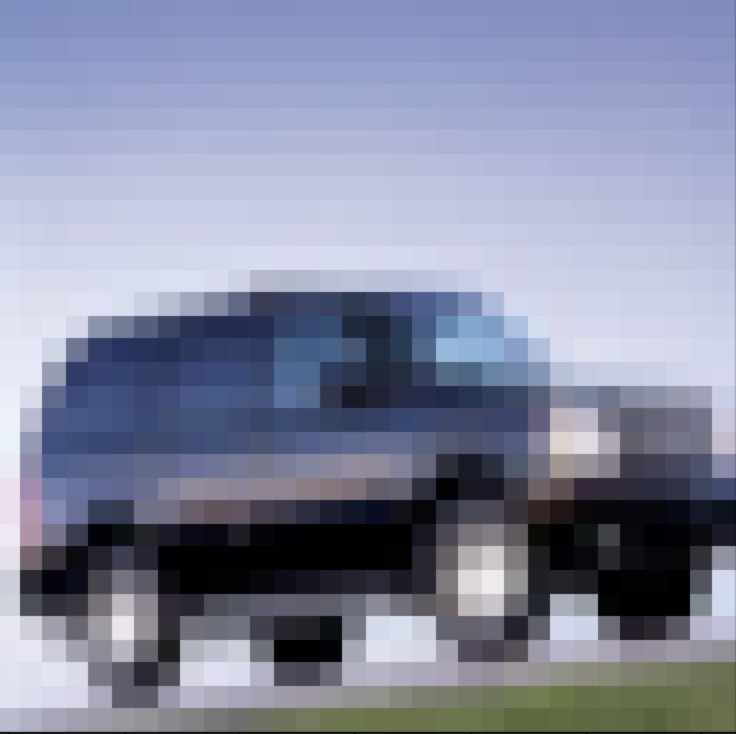}&\includegraphics[scale=.05]{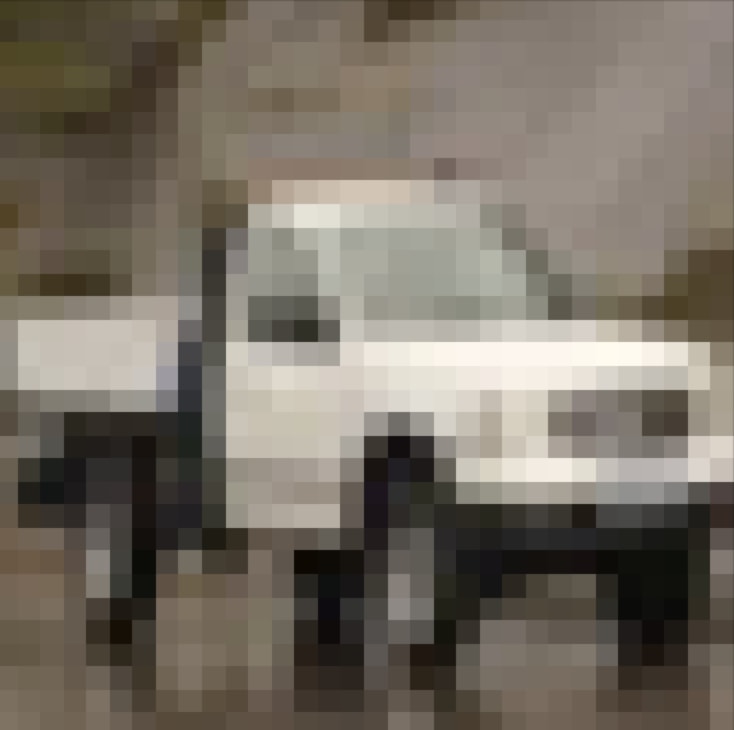}&\includegraphics[scale=.05]{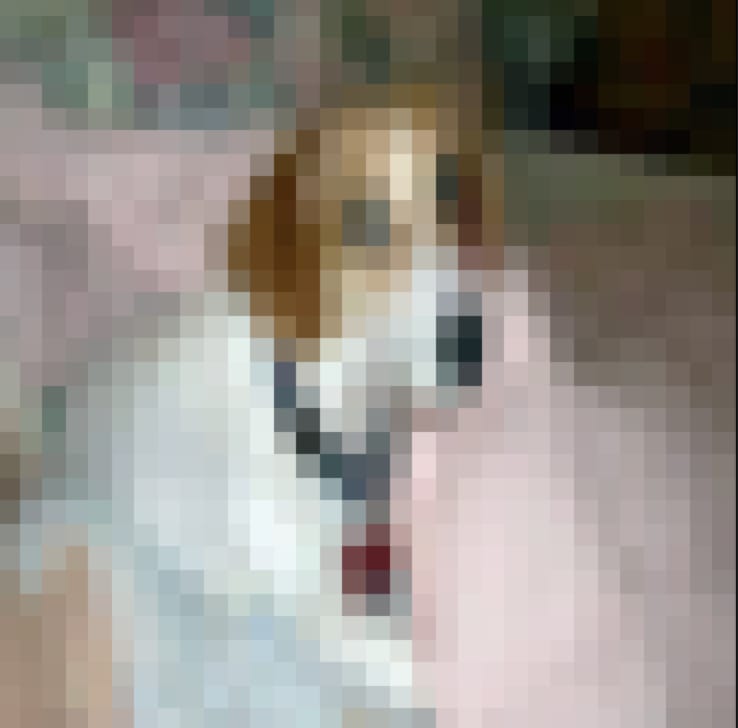}\\
		Plane & Car & Truck& Dog\\
		\\
		\includegraphics[scale=.05]{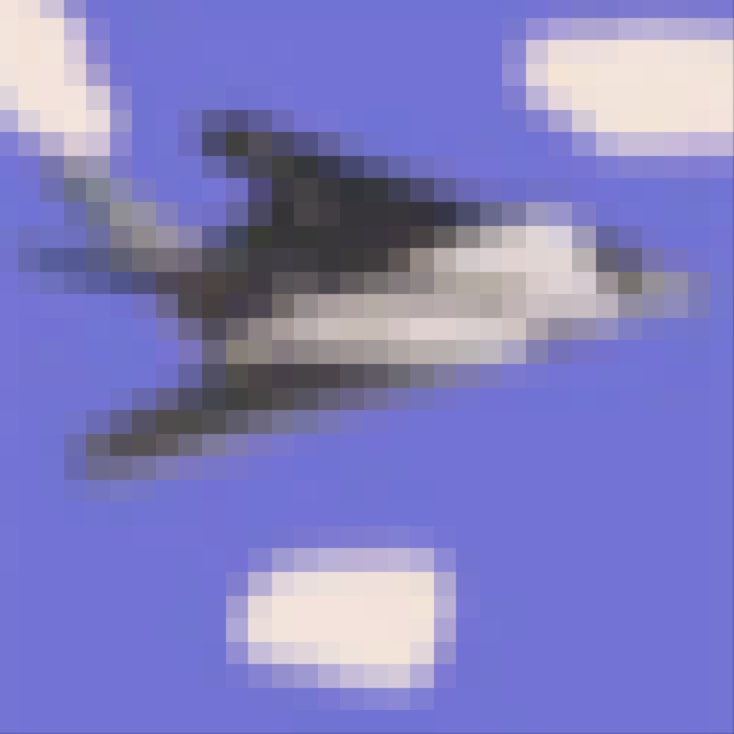}&\includegraphics[scale=.05]{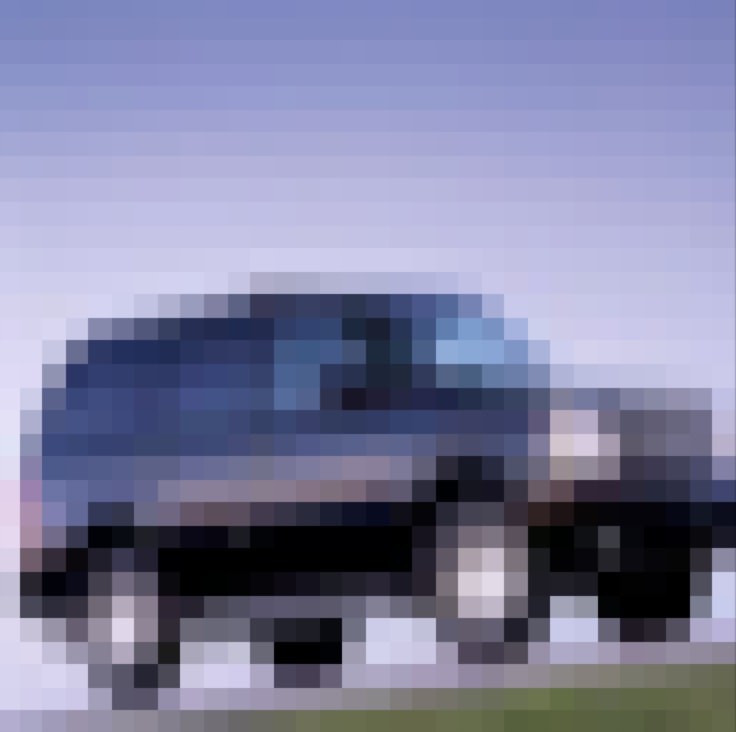}&\includegraphics[scale=.05]{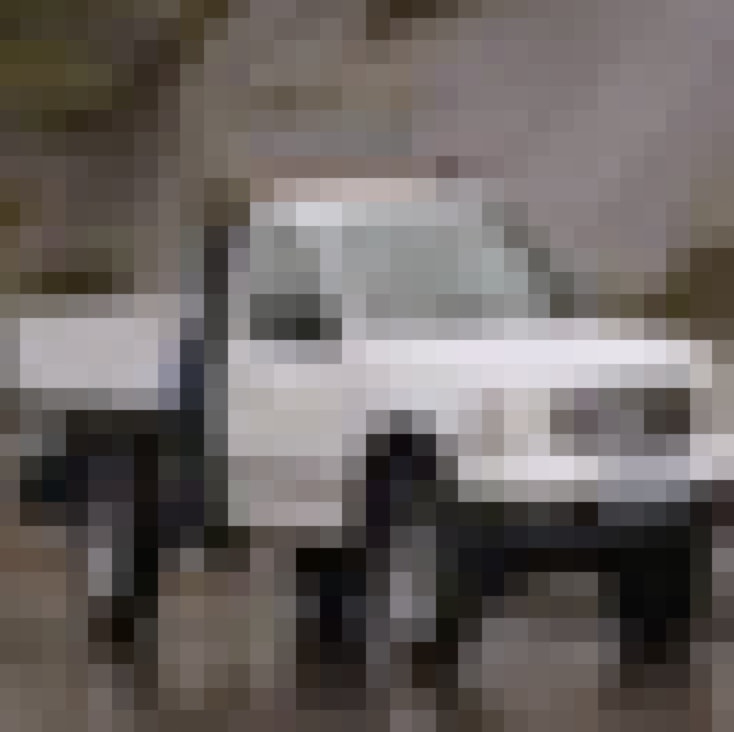}&\includegraphics[scale=.05]{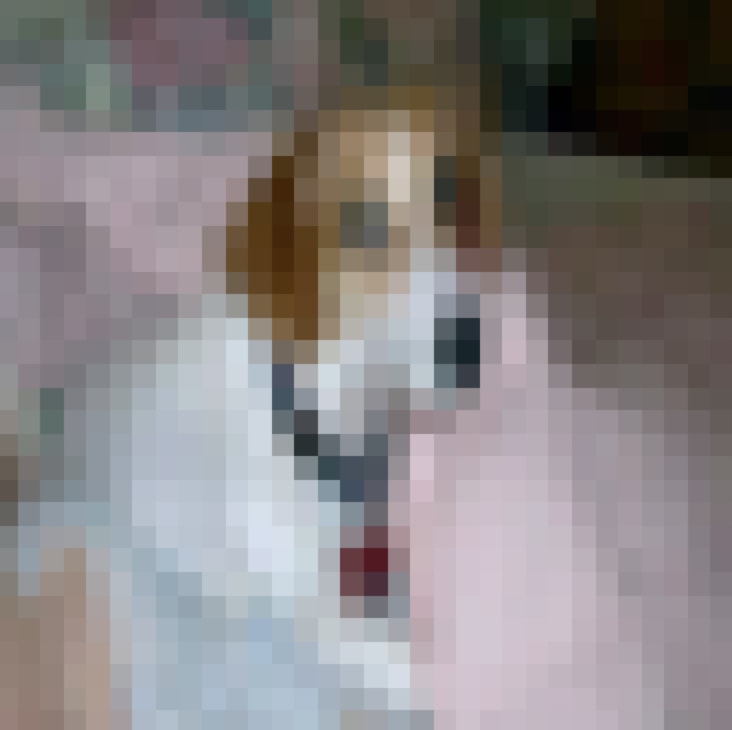}\\
		Ship & Truck & Car & Deer\\
	\end{tabular}
	\caption{Top: Original examples from CIFAR10 dataset; Bottom: Misclassified examples generated by CMA-ES for inputs from top row using color combination change}
	\label{fig:CIFAR_ex_col}
\end{figure}

As with GTSRB, we ran CMA-ES with perturbations $\beta_r,\beta_g,\beta_b$ on darkening colors.
We ran CMA-ES for 3 iterations, with a population size of $L = 240$ and parent size of $K = 120$.
Fig.~\ref{fig:CIFAR_ex_col} shows four of the original and the corresponding adversarial examples.  
The success rate was again $100\%$ with average perturbation $17\%$.

%
%

\subsection{Benchmarking CMA-ES using $L_{\infty}$ norm}

In this section, we briefly discuss the results on using $L_{\infty}$ norm in the fitness function. 
We ran CMA-ES on the MNIST dataset using $L_{\infty}$ norm. 
In this experiment, we randomly selected $10$ input images from each class of MNIST and set iteration number, population number ($L$) and parent number ($K$) 
to $3$, $240$ and $120$ respectively. 
Table~\ref{tb:compare_bisection} shows the average perturbations 
in the $L_{\infty}$ sense to find misclassified examples with and without bisection. 
One can notice that bisection significantly improves the performance of CMA-ES in this experiment,
although the perturbations were relatively larger than all the cases when $L_1$ are used.

\begin{table*}[t]
	\begin{center}
		\caption{Comparing performance of CMA-ES using $L_{\infty}$ norm with and without bisection}\label{tb:compare_bisection}
		
		\begin{tabular}{|c|c|c|c|c|c|c|c|c|c|c|}
			Source class & 0& 1 &  2 & 3 & 4 & 5 & 6 & 7&8 &9 \\\hline
			Using bisection & 17.95\%&  22.16\% & 20.82\% & 15.1\% & 28.63\% & 13.89\% & 36.68\%&21.4\% & 20.92\%&26.31\% \\\hline
			Pure CMA-ES & $52.36\%$ & $58.69\%$ & $63.63\%$ &$60.04\%$& $51.01\%$ & $62.03\%$ & $66.33\%$ &$60.48\%$& $58.21\%$ & $51.86\%$\\\hline
		\end{tabular}
	\end{center}
\end{table*}

\smallskip
Overall, our experiments indicate that one can find adversarial examples using CMA-ES on a wide class of image recognition
benchmarks with a small number of iterations, and small population sizes.

\section{Experiments with Perception in the Loop}
\label{sec:human}

In this section, we describe our experiments running CMA-ES with human perception as the fitness function.
We describe three experiments.
The first experiment (Section~\ref{subsec:human_L1}) is a human subject experiment to check that perceptual
preferences is robust to individual participants.
This is important to ensure our subsequent study is not biased by the specific user.
The second experiment runs CMA-ES to find adversarial examples.
Finally, the third experiment shows that a DNN trained to be robust against $L_\infty$ norm can nevertheless
misclassify examples when run with perception-in-the-loop.

\subsection{Comparing human perception with $L_1$ norm}
\label{subsec:human_L1}

In this experiment, we study the robustness of perceptual similarity across human subjects.
Twenty students and researchers ($15$ male, $5$ female; age range 22-36 years) participated in this experiment. 
The observers were na\"ive to the purpose and the nature of the experiment. 
All participants had normal or corrected-to-normal vision. 
They volunteered to participate in the experiment and gave informed consent in accordance to the policies of 
our university's Committee for the Protection of Human Subjects, 
which approved the experimental protocol. 

Each stimulus in the experiment consisted of a single image (the source) and upto $20$ related images.
The participants were shown the set of images simultaneously and had to pick the $5$ images closest
to the source.

In order to create the stimuli, we selected $10$ sample images out of test dataset for GTSRB and CIGFAR10. 
For each image, we run perception based CMA-ES for $3$ iterations, setting $L=20$ and $K=5$.
After $3$ iterations, we stopped CMA-ES and stored the population at the last iteration. 
Finally, only misclassifying examples were kept. 
Since $L=20$, this number could be $20$ at maximum.  

Suppose that $i^{th}$ participant selects $5$ indexes for the $j^{th}$ image. 
Then, we define $h_{ij}$ as a $20\times 1$ binary vector with $1$s at places corresponding to the selected indexes and $0$s elsewhere. 
Doing so, we can compute mean vector $\bar h_i$ for the $i^{th}$ image. 
Therefore, we can measure spread of choices, using the following relation:
\[
\epsilon=\frac{1}{(n_p\times n_e)} \sum_{j=1}^{n_e}\sum_{i=1}^{n_p} \frac{ ||h_{ij}-\bar h_j||_1}{2}
\]
In above relation, $n_p$ and $n_e$ denote number of participants and examples, respectively. 
In our case, $n_p=20$ and $n_e=10$. 
Furthermore, division by $2$ is included since $h_{ij}$s are complementary for elements at which they are different. 
In our case, $\epsilon=0.7327$. 
Intuitively, this means that participants are different in less than one choice in average. 

We also computed $H_j$ similar to $h_{ij}$, but with respect to $L_1$ norm ranking. Again, we define
\[
E=\frac{1}{n_e} \sum_{j=1}^{n_e}\frac{||H_{j}-\bar h_j||_1}{2}
\]
to measure difference between the $L_1$ norm selection and average of participants. 
We computed $E=1.1844$.  
Intuitively, this means that participants' ranking (in average) is different from what $L_1$ norm suggests in more than one choice. 

These results show that human perception is robust across participants and moreover, 
the $L_1$ norm based ranking is different compared to a ranking based on human perception. 

	\begin{figure}[t]
	\begin{center}
		\begin{tikzpicture}[auto, node distance=1cm,>=latex',scale=.5]
		
		\node at (5,5)
		{\includegraphics[scale=.25,width=2cm,height=2cm]{Figs/orig_image}};
		
		\node at (0,0)
		{\includegraphics[scale=.25,width=1cm,height=1cm]{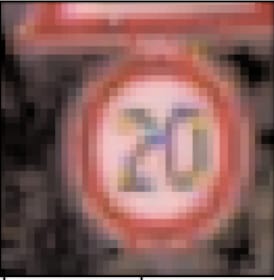}};
		\node at (2.5,0)
		{\includegraphics[scale=.25,width=1cm,height=1cm]{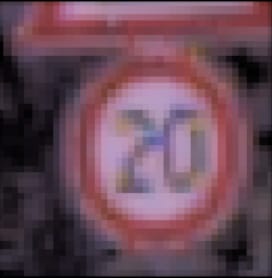}};
		\node at (5,0)
		{\includegraphics[scale=.25,width=1cm,height=1cm]{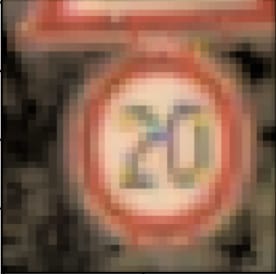}};
		\node at (7.5,0)
		{\includegraphics[scale=.25,width=1cm,height=1cm]{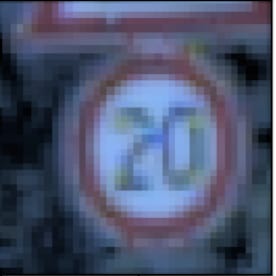}};
		\node at (10,0)
		{\includegraphics[scale=.25,width=1cm,height=1cm]{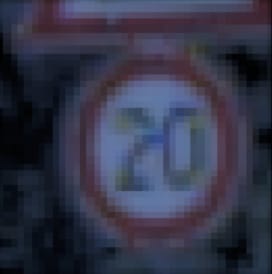}};
		
		\node at (0,-2.5)
		{\includegraphics[scale=.25,width=1cm,height=1cm]{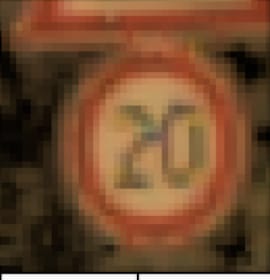}};
		\node at (2.5,-2.5)
		{\includegraphics[scale=.25,width=1cm,height=1cm]{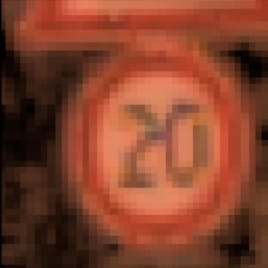}};
		\node at (5,-2.5)
		{\includegraphics[scale=.25,width=1cm,height=1cm]{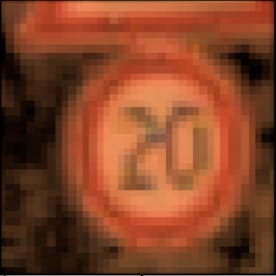}};
		\node at (7.5,-2.5)
		{\includegraphics[scale=.25,width=1cm,height=1cm]{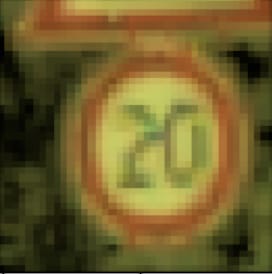}};
		\node at (10,-2.5)
		{\includegraphics[scale=.25,width=1cm,height=1cm]{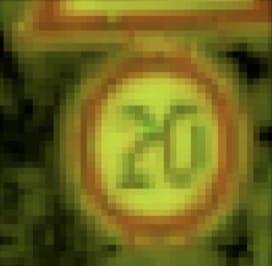}};

		\node at (0,-5)
		{\includegraphics[scale=.25,width=1cm,height=1cm]{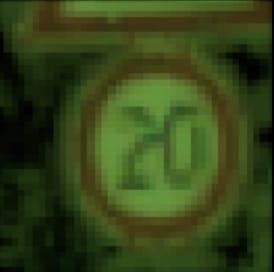}};
		\node at (2.5,-5)
		{\includegraphics[scale=.25,width=1cm,height=1cm]{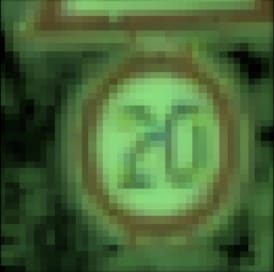}};
		\node at (5,-5)
		{\includegraphics[scale=.25,width=1cm,height=1cm]{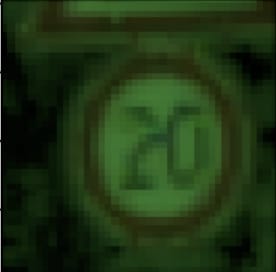}};
		\node at (7.5,-5)
		{\includegraphics[scale=.25,width=1cm,height=1cm]{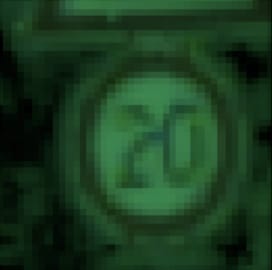}};
		\node at (10,-5)
		{\includegraphics[scale=.25,width=1cm,height=1cm]{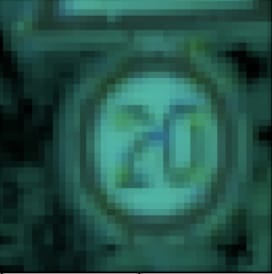}};
		
		\node at (0,-7.5)
		{\includegraphics[scale=.25,width=1cm,height=1cm]{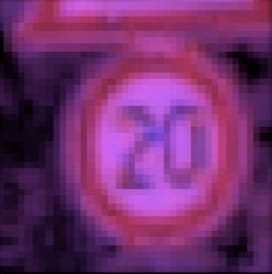}};
		\node at (2.5,-7.5)
		{\includegraphics[scale=.25,width=1cm,height=1cm]{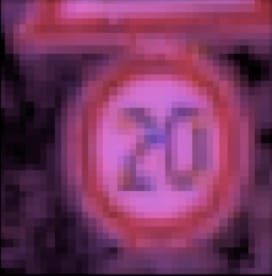}};
		\node at (5,-7.5)
		{\includegraphics[scale=.25,width=1cm,height=1cm]{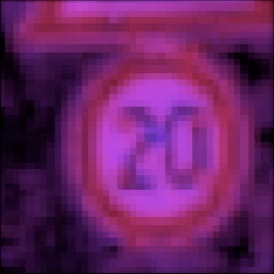}};
		\node at (7.5,-7.5)
		{\includegraphics[scale=.25,width=1cm,height=1cm]{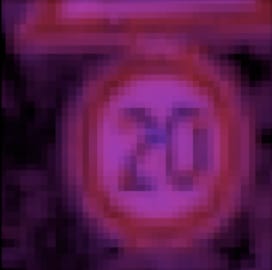}};

		\node at (0,0-12)
		{\includegraphics[scale=.25,width=1cm,height=1cm]{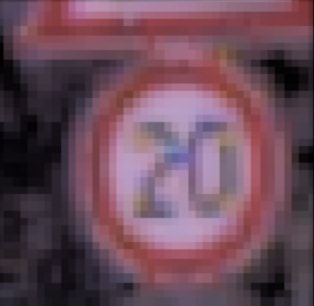}};
		\node at (2.5,0-12)
		{\includegraphics[scale=.25,width=1cm,height=1cm]{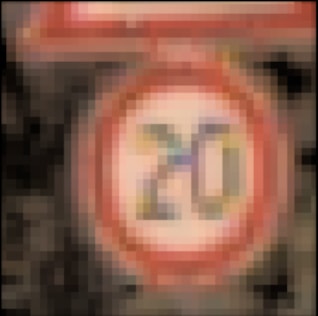}};
		\node at (5,0-12)
		{\includegraphics[scale=.25,width=1cm,height=1cm]{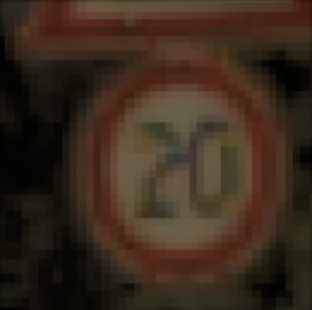}};
		\node at (7.5,0-12)
		{\includegraphics[scale=.25,width=1cm,height=1cm]{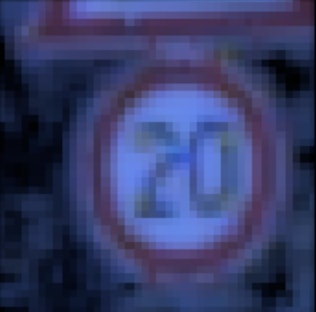}};
		\node at (10,0-12)
		{\includegraphics[scale=.25,width=1cm,height=1cm]{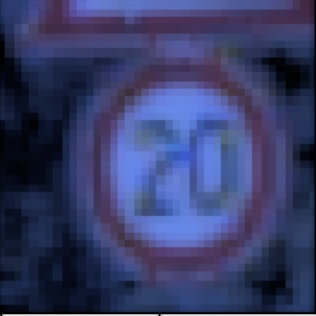}};
		
		\node at (0,-2.5-12)
		{\includegraphics[scale=.25,width=1cm,height=1cm]{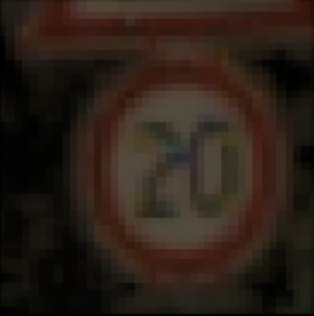}};
		\node at (2.5,-2.5-12)
		{\includegraphics[scale=.25,width=1cm,height=1cm]{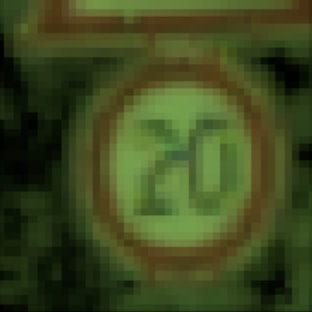}};
		\node at (5,-2.5-12)
		{\includegraphics[scale=.25,width=1cm,height=1cm]{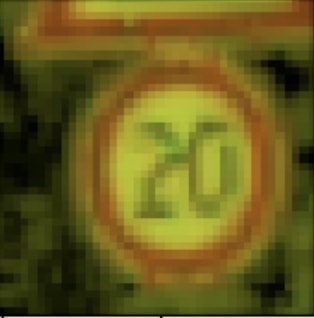}};
		\node at (7.5,-2.5-12)
		{\includegraphics[scale=.25,width=1cm,height=1cm]{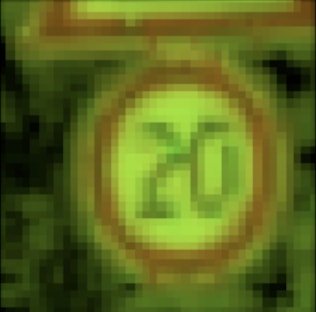}};
		\node at (10,-2.5-12)
		{\includegraphics[scale=.25,width=1cm,height=1cm]{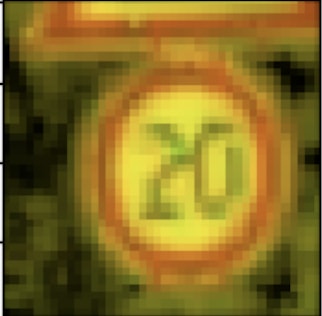}};

		\node at (0,-5-12)
		{\includegraphics[scale=.25,width=1cm,height=1cm]{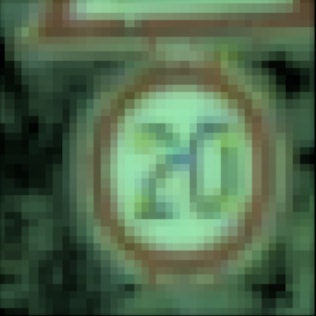}};
		\node at (2.5,-5-12)
		{\includegraphics[scale=.25,width=1cm,height=1cm]{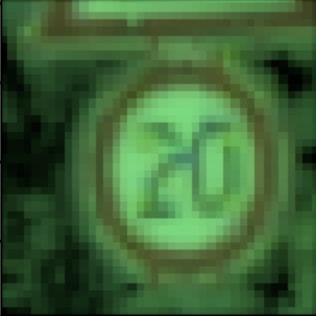}};
		\node at (5,-5-12)
		{\includegraphics[scale=.25,width=1cm,height=1cm]{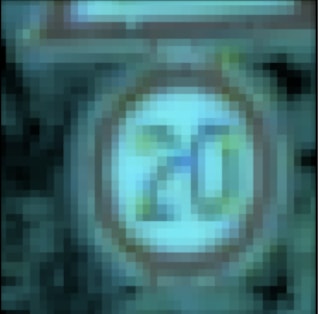}};
		\node at (7.5,-5-12)
		{\includegraphics[scale=.25,width=1cm,height=1cm]{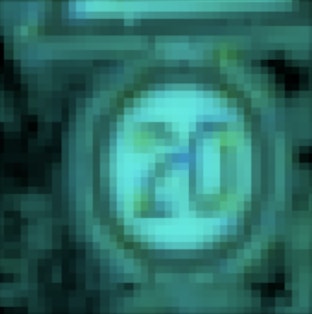}};
		\node at (10,-5-12)
		{\includegraphics[scale=.25,width=1cm,height=1cm]{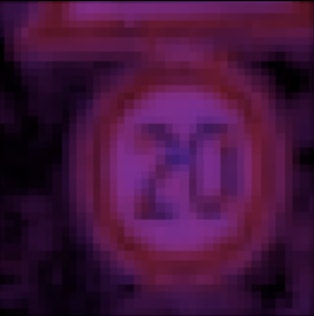}};
		
		\node at (0,-7.5-12)
		{\includegraphics[scale=.25,width=1cm,height=1cm]{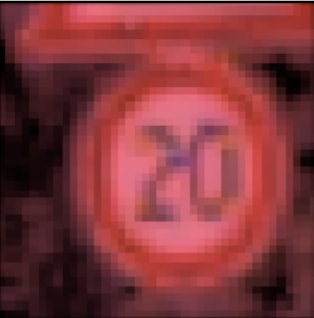}};
		\node at (2.5,-7.5-12)
		{\includegraphics[scale=.25,width=1cm,height=1cm]{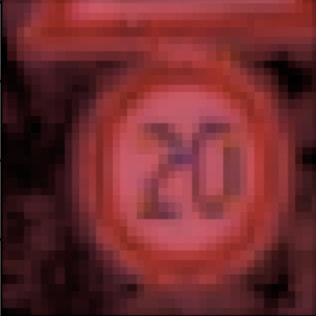}};
		\node at (5,-7.5-12)
		{\includegraphics[scale=.25,width=1cm,height=1cm]{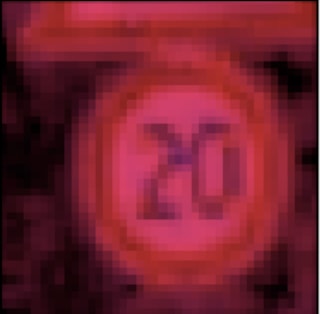}};

		\node at (0,0-24)
		{\includegraphics[scale=.25,width=1cm,height=1cm]{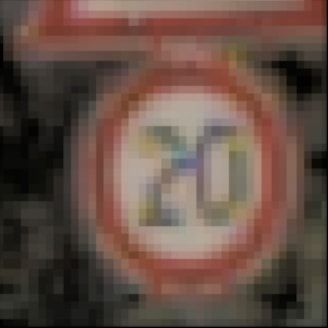}};
		\node at (2.5,0-24)
		{\includegraphics[scale=.25,width=1cm,height=1cm]{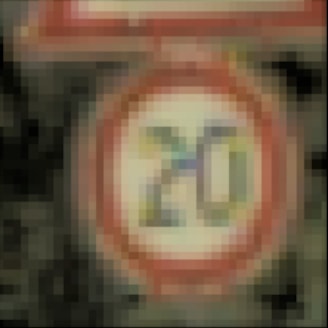}};
		\node at (5,0-24)
		{\includegraphics[scale=.25,width=1cm,height=1cm]{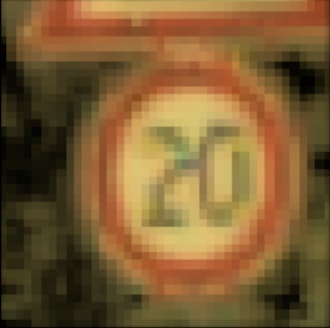}};
		\node at (7.5,0-24)
		{\includegraphics[scale=.25,width=1cm,height=1cm]{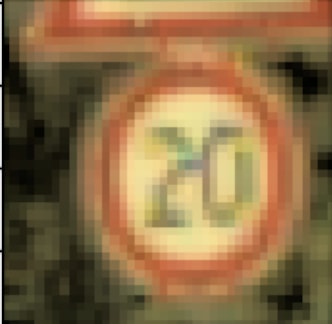}};
		\node at (10,0-24)
		{\includegraphics[scale=.25,width=1cm,height=1cm]{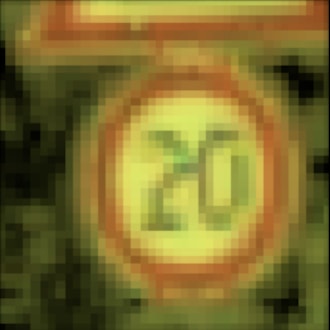}};
		
		\node at (0,-2.5-24)
		{\includegraphics[scale=.25,width=1cm,height=1cm]{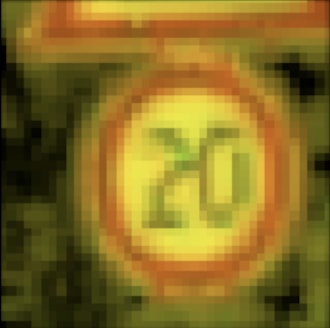}};
		\node at (2.5,-2.5-24)
		{\includegraphics[scale=.25,width=1cm,height=1cm]{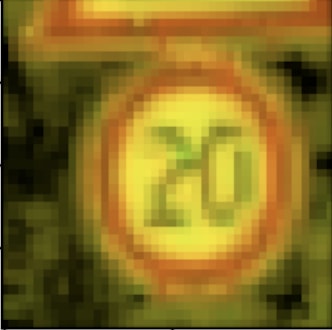}};
		\node at (5,-2.5-24)
		{\includegraphics[scale=.25,width=1cm,height=1cm]{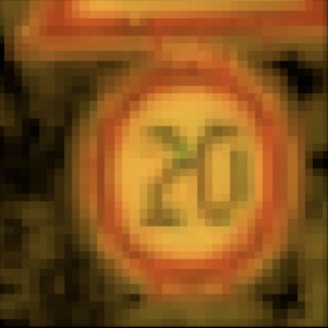}};
		\node at (7.5,-2.5-24)
		{\includegraphics[scale=.25,width=1cm,height=1cm]{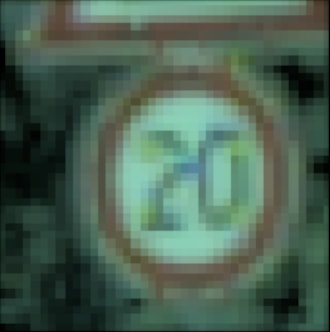}};
		\node at (10,-2.5-24)
		{\includegraphics[scale=.25,width=1cm,height=1cm]{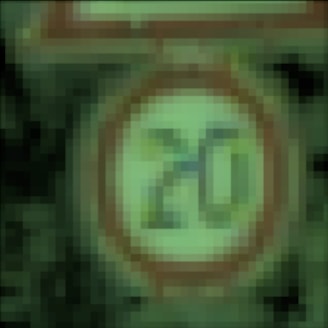}};

		\node at (0,-5-24)
		{\includegraphics[scale=.25,width=1cm,height=1cm]{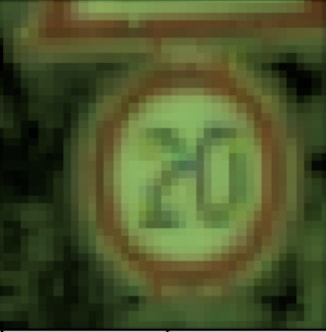}};
		\node at (2.5,-5-24)
		{\includegraphics[scale=.25,width=1cm,height=1cm]{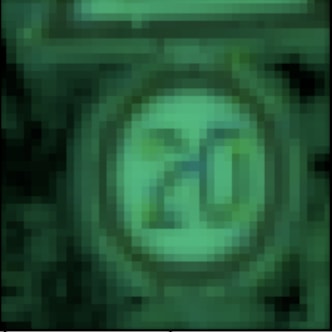}};
		\node at (5,-5-24)
		{\includegraphics[scale=.25,width=1cm,height=1cm]{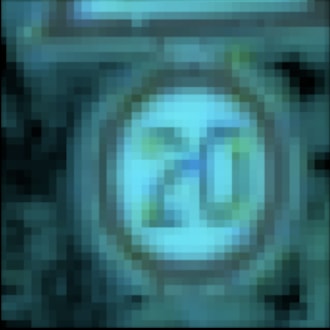}};
		\node at (7.5,-5-24)
		{\includegraphics[scale=.25,width=1cm,height=1cm]{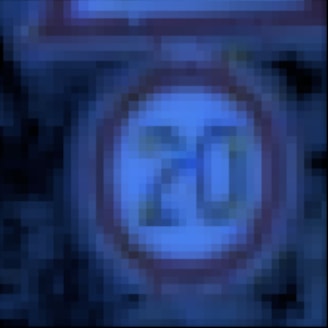}};
		\node at (10,-5-24)
		{\includegraphics[scale=.25,width=1cm,height=1cm]{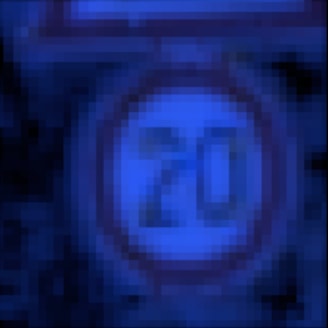}};
		
		\node at (0,-7.5-24)
		{\includegraphics[scale=.25,width=1cm,height=1cm]{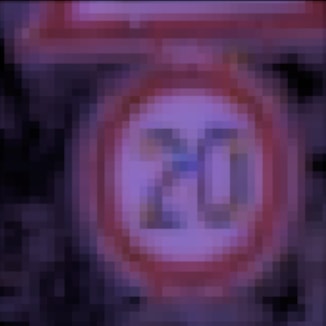}};
		\node at (2.5,-7.5-24)
		{\includegraphics[scale=.25,width=1cm,height=1cm]{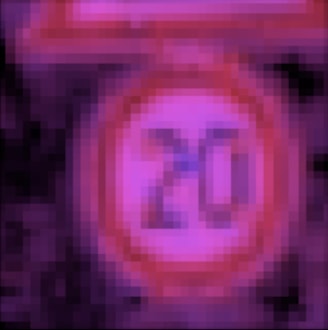}};
		\node at (5,-7.5-24)
		{\includegraphics[scale=.25,width=1cm,height=1cm]{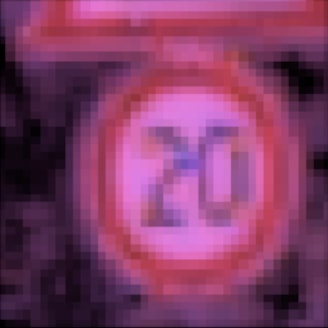}};
		\node at (5,2.5) {(a)};
		\node at (5,-9.5) {(b)};
		\node at (5,-21.5) {(c)};
		\node at (5,-33.5) {(d)};
		\end{tikzpicture}
	\end{center}
	\caption{Results of running CMA-ES using human perception for an example from GTSRB dataset; (a): Input image; (b): misclassified examples within the first iteration of CMA-ES run; (b): misclassified examples within the second iteration of CMA-ES run; (c): misclassified examples within the third iteration of CMA-ES run}\label{fig:CMAES_GTSRB}
\end{figure}

\begin{figure}[t]
	\begin{center}
		\begin{tikzpicture}[auto, node distance=1cm,>=latex',scale=.5]

		\node at (5,5)
		{\includegraphics[scale=.25,width=2cm,height=2cm]{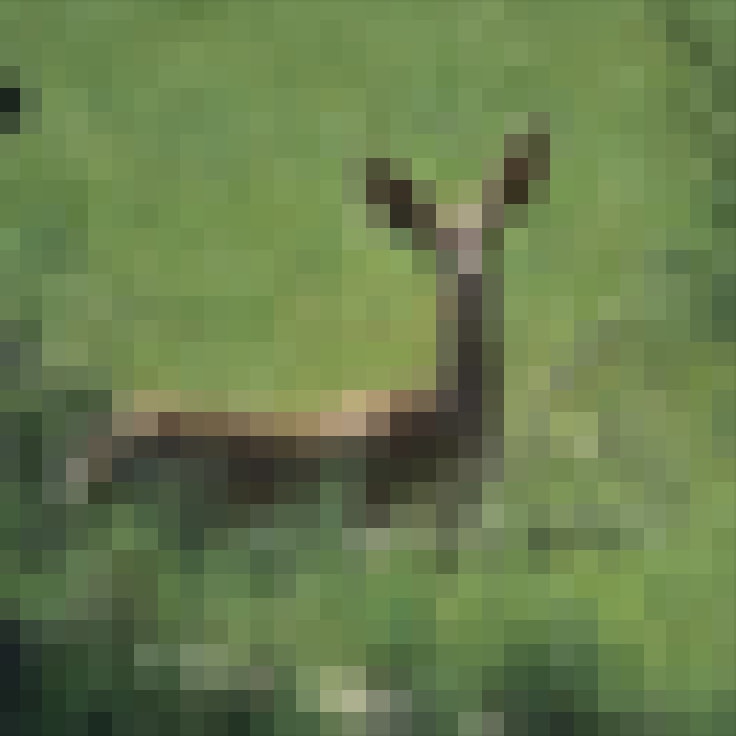}};

		\node at (0,0)
		{\includegraphics[scale=.25,width=1cm,height=1cm]{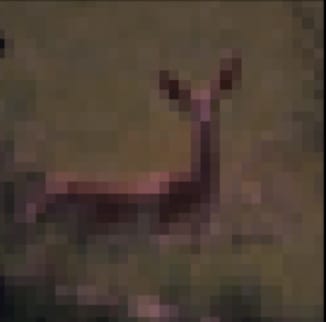}};
		\node at (2.5,0)
		{\includegraphics[scale=.25,width=1cm,height=1cm]{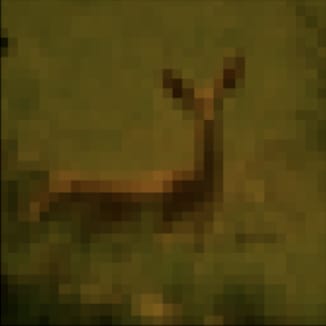}};
		\node at (5,0)
		{\includegraphics[scale=.25,width=1cm,height=1cm]{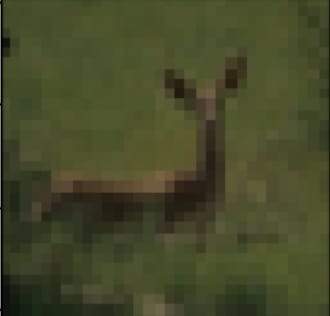}};
		\node at (7.5,0)
		{\includegraphics[scale=.25,width=1cm,height=1cm]{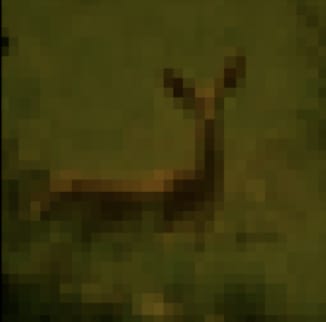}};
		\node at (10,0)
		{\includegraphics[scale=.25,width=1cm,height=1cm]{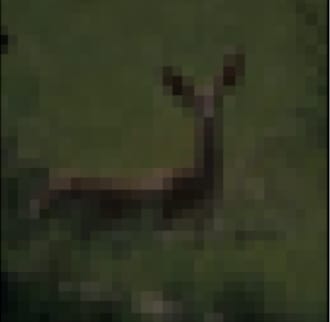}};
		
		\node at (0,-2.5)
		{\includegraphics[scale=.25,width=1cm,height=1cm]{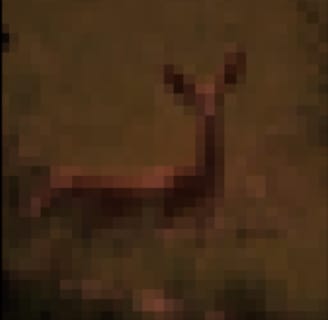}};
		\node at (2.5,-2.5)
		{\includegraphics[scale=.25,width=1cm,height=1cm]{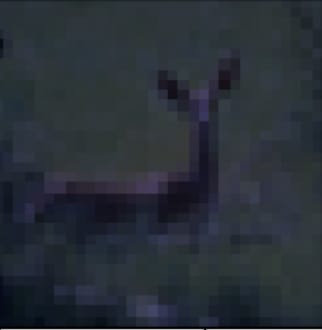}};
		\node at (5,-2.5)
		{\includegraphics[scale=.25,width=1cm,height=1cm]{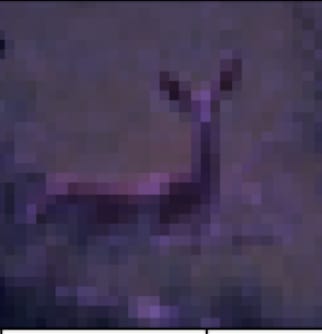}};
		\node at (7.5,-2.5)
		{\includegraphics[scale=.25,width=1cm,height=1cm]{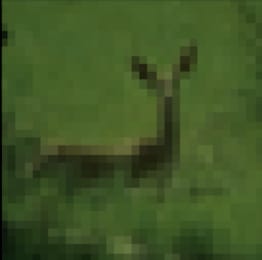}};
		\node at (10,-2.5)
		{\includegraphics[scale=.25,width=1cm,height=1cm]{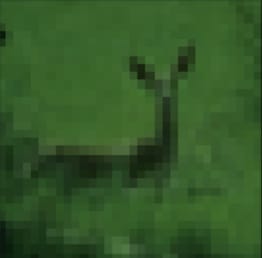}};

		\node at (0,-5)
		{\includegraphics[scale=.25,width=1cm,height=1cm]{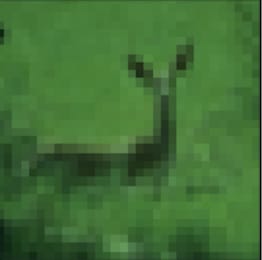}};
		\node at (2.5,-5)
		{\includegraphics[scale=.25,width=1cm,height=1cm]{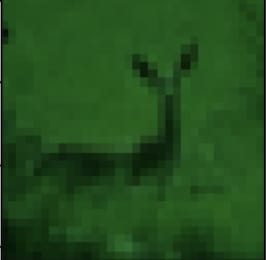}};
		\node at (5,-5)
		{\includegraphics[scale=.25,width=1cm,height=1cm]{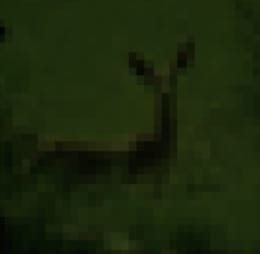}};
		\node at (7.5,-5)
		{\includegraphics[scale=.25,width=1cm,height=1cm]{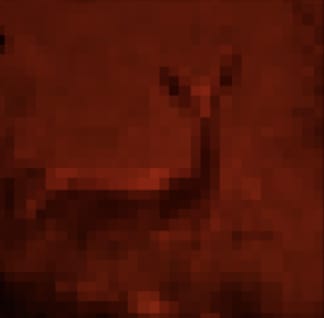}};
		\node at (10,-5)
		{\includegraphics[scale=.25,width=1cm,height=1cm]{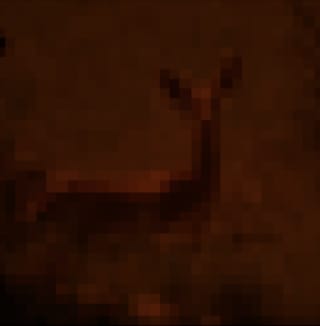}};
		
		\node at (0,-7.5)
		{\includegraphics[scale=.25,width=1cm,height=1cm]{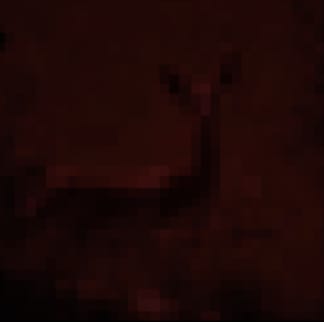}};
		\node at (2.5,-7.5)
		{\includegraphics[scale=.25,width=1cm,height=1cm]{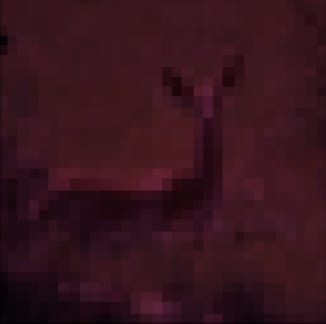}};
		\node at (5,-7.5)
		{\includegraphics[scale=.25,width=1cm,height=1cm]{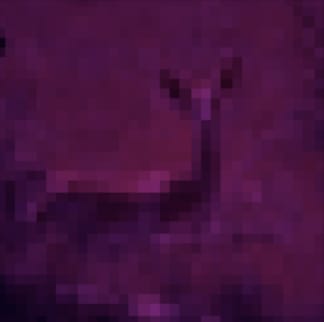}};
		\node at (7.5,-7.5)
		{\includegraphics[scale=.25,width=1cm,height=1cm]{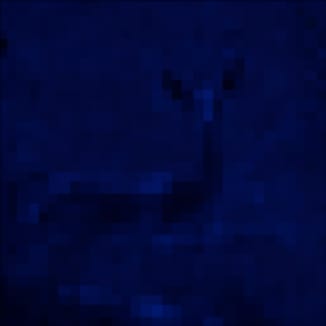}};

		\node at (0,0-12)
		{\includegraphics[scale=.25,width=1cm,height=1cm]{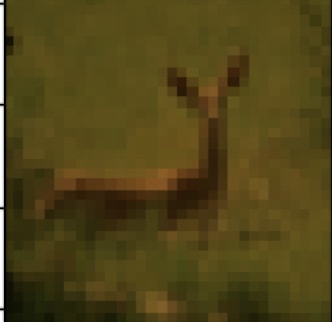}};
		\node at (2.5,0-12)
		{\includegraphics[scale=.25,width=1cm,height=1cm]{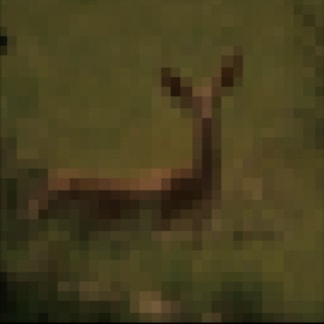}};
		\node at (5,0-12)
		{\includegraphics[scale=.25,width=1cm,height=1cm]{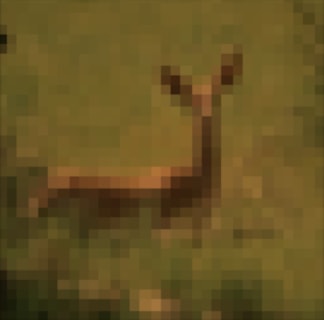}};
		\node at (7.5,0-12)
		{\includegraphics[scale=.25,width=1cm,height=1cm]{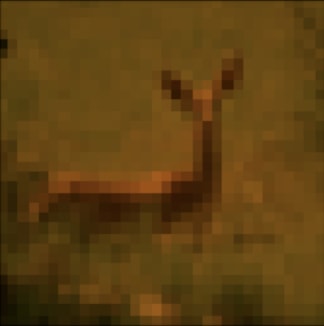}};
		\node at (10,0-12)
		{\includegraphics[scale=.25,width=1cm,height=1cm]{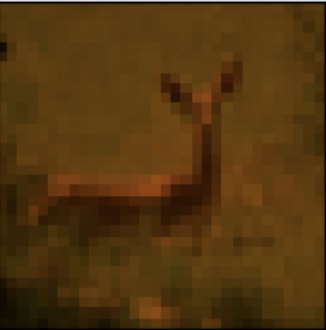}};
		
		\node at (0,-2.5-12)
		{\includegraphics[scale=.25,width=1cm,height=1cm]{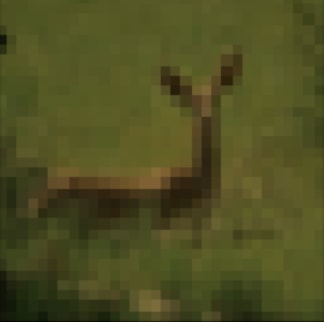}};
		\node at (2.5,-2.5-12)
		{\includegraphics[scale=.25,width=1cm,height=1cm]{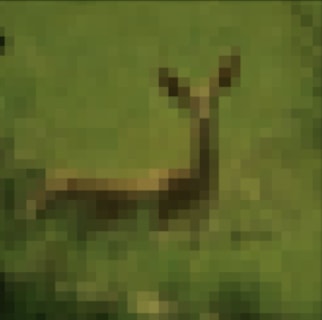}};
		\node at (5,-2.5-12)
		{\includegraphics[scale=.25,width=1cm,height=1cm]{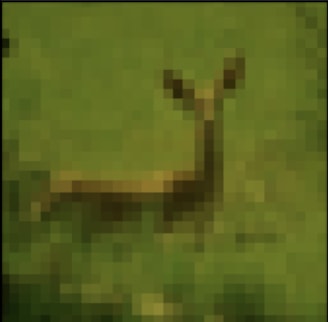}};
		\node at (7.5,-2.5-12)
		{\includegraphics[scale=.25,width=1cm,height=1cm]{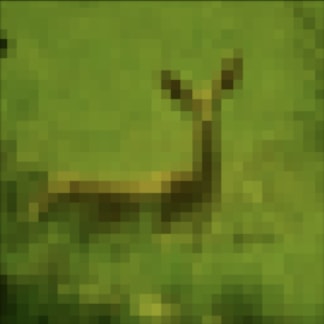}};
		\node at (10,-2.5-12)
		{\includegraphics[scale=.25,width=1cm,height=1cm]{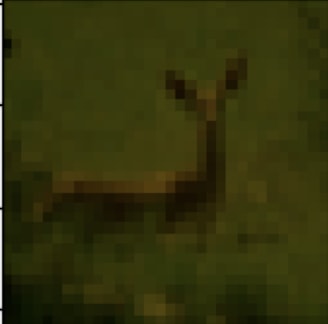}};

		\node at (0,-5-12)
		{\includegraphics[scale=.25,width=1cm,height=1cm]{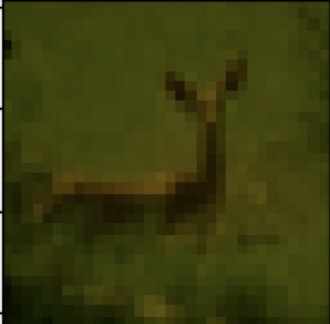}};
		\node at (2.5,-5-12)
		{\includegraphics[scale=.25,width=1cm,height=1cm]{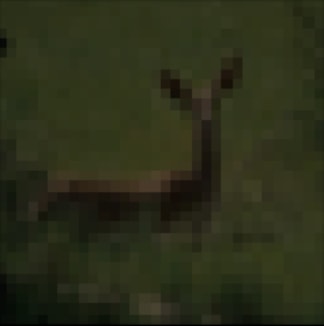}};
		\node at (5,-5-12)
		{\includegraphics[scale=.25,width=1cm,height=1cm]{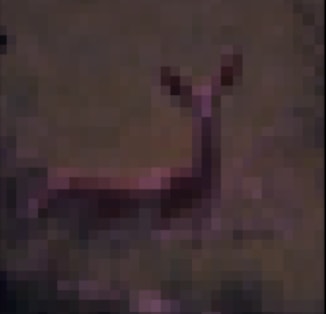}};
		\node at (7.5,-5-12)
		{\includegraphics[scale=.25,width=1cm,height=1cm]{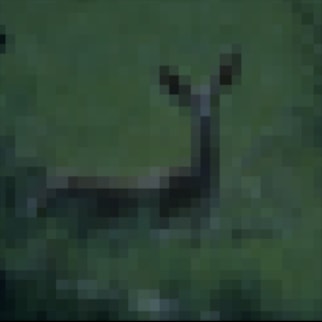}};
		\node at (10,-5-12)
		{\includegraphics[scale=.25,width=1cm,height=1cm]{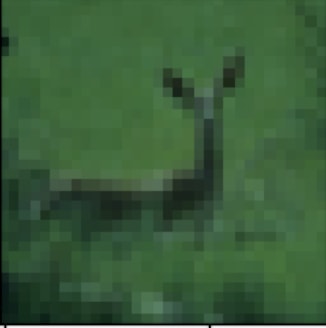}};
		
		\node at (0,-7.5-12)
		{\includegraphics[scale=.25,width=1cm,height=1cm]{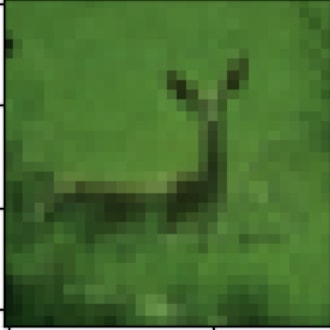}};
		\node at (2.5,-7.5-12)
		{\includegraphics[scale=.25,width=1cm,height=1cm]{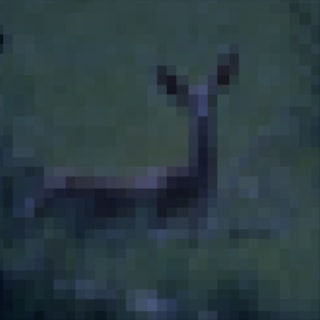}};
		\node at (5,-7.5-12)
		{\includegraphics[scale=.25,width=1cm,height=1cm]{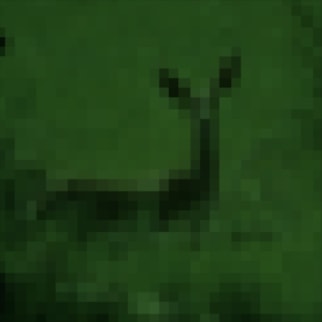}};
		\node at (7.5,-7.5-12)
		{\includegraphics[scale=.25,width=1cm,height=1cm]{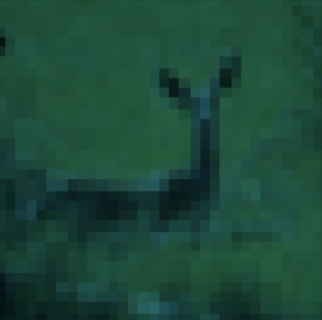}};
		\node at (10,-7.5-12)
		{\includegraphics[scale=.25,width=1cm,height=1cm]{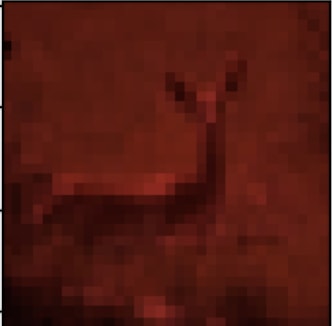}};

		\node at (0,0-24)
		{\includegraphics[scale=.25,width=1cm,height=1cm]{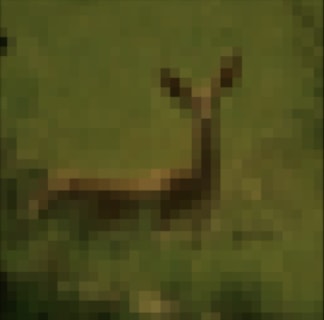}};
		\node at (2.5,0-24)
		{\includegraphics[scale=.25,width=1cm,height=1cm]{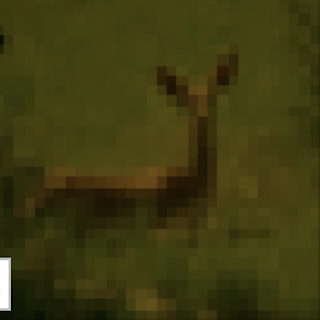}};
		\node at (5,0-24)
		{\includegraphics[scale=.25,width=1cm,height=1cm]{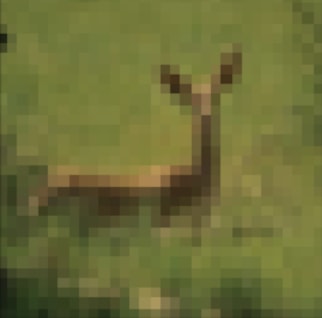}};
		\node at (7.5,0-24)
		{\includegraphics[scale=.25,width=1cm,height=1cm]{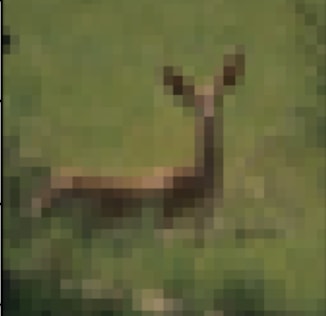}};
		\node at (10,0-24)
		{\includegraphics[scale=.25,width=1cm,height=1cm]{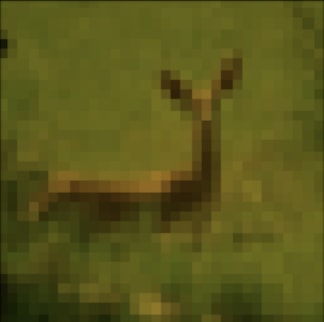}};
		
		\node at (0,-2.5-24)
		{\includegraphics[scale=.25,width=1cm,height=1cm]{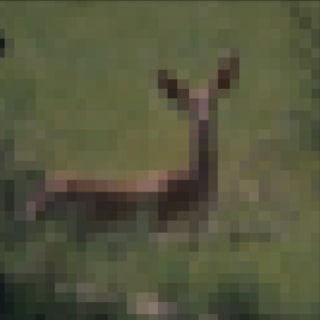}};
		\node at (2.5,-2.5-24)
		{\includegraphics[scale=.25,width=1cm,height=1cm]{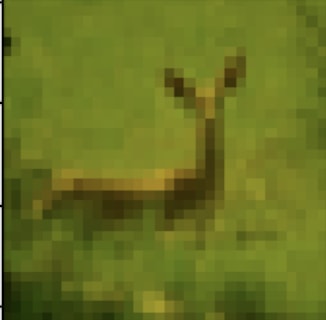}};
		\node at (5,-2.5-24)
		{\includegraphics[scale=.25,width=1cm,height=1cm]{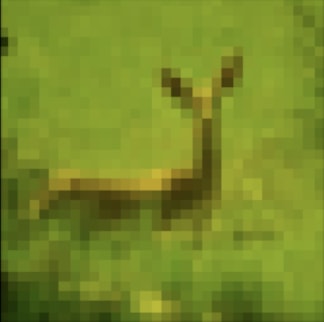}};
		\node at (7.5,-2.5-24)
		{\includegraphics[scale=.25,width=1cm,height=1cm]{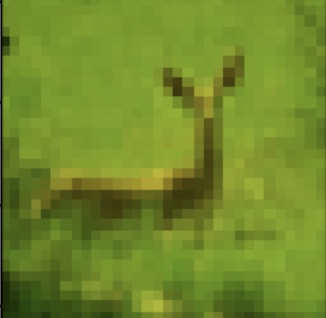}};
		\node at (10,-2.5-24)
		{\includegraphics[scale=.25,width=1cm,height=1cm]{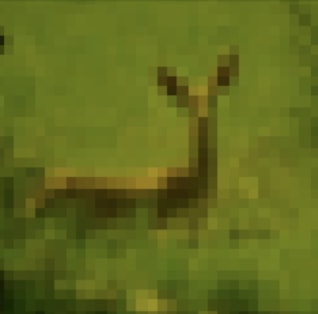}};

		\node at (0,-5-24)
		{\includegraphics[scale=.25,width=1cm,height=1cm]{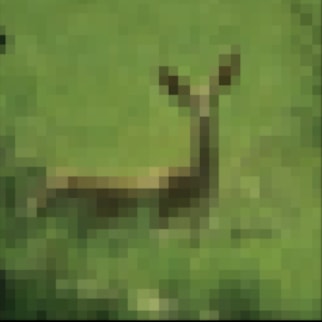}};
		\node at (2.5,-5-24)
		{\includegraphics[scale=.25,width=1cm,height=1cm]{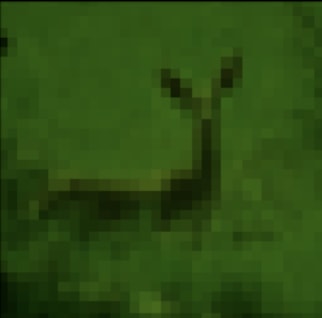}};
		\node at (5,-5-24)
		{\includegraphics[scale=.25,width=1cm,height=1cm]{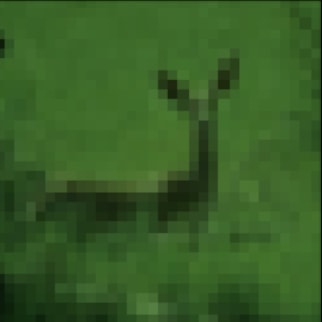}};
		\node at (7.5,-5-24)
		{\includegraphics[scale=.25,width=1cm,height=1cm]{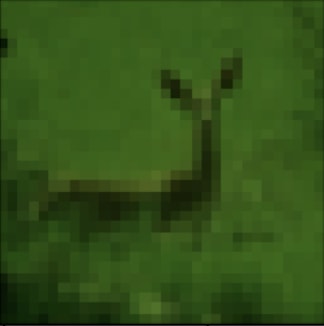}};
		\node at (10,-5-24)
		{\includegraphics[scale=.25,width=1cm,height=1cm]{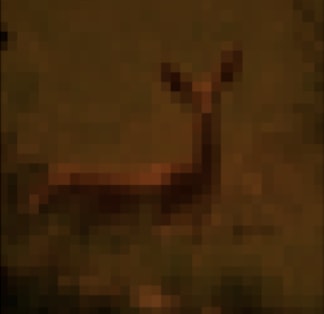}};
		
		\node at (0,-7.5-24)
		{\includegraphics[scale=.25,width=1cm,height=1cm]{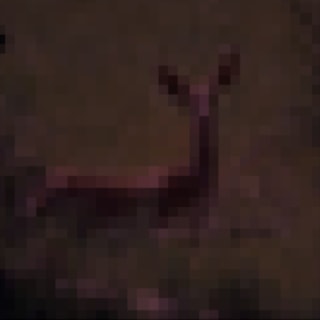}};
		\node at (2.5,-7.5-24)
		{\includegraphics[scale=.25,width=1cm,height=1cm]{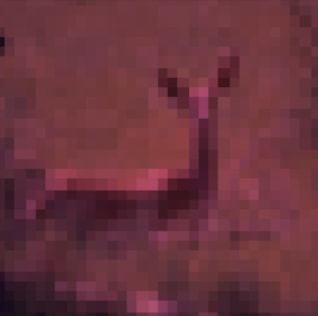}};
		\node at (5,-7.5-24)
		{\includegraphics[scale=.25,width=1cm,height=1cm]{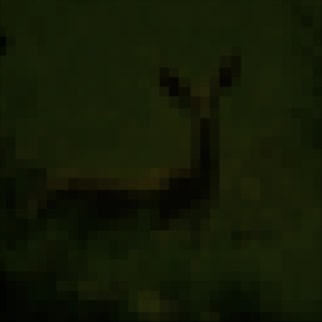}};
		\node at (7.5,-7.5-24)
		{\includegraphics[scale=.25,width=1cm,height=1cm]{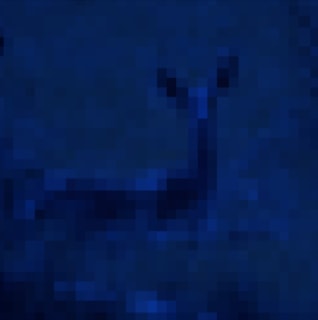}};
		\node at (5,2.5) {(a)};
		\node at (5,-9.5) {(b)};
		\node at (5,-21.5) {(c)};
		\node at (5,-33.5) {(d)};
		\end{tikzpicture}
	\end{center}
	\caption{Results of running CMA-ES using human perception for an example from CIFAR10 dataset; (a): Input image; (b): misclassified examples within the first iteration of CMA-ES run; (b): misclassified examples within the second iteration of CMA-ES run; (c): misclassified examples within the third iteration of CMA-ES run}\label{fig:CMAES_CIFAR}
\end{figure}

\begin{figure}[t]
	\centering
	\begin{tabular}{ccc}
		\includegraphics[height=2cm,width=2cm]{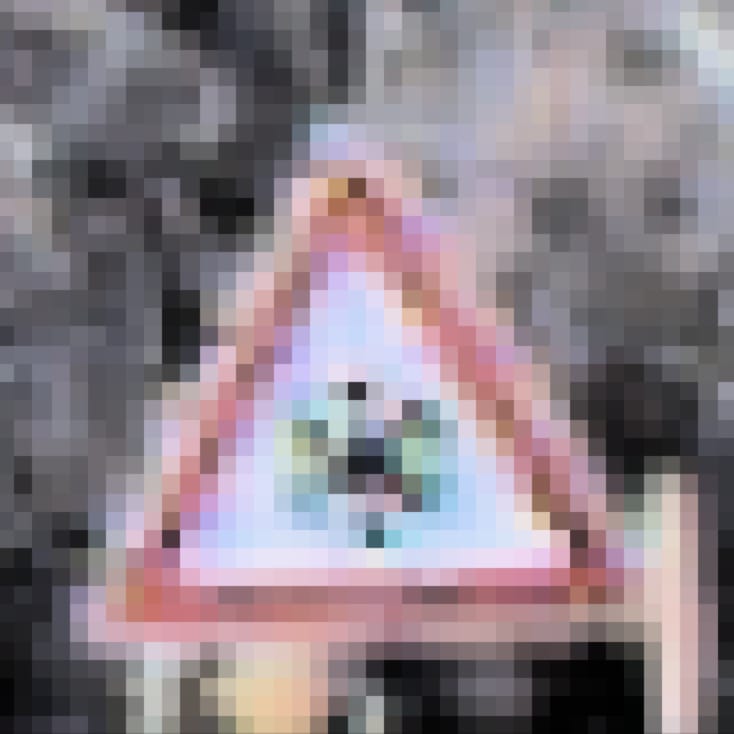}&\includegraphics[height=2cm,width=2cm]{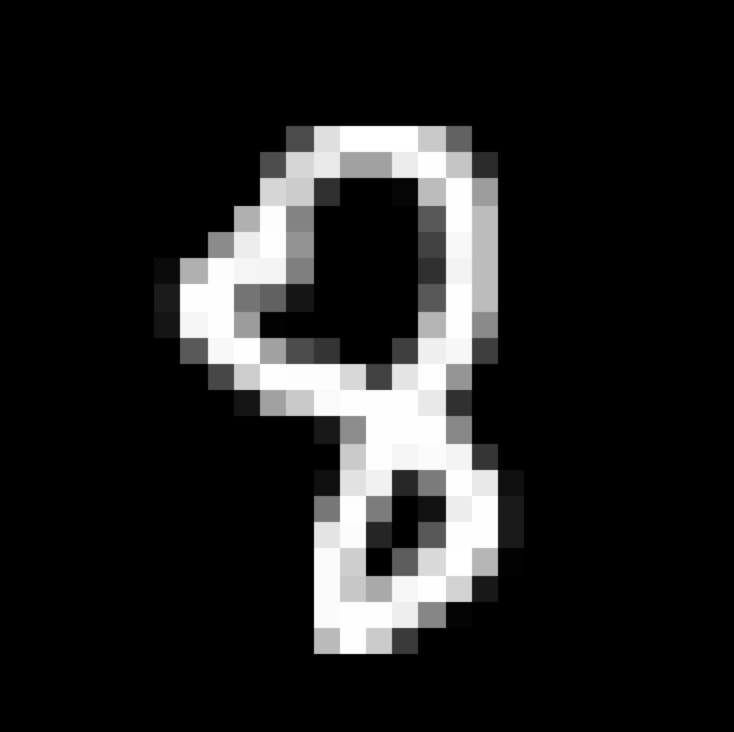}&\includegraphics[height=2cm,width=2cm]{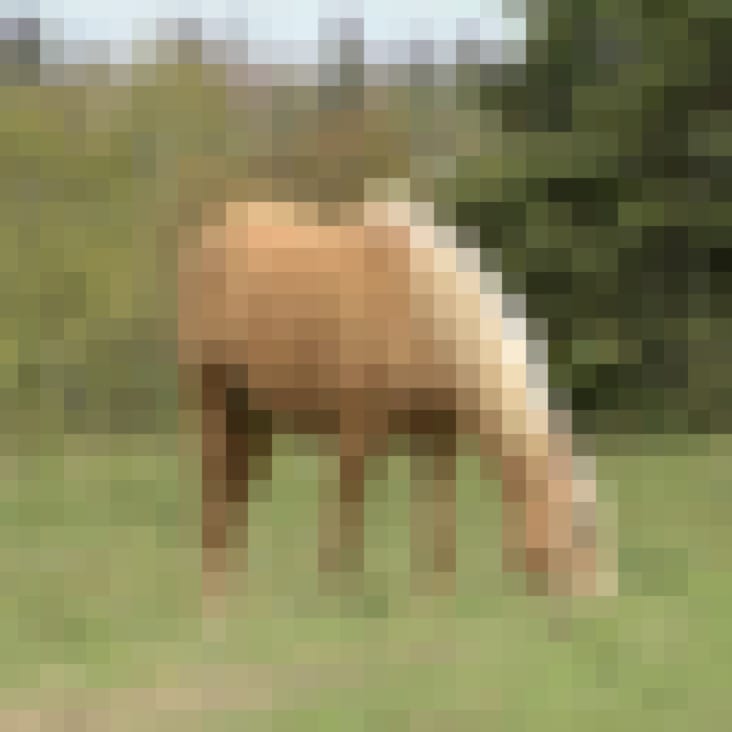}\\
		
		\includegraphics[scale=.05]{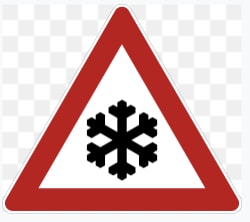}&(8)&Horse\\
		
		\includegraphics[height=2cm,width=2cm]{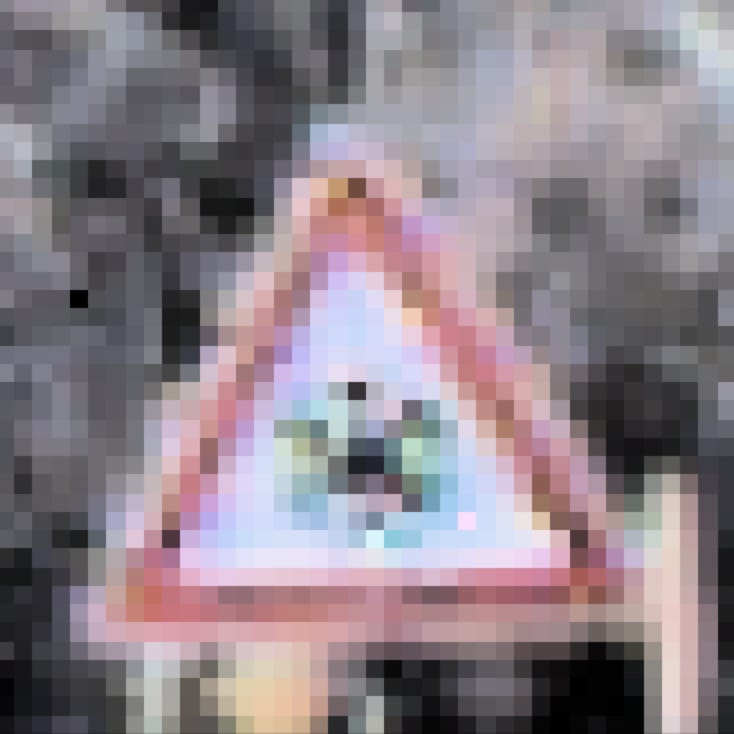}&\includegraphics[height=2cm,width=2cm]{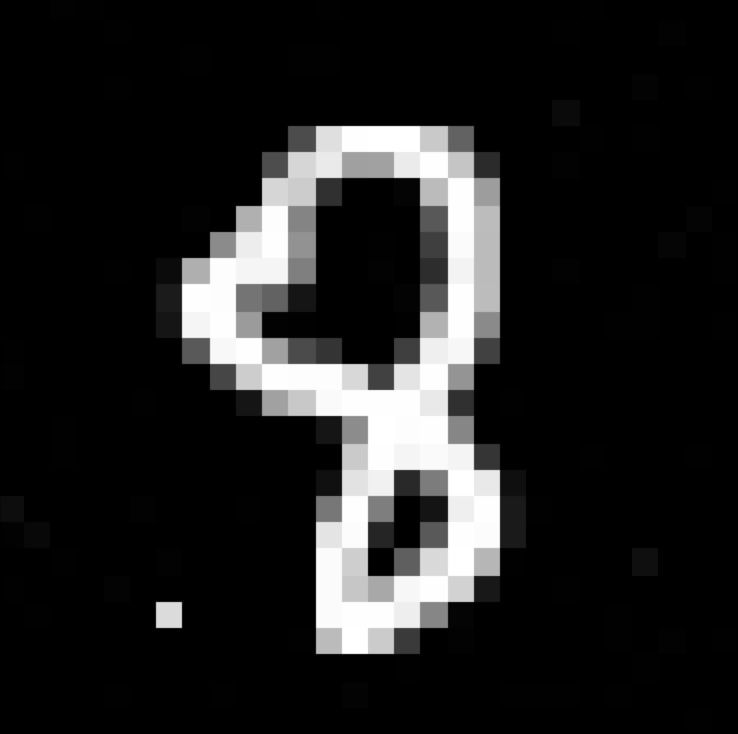}&\includegraphics[height=2cm,width=2cm]{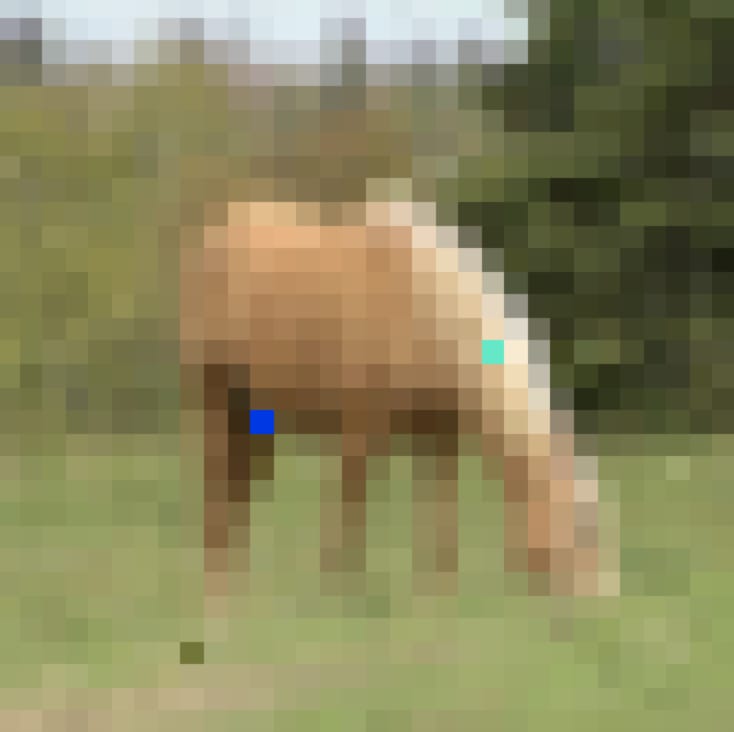}\\
		
		\includegraphics[scale=.05]{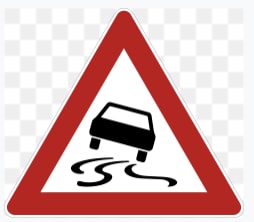}&(9)&Cat\\
		
(a)&(b)&(c)\\
		
	\end{tabular}
	\caption{Adversarial example generated by running human perception based CMA-ES; top: Legitimate samples from CIFAR10 dataset; bottom: Misclassified examples generated by CMA-ES for inputs from top row}
	\label{fig:perception_ex}
\end{figure}

\subsection{CMA-ES with Perception-in-the-Loop}
\label{subsec:human}

We now present adversarial input generation using CMA-ES with human perception.
The goal is to come up with more realistic adversarial examples with respect to human's judgment. 
In order to achieve this goal we do the following. 
First, CMA-ES is used to generate a population of inputs.
The original input and the entire population is displayed together on a screen.
However, any input classified the same way as the original input is hidden and displayed as a fully black square.
The user looks at the original image and the current set of adversarial examples. 
By comparing each example with the original image, the user selects the top $K$ examples closest to the original
input.
We do not restrict the time of the user to make a decision.

This process is demonstrated schematically in Fig.~\ref{fig:CMAES_Process}. 
For example in Fig.~\ref{fig:CMAES_GTSRB}, the top image shows a traffic sign and the next images show $3$ iterations 
of running CMA-ES.
In this example, the population size was set to $20$ and the user had to pick the top $5$ closest
images. 
Given the choice, CMA-ES updates its parameters and generates second set of examples. 
This process continues until the last iteration in which user selects the best example. 

Fig.~\ref{fig:perception_ex} demonstrates three sample outputs of running CMA-ES using human perception, where 
the number of iterations was $180$, population size was $4$, and parent size was $2$.
Thus, the user had to solve $180$ perception tasks.

\begin{figure}[t]
	\centering
	\begin{tabular}{ccc}
		\includegraphics[height=2cm,width=2cm]{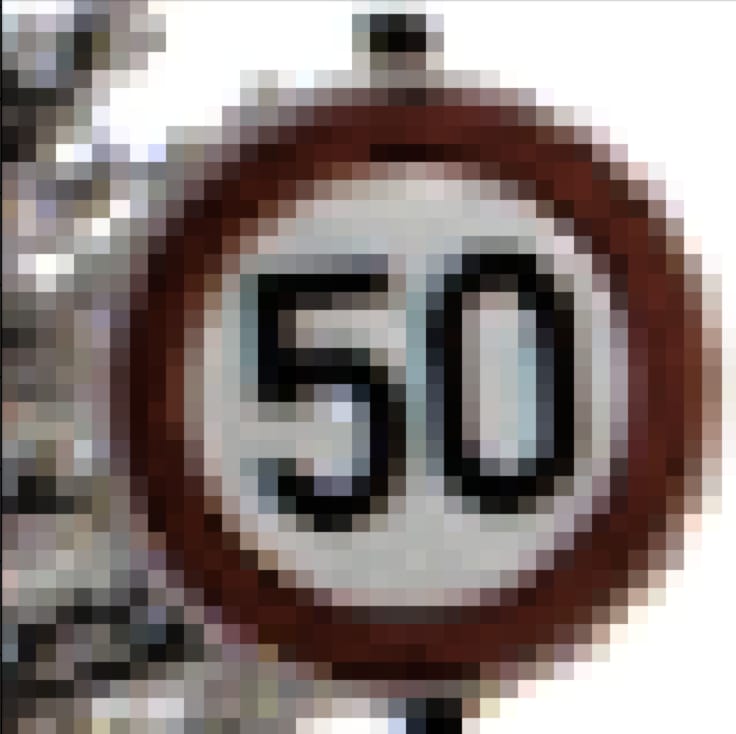}&\includegraphics[height=2cm,width=2cm]{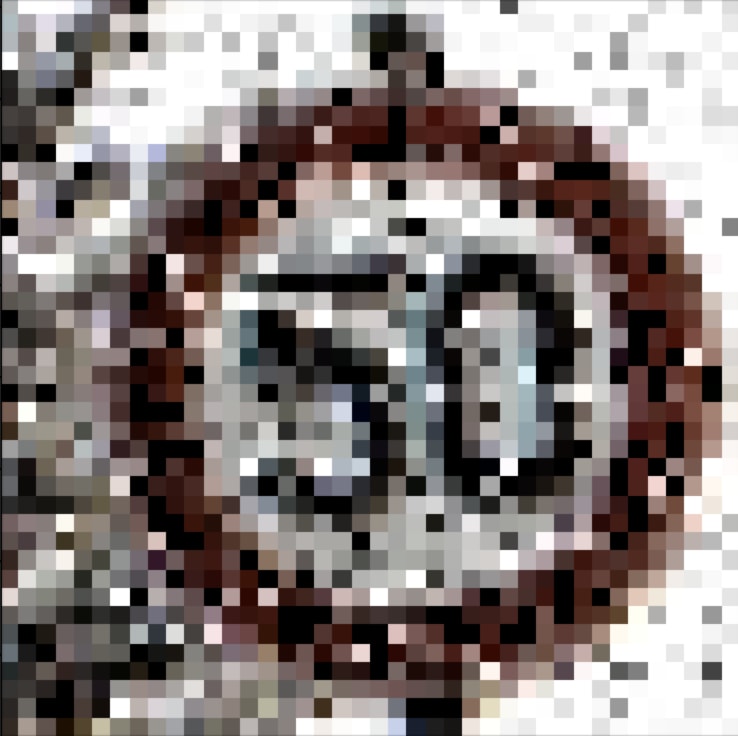}&\includegraphics[height=2cm,width=2cm]{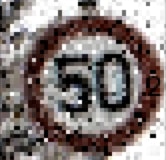}\\
		\includegraphics[scale=.05]{Figs/2_main}&\includegraphics[scale=.05]{Figs/5_main}&\includegraphics[scale=.05]{Figs/5_main}\\
		
		\includegraphics[height=2cm,width=2cm]{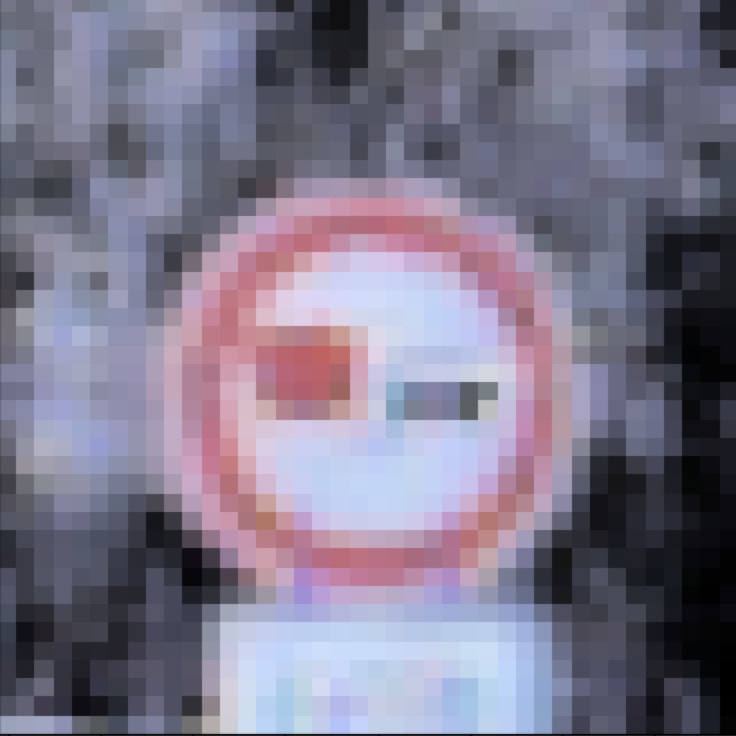}&\includegraphics[height=2cm,width=2cm]{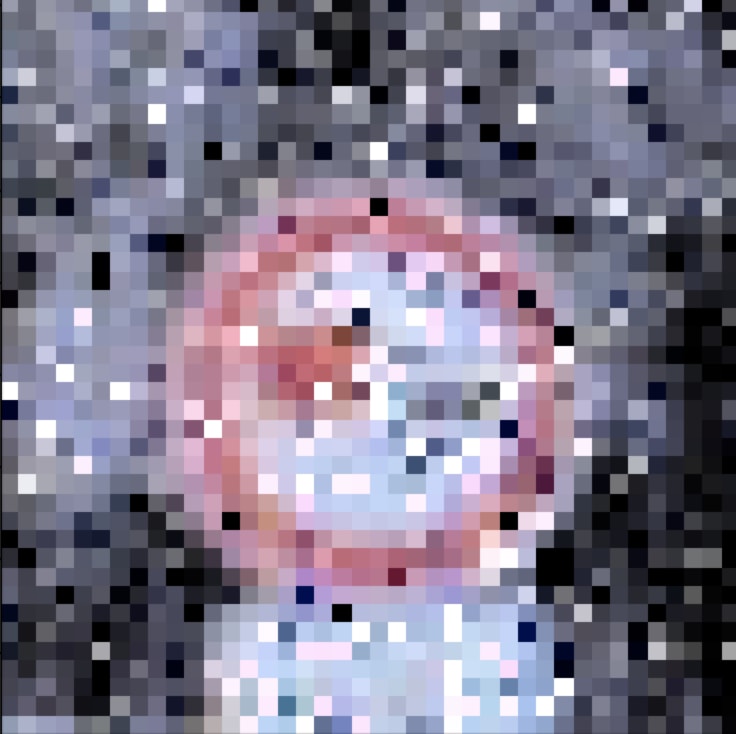}&\includegraphics[height=2cm,width=2cm]{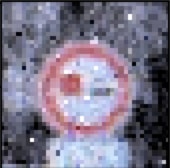}\\
		\includegraphics[scale=.05]{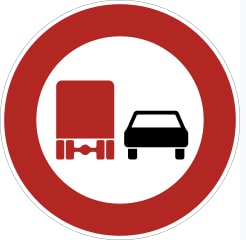}&\includegraphics[scale=.05]{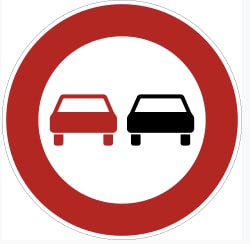}&\includegraphics[scale=.05]{Figs/o1_main}\\(a)&(b)&(c)\\
		
	\end{tabular}
	\caption{Comparing $L_1$ norm and human perception based adversarial example crafting, using pixel perturbation; (a): original image; (b): adversarial example crafted using $L_1$; (c)  adversarial example crafted using human perception}
	\label{fig:1_norm_vs_perception}
\end{figure}

Next, we evaluate CMA-ES with perception-in-the-loop when the number of iterations is reduced, in order to reduce
the burden on the user.
We consider CMA-ES with only $100$ iterations, with a population size of $4$ and parent size of $2$.
Fig.~\ref{fig:1_norm_vs_perception} demonstrates two examples in which images from GTSRB and corresponding adversarial samples crafted using 
$L_1$ norm and human perception, with pixel perturbations.
Subjectively, the human perception version looks ``better.''

\begin{figure}[t]
	
	\centering
	\begin{tabular}{ccc}
		\includegraphics[height=2cm,width=2cm]{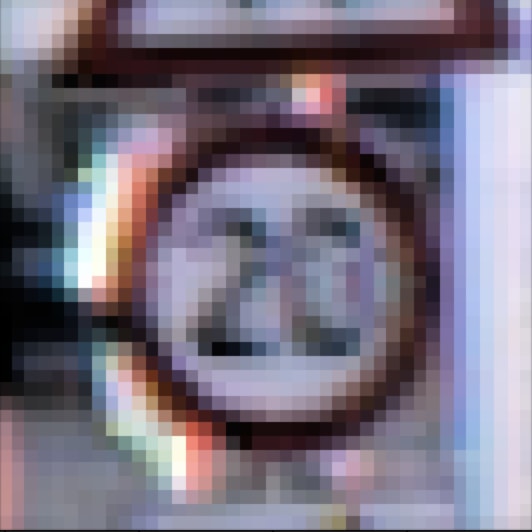}&\includegraphics[height=2cm,width=2cm]{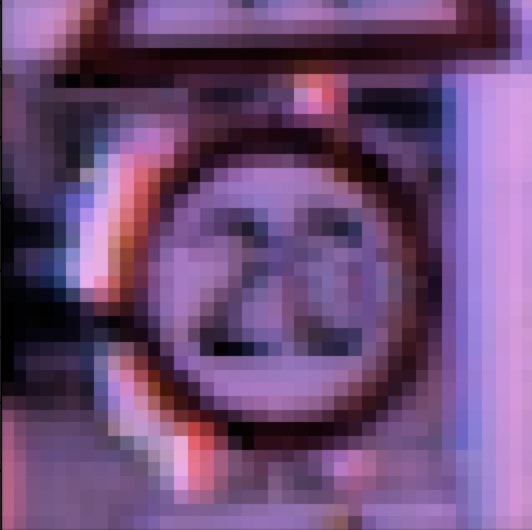}&\includegraphics[height=2cm,width=2cm]{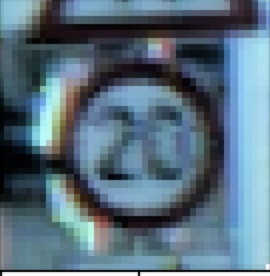}\\
		
		\includegraphics[scale=.05]{Figs/0_main} & \includegraphics[scale=.05]{Figs/2_main} & \includegraphics[scale=.05]{Figs/5_main} \\
		
		\includegraphics[height=2cm,width=2cm]{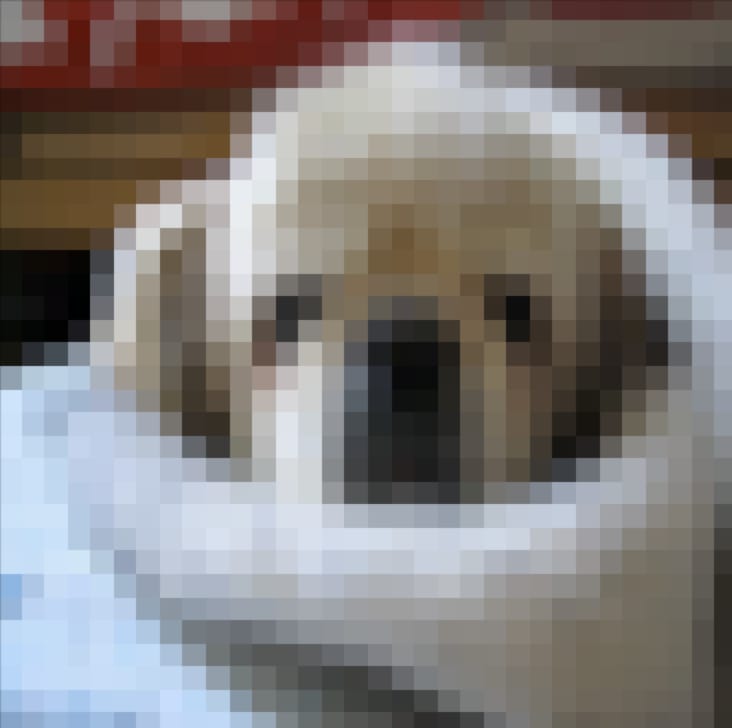}&\includegraphics[height=2cm,width=2cm]{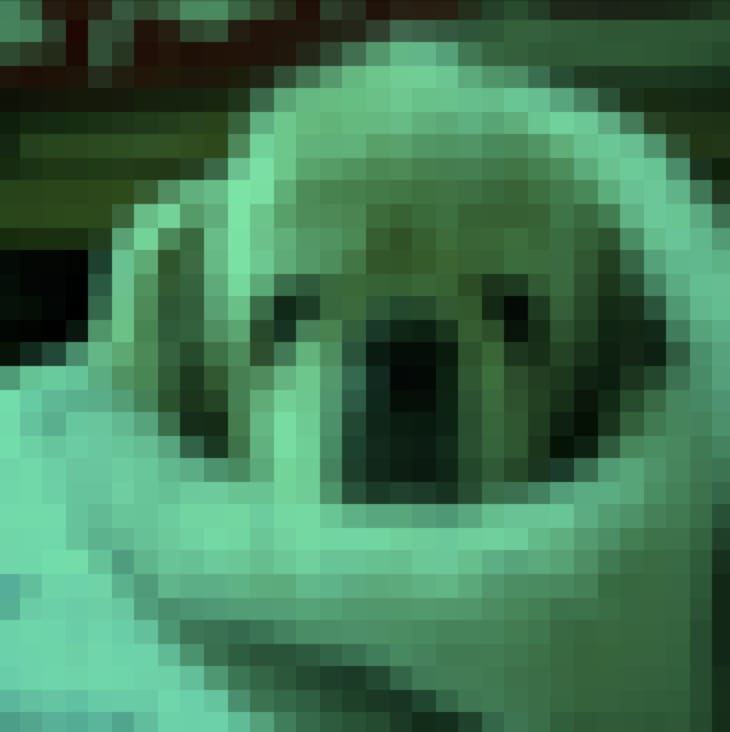}&\includegraphics[height=2cm,width=2cm]{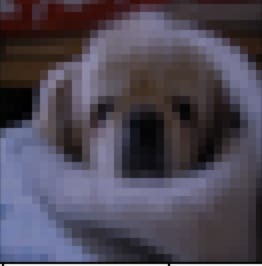}\\
		
		Dog & Frog & Cat \\
		
		\includegraphics[height=2cm,width=2cm]{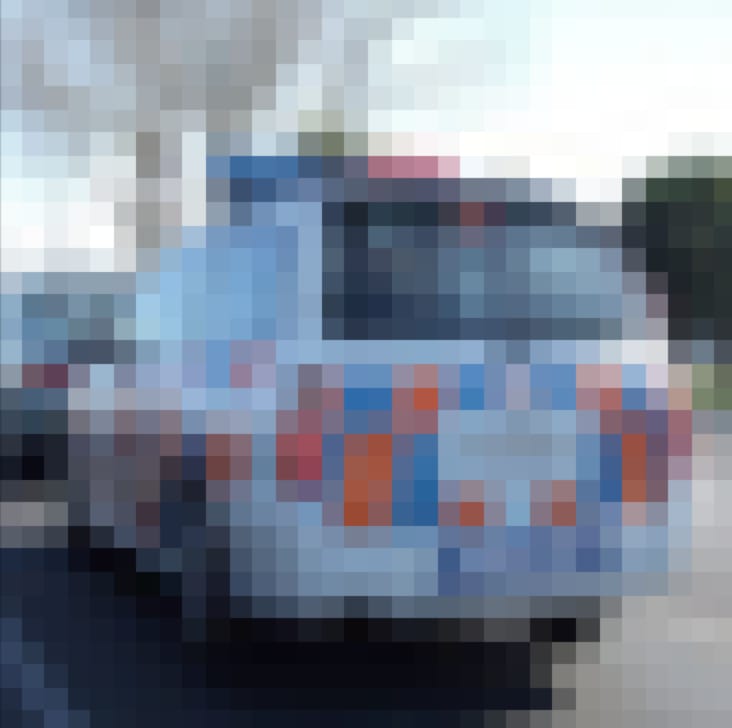}&\includegraphics[height=2cm,width=2cm]{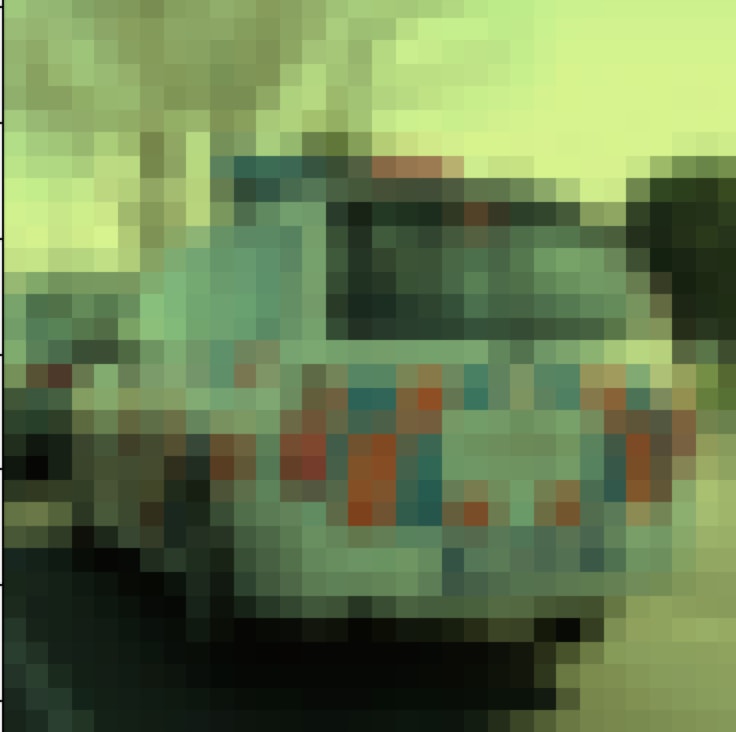}&\includegraphics[height=2cm,width=2cm]{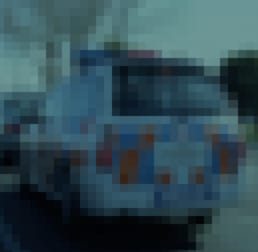}\\
		
	Car & Truck & Truck\\
	(a) & (b) & (c)\\
	\end{tabular}
	\caption{Comparing $L_1$ norm and human perception based adversarial example crafting, using color recombination; (a): original image; (b): adversarial example crafted using $L_1$; (c)  adversarial example crafted using human perception}
		\label{fig:1_norm_vs_perception2}
\end{figure}

The effects are even more marked when using color recombinations.
Fig.~\ref{fig:1_norm_vs_perception2} demonstrates several examples from GTSRB and CIFAR10 datasets as well as corresponding misclassifying 
image using $L_1$ and human perception, when the number of iterations was $3$ and the population size and parent size were $20$ and $5$, respectively. 
Subjectively, the color schemes for perception-in-the-loop was closer to the original image.

\subsection{Attacking Robustly Trained DNNs}

Finally, we consider adversarial input generation for DNNs which are trained to be robust against perturbations. 
We consider robust training based on abstract interpretation \cite{Vechev:2018}, 
which encodes $\epsilon$-robustness in $L_\infty$ norm into the loss function, for a user-defined parameter $\epsilon$.
The goal is to ensure that each training input has an $\epsilon$-neighborhood with the same classification.

We trained a DNN from \cite{Vechev:2018} for the MNIST dataset.
It has convolutional layers ($\times2$), maxpooling layers ($\times2$), and fully 
connected layer before the softmax layer (DNN C in Table~\ref{tb:DNN_LIST} \cite{Vechev:2018}). 
We trained this network using their ``box'' method and with $4$ different perturbation parameters ($\varepsilon$). 
After training for $180$ epochs, we evaluated the robustness of the network against two attacks: 
fast gradient sign method (FGSM) with the same $\varepsilon$ used in training 
and CMA-ES with perception in the loop. 

We ran CMA-ES for $3$ iterations, with population size $240$ and parent size $120$.
CMA-ES was agnostic to the $\epsilon$ parameter, 
and considered those generated examples which could be classified 
by human eyes correctly, as adversarial examples. 
Table~\ref{tb:robust_DiffAI} summarizes the results. 
While FGSM had a low success rate, CMA-ES was able to synthesize perceptually similar adversarial examples,
albeit with a lower success rate than DNN A or B with standard training.

\begin{figure}[H]
	
	\centering
	\begin{tabular}{cccc}
		\includegraphics[scale=.08]{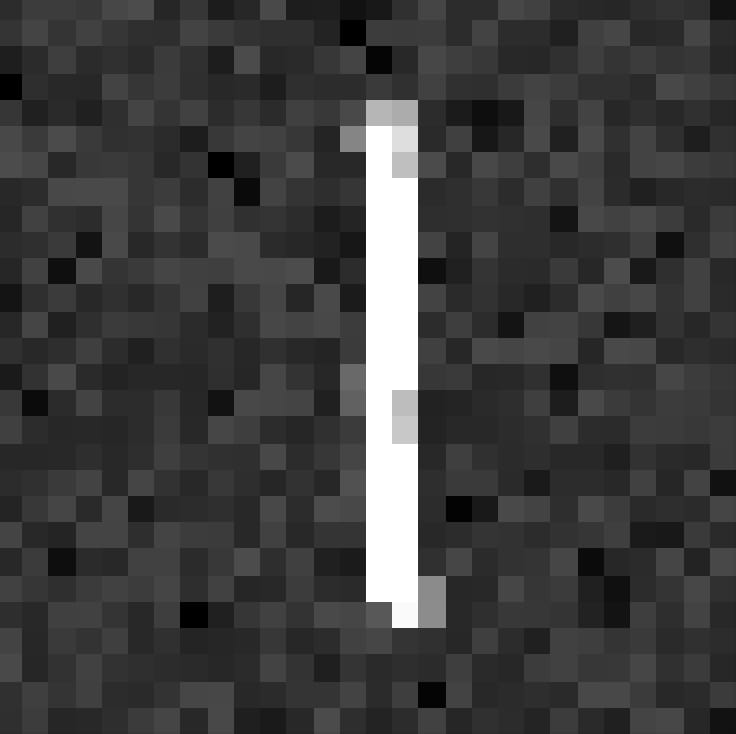}&\includegraphics[scale=.08]{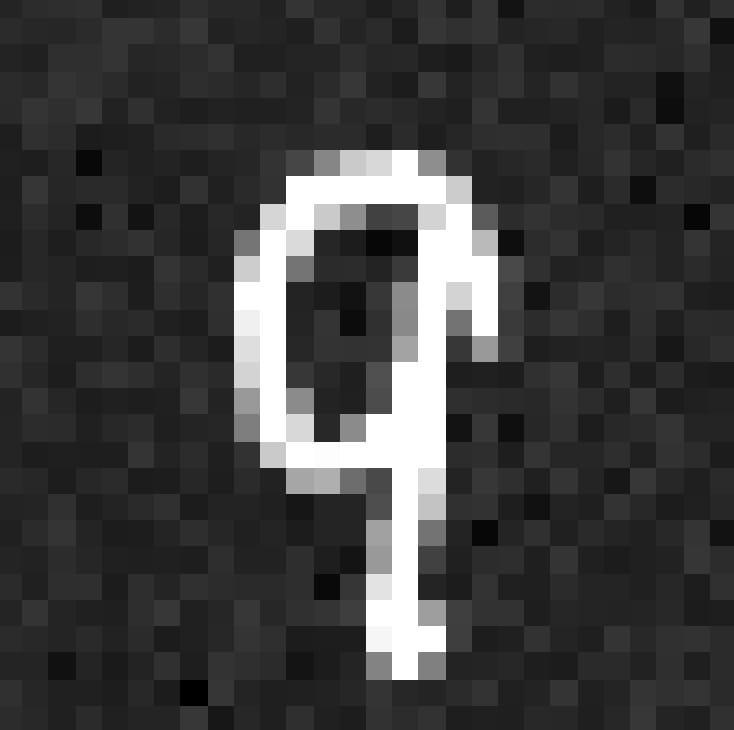}
		
		\includegraphics[scale=.08]{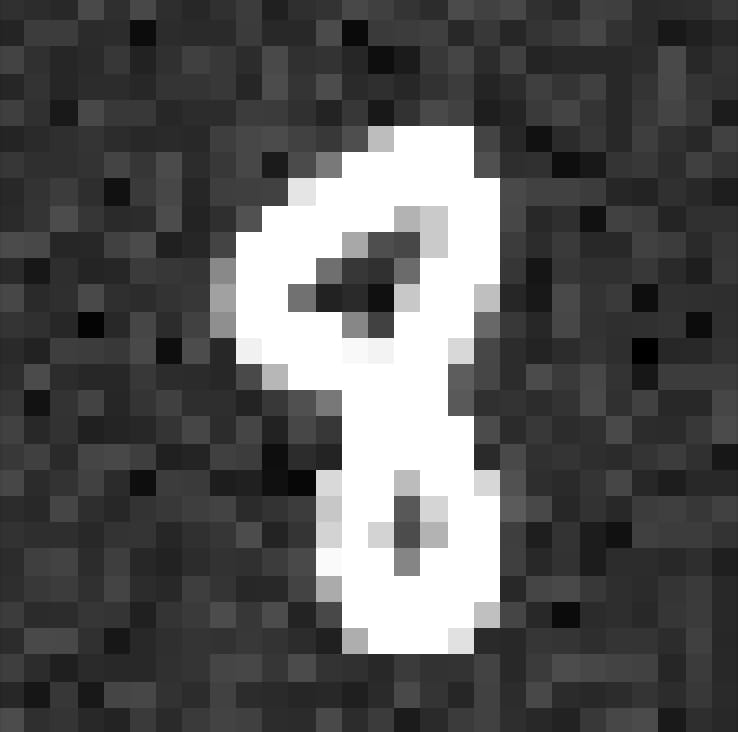}&\includegraphics[scale=.08]{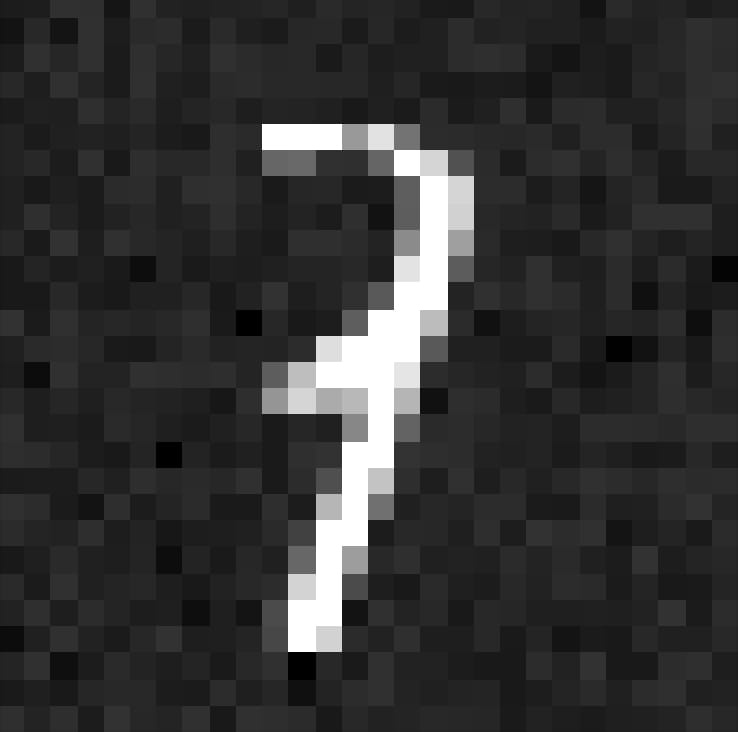}\\
	\end{tabular}
	\caption{Some examples of misclassified images crafted on robustly trained CNN with \cite{Vechev:2018}}
	\label{fig:robust_MNIST}
\end{figure}

An implication of this experiment is that while for some $\varepsilon$, the
network is quite robust with respect to $L_{\infty}$ norm, 
it is still prone to attacks which are based on perceptual norm.
Intuitively, this means that the measuring robustness of trained NNs solely using $L_{p}$ norms is not sufficient. 
Fig.~\ref{fig:robust_MNIST} demonstrates some examples crafted by CMA-ES in $180^{th}$ epoch.

\begin{table}[t]
	\begin{center}
		\caption{  Comparison between success rates of FGSM (based on $L_{\infty}$)and CMA-ES (based on perceptual norm)}\label{tb:robust_DiffAI}
		
		\begin{tabular}{|c|c|c|c|c|c|}
			$\varepsilon$ &0.02& 0.05& 0.1 &  0.15 & 0.2 \\\hline
			FGSM &0\%& 0\%&  10\% & 23.3\% & 36.7\% \\\hline
			CMA-ES  & $46.7\%$ &46.7\% & $73.3\%$ & $76.7\%$ &$83.3\%$\\\hline
		\end{tabular}
	\end{center}
\end{table}




\section{Conclusion}

We have presented a technique to construct adversarial examples based on CMA-ES, an efficient black box optimization technique.
Our technique is black box and can directly use human perceptual preferences.
We empirically demonstrate that CMA-ES can  efficiently find perceptually similar adversarial inputs on several
widely-used benchmarks.
Additionally, we demonstrate that one can generate adversarial examples which are perceptually similar,
even when the underlying network is trained to be robust using $L_p$ norm.
While we focus on incorporating perceptual preferences in crafting adversarial inputs, the role of direct
perceptual preferences in mitigating against attacks is left for future work.

\nocite{*} 


\end{document}